%% file: main.tex
\documentclass[conference]{IEEEtran}
\IEEEoverridecommandlockouts
\input{preamble}

\def\BibTeX{{\rm B\kern-.05em{\sc i\kern-.025em b}\kern-.08em
    T\kern-.1667em\lower.7ex\hbox{E}\kern-.125emX}}
\begin{document}

\title{Non-Invasive Fairness in Learning through the Lens of Data Drift}

\author{\IEEEauthorblockN{Ke Yang}
\IEEEauthorblockA{\textit{Manning College of Information \& Computer Sciences} \\
University of Massachusetts\\
Amherst, MA, U.S.A. \\
key@umass.edu}
\and
\IEEEauthorblockN{Alexandra Meliou}
\IEEEauthorblockA{\textit{Manning College of Information \& Computer Sciences} \\
University of Massachusetts\\
Amherst, MA, U.S.A. \\
ameli@cs.umass.edu}}

\maketitle

\input{abstract}

\begin{IEEEkeywords}
data management, fairness, data profiling
\end{IEEEkeywords}

\input{intro}
\input{ps}

\input{overview}
\input{approach}
\input{exp_supp}

\input{related}

\input{conc}

\bibliographystyle{IEEEtran}
\bibliography{IEEEabrv,fairCC}

\end{document}

%% file: preamble.tex
\usepackage{cite}
\usepackage{amsmath,amssymb}
\usepackage{stmaryrd}

\usepackage{textcomp}
\usepackage{xcolor}

\usepackage{amsthm}
\usepackage{bbm}
\usepackage{amsfonts}       
\usepackage{nicefrac}       

\usepackage{xspace}
\usepackage{listings}
\usepackage[shortlabels]{enumitem}
\usepackage{algorithm}
\usepackage{algorithmicx}
\usepackage[noend]{algpseudocode}
\input{algPatch}

\algnewcommand{\LineComment}[1]{\Statex{\color{gray}{\small$\triangleright$\textit{#1}}}}

\usepackage{mathtools}
\usepackage[position=bottom]{subfig}
\usepackage{url}    
\usepackage{booktabs}
\usepackage{array}
\usepackage{multirow}
\usepackage{dsfont}
\usepackage{makecell}
\usepackage{mdframed}
\usepackage{diagbox}
\usepackage{adjustbox}
\usepackage{balance}

\usepackage{csquotes}
\usepackage{hyperref}
\hypersetup{%
    colorlinks = true,
    linkcolor = black,
    citecolor = black,
    urlcolor = black,
}

\usepackage{tabularx}

\usepackage{colortbl}

\definecolor{lightgray}{gray}{0.9}

\newtheorem{definition}{Definition}

\newtheorem{example}{Example}

\newcommand{\rev}[1]{\textcolor{black}{#1}}
\newcommand{\revs}[1]{\textcolor{black}{#1}}

\newcommand{\base}{\textsc{no-intervention}\xspace}
\newcommand{\mbase}{\textsc{MultiModel}\xspace}
\newcommand{\mcc}{\textsc{DifFair}\xspace}
\newcommand{\scc}{\textsc{ConFair}\xspace}
\newcommand{\kam}{\textsc{KAM}\xspace}
\newcommand{\capu}{\textsc{CAP}\xspace}
\newcommand{\omn}{\textsc{OMN}\xspace} 

\newcommand{\dro}{\textsc{DRO}\xspace}

\newcommand{\reffig}[1]{Fig.\ref{#1}}

\newcommand{\dmp}{\textit{MEPS}\xspace}
\newcommand{\dlg}{\textit{LSAC}\xspace}
\newcommand{\dcr}{\textit{Credit}\xspace}
\newcommand{\dap}{\textit{ACSP}\xspace}
\newcommand{\dah}{\textit{ACSH}\xspace}
\newcommand{\dae}{\textit{ACSE}\xspace}
\newcommand{\dai}{\textit{ACSI}\xspace}

\newcommand{\ebacc}{\textit{BalAcc}\xspace}
\newcommand{\edi}{\textit{DI}\xspace}
\newcommand{\eeo}{\textit{Equalized Odds}\xspace}
\newcommand{\ead}{\textit{AOD}\xspace}

\newcommand{\eg}{e.g.,\xspace}
\newcommand{\ie}{i.e.,\xspace}
\newcommand{\etal}{et al.\xspace}

\usepackage{graphicx,calc}
\newlength\myheight
\newlength\mydepth
\settototalheight\myheight{Xygp}
\newcommand*\inlinegraphics[1]{%
  \settototalheight\myheight{Xygp}%
  \settodepth\mydepth{Xygp}%
  \raisebox{-\mydepth}{\includegraphics[height=\myheight]{#1}}%
}

%% file: abstract.tex

\begin{abstract}
Machine Learning (ML) models are widely employed to drive many modern data
systems. While they are undeniably powerful tools, ML models often demonstrate
imbalanced performance and unfair behaviors. The root of this problem often
lies in the fact that different subpopulations commonly display divergent
trends: as a learning algorithm tries to identify trends in the data, it
naturally favors the trends of the majority groups, leading to a model that
performs poorly and unfairly for minority populations. Our goal is to improve
the fairness and trustworthiness of ML models by applying only \emph{non-invasive} interventions, i.e., without altering the data or the learning algorithm. We use a simple but key insight: the divergence of trends between different populations, and, consecutively, between a learned model and minority populations, is analogous to data drift, which indicates the poor conformance between parts of the data and the trained model.

We explore two strategies (model-splitting and reweighing) to resolve this drift,
aiming to improve the overall \emph{conformance} of models to the underlying
data. Both our methods introduce novel ways to employ the recently-proposed
data profiling primitive of Conformance Constraints.
Our splitting approach is based on a simple data drift strategy: training
separate models for different populations. Our \mcc algorithm
enhances this simple strategy by employing conformance constraints, learned
over the data partitions, to select the appropriate model to use for predictions on each serving tuple. 
However, the performance of such a multi-model strategy can degrade severely
under poor representation of some groups in the data. We thus propose a single-model, reweighing strategy, \scc, to overcome this limitation. \scc employs conformance constraints in a novel way to derive weights for training data, which are then used to build a single model.
Our experimental evaluation over 7 real-world datasets shows that both
\mcc and \scc improve the fairness of ML models.
We demonstrate scenarios where \mcc has an edge, though
\scc has the greatest practical impact and outperforms other baselines.  Moreover, as a model-agnostic technique, \scc stays robust when used against different models than the ones on which the weights have been learned, which is not the case for other states of the art. 

\end{abstract}

%% file: intro.tex

\section{Introduction}
\label{sec:intro}
While Machine Learning (ML) models are widely employed in many modern data systems for their undeniable predicting power, they often demonstrate imbalanced performance and unfair behaviors \rev{(\eg different model performance across subpopulations)}. Such fairness issues in ML models have been extensively studied within the machine learning and data management communities, among others, in the past decade~\cite{DBLP:journals/datamine/CaldersV10,DBLP:journals/kais/KamiranC11,DBLP:conf/icdm/KamiranKZ12,DBLP:conf/pkdd/KamishimaAAS12,feldman2015certifying,DBLP:conf/nips/HardtPNS16,DBLP:conf/nips/CalmonWVRV17,DBLP:conf/kdd/ZhangWW17,DBLP:conf/nips/KusnerLRS17,DBLP:conf/aies/ZhangLM18,DBLP:journals/corr/abs-1808-00023,DBLP:conf/nips/LiptonMC18,DBLP:conf/icml/KearnsNRW18,DBLP:conf/icml/AgarwalBD0W18,DBLP:conf/icml/AgarwalDW19,johndrow2019algorithm,DBLP:conf/fat/CelisHKV19,SalimiRHS2019,doi:10.1146/annurev-statistics-042720-125902}. 

\emph{In this paper, we recast these fairness issues as a problem of data drift, and we address it with solutions that directly aim to improve the conformance between data and model.} Data drift may occur when the deployed data of a model changes compared to the training set, leading to degradation in the model's performance. Similarly, data drift over subpopulations (or groups for brevity) can cause a model to perform poorly over minority groups in the development and deployment of the model.
This is because different groups commonly exhibit distinct patterns in the distribution of their attributes and labels. As a learning algorithm attempts to identify a pattern within a given population, it tends to prioritize the pattern of the majority group\footnote{We use majority and minority groups to refer to the populations that are over- and under-represented, respectively, in the data or in the preferred ground truth labels for ML models.} due to their prevalence. The produced model thus does not \emph{conform} to the minority group, and, as a result, its predictions for members of that group are less reliable. 

\begin{figure}[t]
\centering
\includegraphics[width=0.4\textwidth]{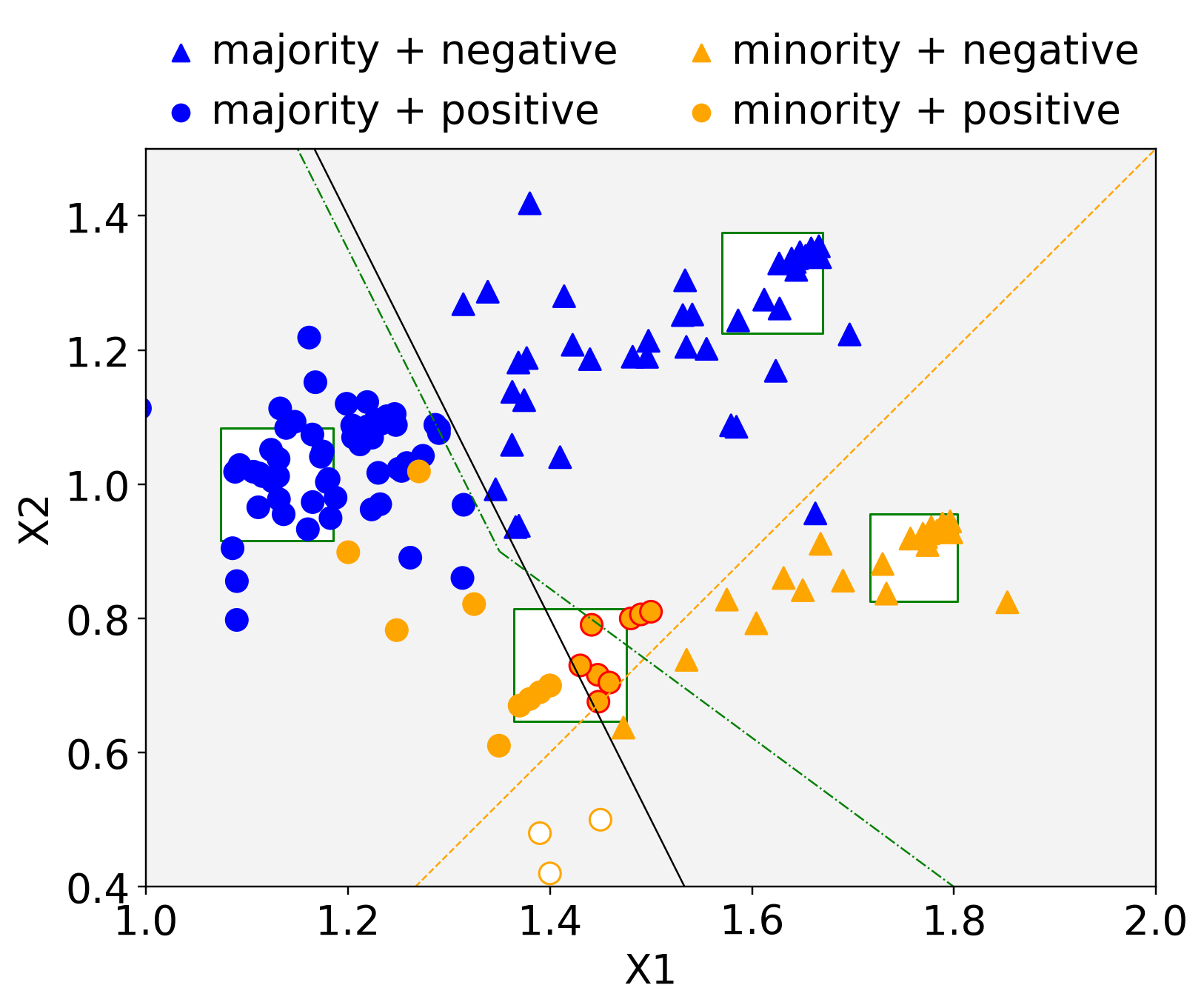}
\vspace{-1em}	
\caption{\looseness-1
An example of input data containing two groups: majority and minority color-coded in blue and orange, respectively. The attributes $X_1$ and $X_2$ of these groups show dissimilar distributions, indicating a drift over groups. An unfair model (black line) prioritizes the pattern of the majority group (blue points), and predicts poorly \rev{(with fewer positive outputs)} for minorities (\eg orange points with red outline). \mcc improves the conformance between data and model by building separate models for different groups (\eg orange dashed line for minorities). \scc improves the conformance by deriving a single model (green dash-dot line) that emphasizes the densest areas (green squares) of the input data for both groups.
}
\vspace{-1em}
\label{fig:example}
\end{figure}

\begin{example}
The dataset in \reffig{fig:example} contains two groups, which are color-coded in blue and orange. The attributes $X1$ and $X2$ of these groups show dissimilar distributions, as can be observed from the $x$ and $y$-axis, respectively, indicating a drift over groups.  The positive and negative ground truth labels for a classification model are marked by circles and triangles, respectively. A model trained on this dataset (black line) tends to conform to the majority group (blue points). As a result, a significant number of minority records (orange points) receive incorrect predictions (orange points with red outline). 
\label{exp:unfair}
\end{example}

Our solutions to address this drift over groups are \emph{non-invasive}, \ie they do not alter the data or the learning algorithm. Instead, we aim to \emph{improve the conformance between data and model} with two strategies: \emph{model-splitting} and \emph{reweighing}.

\textbf{Our model-splitting approach, \mcc,} is designed around a simple strategy to address data drift over groups: training separate models for different groups, such that each produced model conforms to its training data better. 

\begin{example}
For the dataset in \reffig{fig:example}, \mcc produces a separate model (orange dashed line) for the minority group (orange points), which is better served by such a model that is significantly different from the overall model (black line). \mcc also produces a model for the majority group, which is closely aligned with the black line in \reffig{fig:example} (not displayed to avoid visual clutter).
By building models that conform to different groups, \mcc reduces the number of incorrect predictions for the minority, leading to a fair outcome, \rev{\ie the ratio of positive model outcomes is similar for both groups.} 
\label{exp:mcc}
\end{example}

Similar strategies of developing and deploying multiple models to address drift in unseen data have also been employed in production settings of ML models~\cite{DBLP:journals/jmlr/BickelBS09,
DBLP:journals/sigmod/KumarMNP15, DBLP:conf/icml/LiptonWS18,
DBLP:journals/sigmod/PolyzotisRWZ18, DBLP:conf/sigmod/SchelterRB20}. 
\rev{For example, ensemble learning combines the output of several models in some way to derive predictions; \mcc can be viewed as a technique within this family, where the combination of classifier outputs is based on data drift.}
The novelty of our approach lies in the use of Conformance Constraints (CCs) \rev{to identify data drift}, a data profiling primitive that automatically learns from a given dataset numerical constraints that summarize the densest areas in the input data~\cite{DBLP:conf/sigmod/Fariha0RGM21}. 

\begin{example}
Profiling the four subsets of the data in \reffig{fig:example} (blue circles, blue triangles, orange circles, and orange triangles) using CCs, results in four sets of constraints (depicted as green rectangles). 
Each set of constraints describes the densest areas in the corresponding subset of the data by some distributive patterns of the attributes. For example, the constraints for the minority positive group (orange circles) specify the rectangular region $1.38 \leq X1 \leq 1.5 \wedge 0.68 \leq X2 \leq 0.8$. The distance of a point from this region positively corresponds to the point's violation of these constraints, while points inside the region get zero violations. 
\label{exp:cc_rule}
\end{example}

Based on tuples' violation of the produced sets of constraints, \mcc selects an appropriate model to use during the deployment of group-dependent models, such that the chosen model best conforms to the serving data.
\mcc has three advantages over the naive approach of using group membership to separate models: (1)~it may not be possible to use group membership related to sensitive or protected attributes (e.g., gender, race, disability status, etc.) due to legal and discrimination considerations; (2)~demographic attributes may often be unreliable due to privacy and discrimination considerations, which is an issue more likely to affect minority groups~\cite{kappelhof2017survey}; (3)~individuals may deviate from their own group's pattern and may be served better by another group's model, requiring a more sophisticated way to deploy models. \rev{The alternative of learning a separate model to predict which model to deploy is prone to mistakes or drift in the labels, while \mcc is more robust to such issues.}

A general limitation of model-splitting approaches, however, is that their performance can degrade significantly when a group's representation is particularly poor in the data. For example, if the population of a group is very small, or if its labels are severely skewed (e.g., mostly negative labels), the model trained on such data will likely be of low quality. Suppose a minority group has 90\% negative and 10\% positive labels; a model that always assigns a negative prediction may achieve high accuracy over this data, but it is clearly unreliable.  This limitation is hard to address with a model-splitting approach, as small data size and skewed representation offer little opportunity for improving the models.

\textbf{Our reweighing approach, \scc,} is a single-model strategy designed to overcome this limitation. \scc profiles the data using CCs to identify the densest areas of the input, and assigns weights to each tuple based on its violation of the produced sets of constraints. High weights are assigned to those tuples that conform to the corresponding set of constraints (\ie with zero violation). These weights are then used in model training. For models that do not support weights directly, they can still employ a weighted sampling strategy to preprocess the training data accordingly. 

\begin{example}
For the dataset in \reffig{fig:example}, \scc assigns high weights to the tuples located within the constraint areas (e.g., points inside green squares). The produced model (green dash-dot line) manages to correct several erroneous predictions of the original model (black line), \ie most of the red-outlined orange points are now correctly classified.  \rev{The two groups also get similar ratios of positive predictions,}
indicating a fair outcome.
\label{exp:scc}
\end{example}

\looseness-1
Reweighing strategies have been used in prior art to improve the fairness of ML models~\cite{DBLP:journals/kais/KamiranC11,hashimoto2018fairness,SalimiRHS2019,lahoti2020fairness,zhang2021omnifair}. The intuition of such strategies is that balancing the weighted representation of groups can amplify the loss of the minority group during training, thus leading to models that better optimize for this loss. Much of the prior work focuses on adjusting the weights during iterative training of a model~\cite{hashimoto2018fairness,lahoti2020fairness,jiang2020identifying}. \rev{Such interventions learn the weights through a black-box training process, which cannot be audited or adjusted.} 
In contrast, \scc supports \emph{flexible intervention}:
by allowing users to control the reweighing impact, they can adjust the tradeoff between fairness and accuracy. Moreover, \scc follows a \emph{non-invasive} strategy that does not alter the data or model.
Among other non-invasive techniques~\cite{DBLP:journals/kais/KamiranC11, zhang2021omnifair}, \scc stands out by allowing variability in the weights assigned to the members of a subpopulation. Instead of assigning identical weights to all tuples within a minority group, \scc only increases the weights of those individuals that conform to the densest part of the group's data.  This way, \scc avoids amplifying outliers and noise, which could mislead the training and harm model accuracy.
Figure~\ref{tab:scc_comparison} summarizes these points of comparison between \scc and prior art.

\input{figs/tab-scc-novelty.tex}

\textbf{Model-splitting vs reweighing.}
\scc and \mcc are designed to support different scenarios of data drift over groups. In cases of significant drift, \mcc is generally better, as it may not be possible to build a single well-conforming model (see evaluation in Section~\ref{sec:exp:mcc}).

\begin{example}
For the dataset in \reffig{fig:example}, \scc does not resolve all erroneous predictions for the minority group, \ie red-outlined points still fall on the wrong side of the green dash-dot line.  In contrast, \mcc can produce a model (orange dashed line) that better conforms to the minority group.
\label{exp:scc_lim}
\end{example}

When drift over groups is less stark, \scc can be more effective than \mcc as it applies an early-stage intervention (focusing on the training data), while avoiding the loss of predicting power in splitting input and developing multiple group-dependent models. 

\smallskip
\noindent\textbf{Scope.} 
In this paper, we aim to improve the fairness of ML models by increasing the conformance between the model and data. Our work focuses on \emph{group fairness}, which characterizes if any group, collectively, is discriminated against. \rev{In this paper, we focus on group fairness measured by disparate impact~\cite{DBLP:journals/datamine/CaldersV10,DBLP:conf/innovations/DworkHPRZ12,feldman2015certifying}, but our approach also supports other fairness metrics (\eg Equalized Odds), discussed in Section~\ref{sec:exp:scc}.} 
These groups are often defined by demographic attributes, such as gender, race, disability status, etc., but this is not a requirement for our methods.  

\textbf{In relation to methods in the fairness literature}, our approach focuses on data-oriented interventions but requires no invasive changes to the data itself. 
Compared to those methods that alter the data directly (known as pre-processing interventions)~\cite{DBLP:journals/kais/KamiranC11,feldman2015certifying,DBLP:conf/nips/CalmonWVRV17,DBLP:conf/kdd/ZhangWW17,SalimiRHS2019}, our approach may be less powerful due to the non-invasive setting, while the former allows arbitrary changes to the data such that one can achieve greater flexibility in obtaining desired fairness improvement.
However, by being non-invasive, our approach poses a lower risk of introducing unintended drift between the training and serving data. Furthermore, we take into account the distribution of numerical attributes, providing a rich space for fine-tuning the balance between fairness and utility, and enabling our approach to be easily combined with others that operate in the categorical domain. 
Our approach is also different from the methods that alter the learning algorithms or the outcomes directly~\cite{DBLP:conf/pkdd/KamishimaAAS12,DBLP:conf/icdm/KamiranKZ12,DBLP:conf/nips/HardtPNS16,DBLP:conf/nips/PleissRWKW17,DBLP:conf/aistats/ZafarVGG17,DBLP:conf/aies/ZhangLM18,DBLP:conf/icml/KearnsNRW18,DBLP:conf/icml/AgarwalBD0W18,DBLP:conf/icml/AgarwalDW19,DBLP:conf/fat/CelisHKV19,thomas2019preventing}, known as in- or post-processing interventions. These methods often require access to models or learning algorithms to fine-tune (or reassign) the loss for each data point during the development (or deployment) of fair models, making them less interpretable and difficult to audit due to technical complexity. In contrast, our approach is explicit and easy to interpret and audit. 
\rev{Our techniques rebalance fairness for specified minority and majority groups. This process may lead to imbalances in the treatment of other unidentified subpopulations, which is a common effect in fairness repairs (\eg repairing fairness w.r.t. gender may lead to imbalances w.r.t. race)~\cite{martinez2020minimax,krishnaswamy2021fair}.}

\textbf{In relation to clustering}, our approach is designed for supervised learning tasks rather than unsupervised settings such as clustering tasks. Fair clustering tasks may differ based on their definitions of fairness (we refer the reader to a survey~\cite{chhabra2021overview} for more detail). 
The employment of conformance constraints in our approach resembles clustering, \rev{but the two have different objectives (identifying clusters vs determining dense areas of the input data).}
\rev{Clustering may be repurposed to perform the same task, but it} is not an effective alternative to the use of numerical constraints as in CCs. This is because most clustering techniques are sensitive to the separation of clusters in input data, requiring that clusters are well separated from each other. This assumption is not valid in much of our experimental data, where drift over groups (or clusters) exists but the groups are not clearly separated in the input space. Moreover, clustering methods are less useful in scenarios where individuals may deviate from their own cluster and would receive better outcomes if they were assigned to another cluster (\eg assigned to the model for another group in \mcc). \rev{By analyzing the distribution of attributes, CCs are more robust towards data drift than clustering.}

\textbf{Considering other data profiling primitives}, CCs offer two important advantages. (1)~Focusing on the numerical attributes provides rich data context and great flexibility in achieving desired fairness balance, which has not been exploited in deriving fairness interventions. Additionally, the focus on a continuous domain makes our approach orthogonal to the methods that work in a categorical domain to derive interventions, thus presenting the potential in combining with the latter to further improve the fairness of ML models. (2)~Constraints can be derived efficiently over large datasets (\ie linear in the number of tuples and cubic in the number of attributes), which makes our methodology practical for real-world data. Ultimately, our approach can integrate with other profiling tools that produce similar quantitative descriptions of input data.

\smallskip
\noindent\textbf{Contributions.} 
In this paper, we make the following contributions:

\begin{itemize}[topsep=1pt, leftmargin=2ex]
    \item We recast the problem of fairness in ML models as an issue of drift over groups in input data, and, consecutively, as a problem of conformance of the model to its underlying data. (Section~\ref{sec:ps})
    
    \item We present \mcc, a model-splitting strategy that improves conformance between model and data by deriving group-dependent models and deploying these models based on the similarity of serving tuples to the training data of each model. Experiments show that \mcc is a better solution to improve the fairness of ML models for scenarios, where a single model is impossible to conform to all groups of input data. (Section~\ref{sec:method_driftcc})
    
    \item We present \scc, a single-model strategy that reweighs the training tuples based on the densest areas of input data, thus producing a single model with balanced predictive accuracy across groups. Experiments show that \scc outperforms existing reweighing techniques, and remains robust when its weights are used by different learning algorithms, in contrast with other prior art. (Section~\ref{sec:method_faircc})
    
    \item We augment our techniques with density estimation to improve the tightness of derived conformance constraints. (Section~\ref{sec:method_density})

    \item We evaluate our methods against 7 real-world datasets and 4 alternative approaches. We demonstrate gains against these baselines and show that our methods improve fairness in ML models, while maintaining utility on par with that before interventions. 
    (Section~\ref{sec:exp})
\end{itemize}

%% file: figs/tab-scc-novelty.tex
\begin{figure}
    \setlength\tabcolsep{0.9pt}
\resizebox{\columnwidth}{!}{
    \begin{tabular}{lcccccc}
    \toprule
    & \makecell[c]{\dro \\ \cite{hashimoto2018fairness}} & \makecell[c]{LAH \\ \cite{lahoti2020fairness}} & \makecell[c]{CAP\\ \cite{SalimiRHS2019}} & \makecell[c]{KAM \\ \cite{DBLP:journals/kais/KamiranC11}} & \makecell[c]{OMN \\ \cite{zhang2021omnifair}} & \scc \\ 
    \midrule
        \rowcolor{lightgray}
    non-invasive wrt data & $\checkmark$ & $\checkmark$  & $\times$   & $\checkmark$    & $\checkmark$     & $\checkmark$  \\ 
    non-invasive wrt model & $\times$ & $\times$  & $\checkmark$   & $\checkmark$    & $\checkmark$     & $\checkmark$  \\ 
        \rowcolor{lightgray}
    flexible intervention & $\times$ & $\times$  & $\times$   & $\times$    & $\checkmark$     & $\checkmark$  \\ 
    intra-group variability & $\checkmark$ & $\checkmark$  & $\times$   & $\times$    & $\times$     & $\checkmark$  \\ 
    \bottomrule
    
    \end{tabular}}
    \vspace{-3pt}
    \caption{\scc provides non-invasive and flexible interventions; by allowing for variable weights within the same group, \scc can better balance the tradeoff between fairness and accuracy.
    }
    \vspace{-1em}
    \label{tab:scc_comparison}
\end{figure}

%% file: ps.tex

\section{Framing fairness as data drift}
\label{sec:ps}

In this section, we formalize our notation and problem, we then provide a high-level description of our model-splitting and reweighing strategies, and, finally, we review some necessary background on Conformance Constraints (CCs), a recently-proposed profiling primitive that we use as an off-the-shelf tool in our methods.

\subsection{Notations and problem statement}

We first discuss the notations used in the paper. We denote
variables with upper-case letters, \eg $X$ and $Y$; values with lower-case letters, \eg $n, m, c, i$, and $j$; sets of variables or values with boldface symbols, \eg $\mathbf{X}$ or $\mathbf{t}$; and bags of variables with calligraphic symbols, \eg $\mathcal{D}$.
Figure~\ref{tab:notations} summarizes the main notations that appear in the paper.

\smallskip
\noindent\textbf{Data.} We assume input data $\mathcal{D}$ that consists of $n=|\mathcal{D}|$ tuples. Each tuple is described by a set of attributes $\mathbf{X}$ with cardinality $m=|\mathbf{X}|$ and a target attribute $Y$ with $c$ distinct classes (or labels). 

\smallskip
\noindent\textbf{Groups.} 
For ease of exposition, and without loss of generality, we assume that $\mathcal{D}$ can be partitioned into a majority group $\mathbf{W}$ and a minority group $\mathbf{U}$.\footnote{Our approach can be easily extended to the general case, where the input data contains multiple majority and minority groups.} 
For the purposes of our work, we use the term minority to refer to a group $\mathbf{U}$ that is under-represented in the data, either with respect to the overall population, \ie $|\mathbf{U}|$ is small, or with respect to the target attribute $Y$ within $\mathbf{U}$, \ie there exists $i\in[1,c]$, with $\mathbf{U}_i=\{\mathbf{t}| \mathbf{t}\in\mathbf{U}\wedge \mathbf{t}.Y=i\}$, such that $|\mathbf{U}_i|$ is small.
We further assume a user-specified binary mapping function $g:
\mathbb{R}^{n\times m} \mapsto [0,1]$ that takes as input a tuple $\mathbf{t}$ and maps it to $\mathbf{W}$ or $\mathbf{U}$. 
Typically, $g$ is a simple function over one or more attributes in $\mathbf{X}$. For example, based on the color of the data points in Fig.~\ref{fig:example}, a tuple can be assigned to the
``blue'' majority group or the ``orange'' minority group.

\smallskip
\noindent\textbf{Model.} 
We assume a model $f: \mathbb{R}^{n\times m} \mapsto \mathbb{R}^{n\times c}$, which takes as input a tuple $\mathbf{t} \in \mathcal{D}$ and outputs a prediction as one of the $c$ classes of the target attribute $Y$. We denote the predictions of $f$ on $\mathcal{D}$ by $\hat{Y}$.
We use the following standard process to develop a model $f$. We partition the input $\mathcal{D}$ into three disjoint sets: training $\mathcal{D}^t$, validation $\mathcal{D}^v$, and deploy
$\mathcal{D}^d$. 
We train $f$ on $\mathcal{D}^t$, optimize for its hyperparameters on $\mathcal{D}^v$, and deploy and evaluate it on $\mathcal{D}^d$.
Tuples are assigned into these three sets independently at random (i.i.d.).

\smallskip
\noindent\textbf{Metrics.} 
A fairness metric $\Delta(\mathbf{W}, \mathbf{U})$ quantifies the difference in predictions $\hat{Y}$ between the majority $\mathbf{W}$ and minority $\mathbf{U}$. 
A lower value of $\Delta(\mathbf{W}, \mathbf{U})$ indicates less bias in the predictions of $f$. A utility function $\Sigma(Y, \hat{Y})$
quantifies the similarity between the target attribute $Y$ and the output $\hat{Y}$ of $f$. 
A higher value of $\Sigma(Y, \hat{Y})$ indicates higher utility for the model $f$.

\begin{definition}[Non-invasive fair learning]
    Given a dataset $\mathcal{D}$, a mapping function $g$, and a learning algorithm $f$, our goal is to design a learning framework that, without altering the data in $\mathcal{D}$ or the learner $f$, trains a model $f'$ using learner $f$,
    such that the fairness difference $\Delta(\mathbf{W}, \mathbf{U})$ is minimized, while the utility $\Sigma(Y, \hat{Y})$ is maximized.
\end{definition}

\input{figs/tab-notations}

%% file: figs/tab-notations.tex
\begin{figure}[t]
    \setlength\tabcolsep{1.8pt}

	\begin{tabular}{p{.15\linewidth}p{.84\linewidth}l}
        \toprule
	    $\mathcal{D}$  & 
        input dataset with $n$ tuples\\
        
		$\mathbf{X}$ & 
        attributes in $\mathcal{D}$ with $|\mathbf{X}|=m$  \\
        
		$Y$  & 
        the target attribute in $\mathcal{D}$ with $c$ distinct values \\
        
		$\mathbf{t}$ & 
        a tuple in $\mathcal{D}$ with attributes $\mathbf{X}\cup Y$\\
        
		$\mathbf{W}$ & 
        a majority group (well-represented) in $\mathcal{D}$ \\ 
        
		$\mathbf{U}$ & 
        a minority group (underrepresented) in $\mathcal{D}$\\ 
        
		$g(\mathbf{t})$ & 
        a binary mapping function that assigns $\mathbf{t}$ to $\mathbf{W}$ or $\mathbf{U}$\\ 
        
		$f$, $\hat{Y}$  & a classification model and its output on $\mathcal{D}$\\ 
        
        
		$\Sigma(Y, \hat{Y})$ & 
        a utility function that quantifies the similarity between target attribute $Y$ and model output $\hat{Y}$ \\ 
        
		$\Delta(\mathbf{W}, \mathbf{U})$ & 
        a fairness metric that quantifies the difference between $\mathbf{W}$ and $\mathbf{U}$ \\ 
        
        $\mathbf{\Phi}^w$, $\mathbf{\Phi}^u$ &
        conformance constraints over $\mathbf{W}$ and $\mathbf{U}$, respectively\\
		\bottomrule
	\end{tabular}
    \vspace{-0.6em}
    \caption{Summary of notations used in the paper.}
    \label{tab:notations}
    \vspace{-1em}
\end{figure}

%% file: overview.tex

\subsection{Strategy overview: improving conformance}
\label{sec:overview_own}
We described how data drift (across groups) leads to unfairness in ML models. As a result, the produced model may not conform to the minority group, whose predictions are, thus, not reliable. To improve the conformance between the model and data, we propose two strategies: a model-splitting approach (\mcc) and a reweighing approach (\scc). 

\smallskip
\noindent
\textbf{\mcc} follows a simple strategy: train separate models for different groups and deploy these group-dependent models collectively to improve the conformance between the model and data.
A naive version of this strategy, which we will simply refer to as \mbase, may split the input data based on group membership (\eg blue and orange points in Fig.~\ref{fig:example}), train multiple models (one for each group), and choose a model to use for a serving tuple based on its group membership during deployment.  In contrast to the naive \mbase method, \mcc does not use group membership in assigning models for serving tuples. Instead, it learns constraints to describe each group's training data using CCs.  
For each serving tuple, \mcc chooses the model that minimizes the tuple's violation score against the CCs of the model's training data. 

This strategy has two important advantages compared to simply relying on group membership: 

\begin{enumerate}[leftmargin=12pt]
    
\item \mcc affords compliance with legal considerations regarding discrimination when it does not rely on group membership during deployment. Such membership information can be sensitive and protected when it corresponds to demographic attributes (\eg gender, race, disability status, etc.). Additionally, \mcc is robust to erroneous membership during deployment, \ie individuals with wrong membership information (\eg auto-filled or misclassified) still receive correct predictions.

\item \mcc handles individuality more flexibly compared to \mbase, which deploys a model strictly based on the group membership of a serving tuple. In contrast, \mcc chooses a model for the tuple considering the distribution of its attributes, \ie assigning a model to which a tuple conforms better, regardless of which group the tuple formally belongs to.
\end{enumerate}

The novelty of \mcc lies in its use of CCs to model drift and separate data based on this drift. \mcc builds a simple mechanism around this intuition: it serves each tuple using the model that results in the minimum CC violation score. One can easily augment this with more sophisticated mechanisms (\eg ensemble learning), where conformance constraints can be used as an explicit heuristic for aggregating predictions from involved models.  

\noindent \textbf{\scc} aims to achieve better conformance between the model and data through a reweighing strategy. It assigns weights to tuples in the training data, and an ML model then takes the new weighted data as input. \scc determines these weights based on the conformance constraints that are learned over each group's data. It increases the weights of the tuples that best conform to the produced constraints (\eg points located inside the green rectangles in Fig.~\ref{fig:example}). 
Training a single model makes \scc more robust against the poor representation of groups, whereas model-splitting approaches, such as \mbase and \mcc, are limited by the need for adequate group representation to train a reasonable model. 

\subsection{Background on Conformance Constraints}
\label{sec:overview_cc}
We proceed with a brief overview of conformance constraints~\cite{DBLP:conf/sigmod/Fariha0RGM21}. 
We generally follow the formalism and notations of the original paper, but we omit or simplify some details in the summary we provide here; we refer the reader to Fariha et al.~\cite{DBLP:conf/sigmod/Fariha0RGM21} for more detail.

A conformance constraint is a constraint over arithmetic relationships
involving multiple numerical attributes. 
More formally, a constraint $\phi$ is an expression of the form $\phi \coloneqq \epsilon^{lb} \leq F(\mathbf{X}) \leq \epsilon^{ub}$, where $\epsilon^{lb}$ and $\epsilon^{ub}$ are the lower and upper bounds of the projection $F(\mathbf{X})$. $F(\mathbf{X})$ is a linear combination of numerical attributes $\mathbf{X}$ in data $\mathcal{D}$. 
We use $\mathbf{\Phi}$ to denote a set of conjunctive constraints. For a tuple $\mathbf{t}$, $\mathbf{\Phi}(\mathbf{t})$ is computed as follows:
$\mathbf{\Phi}(\mathbf{t}) \coloneqq \phi_1(\mathbf{t}) \wedge \phi_2(\mathbf{t}) \dots \wedge \phi_r(\mathbf{t})$, and $\phi_i(\mathbf{t}) \coloneqq \epsilon_i^{lb} \leq F_i(\mathbf{t}) \leq \epsilon_i^{ub}, \forall i \in [1, r]$. $F_i(\mathbf{t})$ is simplified from $F_i(\mathbf{t}.\mathbf{X})$ for readability. In this Boolean semantics, a tuple $\mathbf{t}$ satisfies the constraints $\mathbf{\Phi}$ when $\mathbf{\Phi}(\mathbf{t})=1$. Otherwise, $\mathbf{t}$ violates the constraints $\mathbf{\Phi}$.

Fariha et al.~\cite{DBLP:conf/sigmod/Fariha0RGM21} also propose quantitative semantics to measure the level of violation of a tuple $\mathbf{t}$ for constraints $\mathbf{\Phi}$, denoted as $\llbracket \mathbf{\Phi}\rrbracket(\mathbf{t})$. We compute the violation $\llbracket \mathbf{\Phi}\rrbracket(\mathbf{t})$ as follows:
\begin{equation}
\begin{split}
    \llbracket \mathbf{\Phi}\rrbracket(\mathbf{t}) &= \sum_{i=1}^r q_i \cdot \llbracket \phi_i \rrbracket (\mathbf{t}) \\
    \llbracket \phi_i \rrbracket (\mathbf{t}) & = \eta (\frac{dist(F_i, \mathbf{t})}{\sigma(F_i(\mathbf{t}))}), \forall i \in [1, r] \\
    dist(F_i, \mathbf{t}) &= max(0, F_i(\mathbf{t})-\epsilon_i^{ub}, \epsilon_i^{lb} - F_i(\mathbf{t}))\\
    \eta(x)&=1-e^{-x}
\end{split}
\label{eq:cc}
\end{equation}
Where $q_i \in \mathbb{R}^+, \forall i \in [1,r]$ is the coefficient of the
expression $\phi_i \in \mathbf{\Phi}$ and $\sum_{i=1}^r q_i=1$. 
This factor represents the importance of the expression $\phi_i$ and is computed as $q_i=1-\frac{\sigma(F_i)}{max(\sigma(\mathbf{F}))-min(\sigma(\mathbf{F}))}$, where $\mathbf{F}=\{F_1, \dots, F_r\}$ consists of all the projections involved in expressions $\mathbf{\Phi}$. 
It is saying that the lower the standard deviation $\sigma(F_i)$ of the projection $F_i$ is, the more important the expression $\phi_i$ is in computing the violation of the tuple $\mathbf{t}$. 
In other words, the set of constraints, whose projections have low standard deviations, is more effective at characterizing tuples in $\mathcal{D}$.

In this quantitative semantic, a tuple $\mathbf{t}$ satisfies the constraints $\mathbf{\Phi}$ (i.e., $\mathbf{\Phi}(\mathbf{t})=1$) when the violation $\llbracket \mathbf{\Phi}\rrbracket(\mathbf{t})=0$. 
Otherwise, the lower the violation $\llbracket \mathbf{\Phi}\rrbracket(\mathbf{t})$ is, the more $\mathbf{t}$ conforms to $\mathbf{\Phi}$. 
We employ these quantitative semantics in our approach to profile groups' data. In this paper, we use $\mathbf{\Phi}^w$ and $\mathbf{\Phi}^u$
to denote the sets of constraints derived over the majority and minority groups $\mathbf{W}$ and $\mathbf{U}$, respectively. 

\begin{example}
Considering the dataset in Fig.~\ref{fig:example}, an example of the produced constraint for the majority group $\mathbf{W}$ with the positive label (blue circles) is: 
$$\phi^w: 0.708 \leq 0.477 \cdot X1 + 0.265 \cdot X2 \leq 0.902$$

And an example constraint for the minority group $\mathbf{U}$ with positive label (orange circles) is:
$$\phi^u: 0.771 \leq -0.519 \cdot X1 - 0.16 \cdot X2 \leq 0.912 $$

Where the two involved projections of attributes are $F^w(\mathbf{t})=0.477 \cdot X1 + 0.265 \cdot X2 $ with coefficients $\langle 0.477, 0.265 \rangle$, $\epsilon_{w}^{ub}=0.902$ and $\epsilon_{w}^{lb}=0.708$, and $F^u(\mathbf{t})=-0.519 \cdot X1 - 0.16 \cdot X2$ with coefficients $\langle -0.519, -0.16 \rangle $, $\epsilon_{u}^{ub}=0.912$ and $\epsilon_{u}^{lb}=0.771$ for groups $\mathbf{W}$ and $\mathbf{U}$, respectively. 
When the above constraints are applied over the red-outlined points inside the green rectangle in \reffig{fig:example}, their average violations for the constraints $\phi^w$ and $\phi^u$ are:
$$\llbracket \phi^w \rrbracket=0.3, \llbracket \phi^u \rrbracket=0$$

\label{exp:cc}
\end{example}

%% file: approach.tex
\section{Fairness through Conformance}
\label{sec:method}
In this section, we present two methods that aim to improve fairness in learning, by improving the conformance of models to underlying data.  We first describe \mcc, which enhances the naive method \mbase by using conformance constraints to deploy the appropriate model for serving tuples (Section~\ref{sec:method_driftcc}).  
Next, we introduce \scc, which uses conformance constraints to assign weights to the training data and then build a single model over weighted data (Section~\ref{sec:method_faircc}). 
Finally, we present an optimization that improves the effectiveness of the derived CCs: we use density estimation to preprocess the input and filter high-variance data, leading to tighter constraints.

\subsection{\mcc}
\label{sec:method_driftcc}

Our model-splitting approach is designed around a simple data drift strategy: train separate models for different groups, such that each produced model better conforms to its underlying data. 
\mcc augments this simple strategy with conformance constraints: Roughly, \mcc derives CCs from the training data of each model, and calculates the violation of each serving tuple against each set of constraints; it then selects the model that corresponds to the lowest violation to serve the tuple.

\begin{algorithm}[t]
	\caption{\mcc}
	\begin{algorithmic}[1]
		\Require Dataset $\mathcal{D}$ with attributes $\mathbf{X}$, a target attribute $Y$, and a mapping function $g$. 
		\Ensure A fair model $f'$ over $\mathcal{D}$
		\State $Partition(\mathcal{D})\rightarrow\{\mathcal{D}^t, \mathcal{D}^v, \mathcal{D}^d\}$  \label{ln:part}
		\LineComment {Identify majority and minority groups $\mathbf{W}$ and $\mathbf{U}$ in $\mathcal{D}^t$ and $\mathcal{D}^v$}
		\State $\mathbf{W}^t=\{\mathbf{t}| g(\mathbf{t})=0, \mathbf{t} \in \mathcal{D}^t\}$, $\mathbf{U}^t=\{\mathbf{t}| g(\mathbf{t})=1, \mathbf{t} \in \mathcal{D}^t\}$\label{ln:wt}
		\State $\mathbf{W}^v=\{\mathbf{t}| g(\mathbf{t})=0, \mathbf{t} \in \mathcal{D}^v\}$, $\mathbf{U}^v=\{\mathbf{t}| g(\mathbf{t})=1, \mathbf{t} \in \mathcal{D}^v\}$\label{ln:wv}
		\State $\mathbf{C}^w = \emptyset$, $\mathbf{C}^u = \emptyset$ \label{ln:CC}
		\For {$i \leftarrow 1, \dots c$}
		\State $\mathbf{W}^t_i = \{\mathbf{t}| \mathbf{t}.Y=i, \mathbf{t} \in \mathbf{W}^t\}$; $\mathbf{U}^t_i = \{\mathbf{t}| \mathbf{t}.Y=i, \mathbf{t} \in \mathbf{U}^t\}$
		\State $\mathbf{\Phi}^w_i = GetCCs(\mathbf{W}^t_i)$;  $\mathbf{\Phi}^u_i= GetCCs(\mathbf{U}^t_i)$
		\State $\mathbf{C}^w \leftarrow \mathbf{C}^w \cup \mathbf{\Phi}^w_i$, $\mathbf{C}^u \leftarrow \mathbf{C}^u \cup \mathbf{\Phi}^u_i$
		\EndFor \label{ln:CCdone}
		\State Train $f^w$ on $\mathbf{W}^t$; Train $f^u$ on $\mathbf{U}^t$; 
        \label{ln:train}
		\State Validate $f^w$ on $\mathbf{W}^v$; Validate $f^u$ on $\mathbf{U}^v$;\label{ln:val}
        \For {$\mathbf{t} \in \mathcal{D}^d$}
        \LineComment{Produce predictions for all serving tuples}
            \State \textsc{Predict}($t$, $\mathbf{C}^w$, $\mathbf{C}^u$)
        \EndFor
        \State\Return $f' \leftarrow (f^w, f^u, \mathbf{C}^w, \mathbf{C}^u)$
        \Statex
        \Procedure{Predict}{$t$, $\mathbf{C}^w$, $\mathbf{C}^u$}\label{ln:pred}
    		\State $v^w(\mathbf{t}) = min_{\mathbf{\Phi}^w \in \mathbf{C}^w} \llbracket \mathbf{\Phi}^w \rrbracket (\mathbf{t})$ \label{ln:viol1}
            \State $v^u(\mathbf{t})= min_{\mathbf{\Phi}^u \in \mathbf{C}^u} \llbracket \mathbf{\Phi}^u \rrbracket (\mathbf{t})$ \label{ln:viol2}
    		\If {$v^w(\mathbf{t}) < v^u(\mathbf{t})$}
                \label{ln:viol3}
    		\State \Return $f^w(t)$ 
    		\Else
    		\State \Return $f^u(t)$
    		\EndIf
                \label{ln:viol4}
        \EndProcedure \label{ln:predend}
	\end{algorithmic}
	\label{alg:driftcc}
\end{algorithm}
Algorithm~\ref{alg:driftcc} presents this strategy in more detail. 
The algorithm takes as input a dataset $\mathcal{D}$ and a mapping function $g$ to define groups. 
While the pseudo-code assumes a binary function $g$, it can easily generalize to more than two groups. 
The algorithm first splits $\mathcal{D}$ into three disjoint sets: training $\mathcal{D}^t$, validation $\mathcal{D}^v$, and deployment $\mathcal{D}^d$ (line~\ref{ln:part}). 
Then it proceeds to identify groups within the first two sets using the mapping function $g$ (lines~\ref{ln:wt}--\ref{ln:wv}). 
The algorithm proceeds to derive constraints for groups within the training set $\mathcal{D}^t$, and \mcc does so within each set of labels (lines~\ref{ln:CC}--\ref{ln:CCdone}). 
This is because, in practice, individuals with positive and negative labels may display distinct patterns in their attributes (\eg triangles and circles in Fig.~\ref{fig:example}).
\mcc, therefore, leads to a tighter and higher-quality set of constraints.

\mcc proceeds to train two group-dependent models $f^w$ and $f^u$ for the majority and minority, respectively, and optimizes their parameters over the corresponding validation sets (lines~\ref{ln:train}--\ref{ln:val}). 
The \textsc{Predict} procedure (lines~\ref{ln:pred}--\ref{ln:predend}) outputs predictions for each serving tuple $\mathbf{t}$ in $\mathcal{D}^d$ solely by the constraints $\mathbf{C}^w$ and $\mathbf{C}^u$ without referring to the mapping function $g$. The goal of \textsc{Predict} is to identify the best model to deploy.  
First, we determine the label group within the majority and the label group within the minority that tuple $t$ is closest to (has minimal violation (lines~\ref{ln:viol1}--\ref{ln:viol2}).  Then, comparing the majority and minority violation scores, we select the appropriate model (lines~\ref{ln:viol3}--\ref{ln:viol4}).  Note that \mcc picks the model with the best conformance, even if a tuple does not actually belong to that group. By prioritizing conformance, \mcc achieves better accuracy, especially for members of the minority group, leading to improved model fairness.

The run-time complexity of Algorithm~\ref{alg:driftcc} is bounded by the derivation of conformance constraints, which takes $O(q^3)$ with $q$ numerical attributes for computing projections and $O(nm^2)$ with $n$ tuples and $m$ attributes for producing the constraints~\cite{DBLP:conf/sigmod/Fariha0RGM21}. 
The training of models takes $O(nm)$ with $n$ tuples and $m$ attributes using classification algorithms such as Logistic Regression.

\subsection{\scc}
\label{sec:method_faircc}
Multi-model approaches are more susceptible to groups' poor representation in the input. Splitting the dataset to produce multiple models weakens the predictive power over the input with small and often skewed group representation. 
We explore a single-model strategy that boosts the conformance between the model and data (especially of minority groups), without diluting the learning power as in multi-model approaches. 

\begin{algorithm}[t]
	\caption{\scc}
	\begin{algorithmic}[1]
		\Require Dataset $\mathcal{D}$ with attributes $\mathbf{X}$, a target attribute $Y$, a mapping function $g$, and intervention factors $\alpha^w$ and $\alpha^u$ for the majority and minority groups, respectively.
		\Ensure Dataset $\mathcal{D}$, augmented with a weight attribute.
		\State $\mathbf{t}.S=0$, $\forall \mathbf{t} \in \mathcal{D}$  \label{ln:init2}
        \Comment{add weight attribute $S$ with initial value 1}
		\ForAll {$c$}\label{ln:part1}		\Comment {partition $\mathcal{D}$ according to target $Y$ and function $g$}
		\State $\mathbf{W}_c=\{\mathbf{t}| g(\mathbf{t}){=}0, \mathbf{t}.Y{=}c, \mathbf{t} {\in} \mathcal{D}\}$, $\mathbf{U}_c=\{\mathbf{t}| g(\mathbf{t}){=}1, \mathbf{t}.Y{=}c, \mathbf{t} {\in} \mathcal{D}\}$
        \State $\mathbf{\Phi}^w_c = GetCCs(\mathbf{W}_c)$;  $\mathbf{\Phi}^u_c= GetCCs(\mathbf{U}_c)$ \label{ln:getcc} \Comment{derive constraints} 
            \LineComment{Update weights for population and label skew}
		\State $\mathbf{t}.S \leftarrow \mathbf{t}.S + \frac{|\{\mathbf{t}| \mathbf{t}.Y{=}c, \mathbf{t} {\in} \mathcal{D}\}|}{|\mathcal{D}|} *( \frac{g(\mathbf{t})*|\mathbf{U}|}{|\mathbf{U}_c|} + \frac{(1- g(\mathbf{t})) *|\mathbf{W}|}{|\mathbf{W}_c|})$ \label{ln:weightpop}
		
        \LineComment{Find conforming tuples}
		\State $\mathbf{T}_c^w =\{\mathbf{t}| \mathbf{t}\in\mathbf{W}_c, \llbracket \mathbf{\Phi}_c^w \rrbracket (\mathbf{t}) ==0\}$
        \label{ln:cc1}
        \State
        $\mathbf{T}_c^u =\{\mathbf{t}| \mathbf{t}\in\mathbf{U}_c,\llbracket \mathbf{\Phi}_c^u \rrbracket (\mathbf{t}) ==0\}$
        \label{ln:cc2}
		\EndFor\label{ln:part2}
        \LineComment{Increase the weight of tuples conforming with minority positive labels}
		\For {$\mathbf{t} \in \mathbf{T}_1^u$}\label{ln:weight1}
		\State $\mathbf{t}.S \leftarrow \mathbf{t}.S + \alpha^u$ 
		\EndFor
        \LineComment{Increase the weight of tuples conforming with majority negative labels}
		\For {$\mathbf{t} \in \mathbf{T}_0^w$}
		\State $\mathbf{t}.S \leftarrow \mathbf{t}.S + \alpha^w$ 
		\EndFor\label{ln:weight2}
		\Return $\mathcal{D}$
	\end{algorithmic}
	\label{alg:faircc}
\end{algorithm}
In this section, we present \scc in Algorithm~\ref{alg:faircc}, a single-model approach that uses CCs in a novel way to derive weights for the training data. 
For ease of exposition, the pseudo-code assumes binary labels (\ie $c=2$). It further assumes that the positive labels are over-represented in the majority group, while the opposite holds for the minority. 
These assumptions are simply for ease of presentation, and not true restrictions of the framework.
The intervention degrees $\alpha^w$ and $\alpha^u$ are adjustable weight parameters to control the level of intervention that users may wish to apply to the majority and minority groups, respectively. \rev{By default, \scc optimizes for disparate impact by applying these weights to appropriate labels.} \revs{\scc can easily adapt to other fairness measures by appropriately adjusting $\alpha^w$ and $\alpha^u$. For example, to optimize \eeo by FNR, we can set $\alpha^u$ to a positive value (e.g., $\alpha^u=2$) and $\alpha^w$ to zero; \scc would then only increase the weights of tuples within the minority group associated with positive labels, thus decreasing the FNR.} 
\rev{\scc only augments the weights of tuples that conform to the identified CCs, resulting in a monotonic behavior of improvement in fairness with respect to the intervention degree; this facilitates tuning the parameter to each application's fairness requirements. In contrast, prior art~\cite{zhang2021omnifair} augments the weights of all tuples in a group; as data is inevitably noisy, this lead to a non-monotonic relationship between the level of intervention and the achieved fairness (see evaluation in Section~\ref{sec:exp:scc}).} 

\scc adds a weight attribute $S$ to input $\mathbf{D}$ initialized in line~\ref{ln:init2}. 
It then partitions $\mathbf{D}$ based on a mapping function $g$ and target attribute $Y$. For example, the dataset in Fig.~\ref{fig:example} would be separated into four parts: the majority group with positive labels (blue circles), the majority group with negative labels (blue triangles), the minority group with positive labels (orange circles), and the minority group with negative labels (orange triangles). \scc proceeds to derive constraints on each part (line~\ref{ln:getcc}). 
It then balances the weights of tuples in each part according to the skew in the groups' population and labels, \ie increase weights for the minority and decrease values for the majority (line~\ref{ln:weightpop}).
Next, \scc focuses on the tuples that conform to each part (lines~\ref{ln:cc1} and~\ref{ln:cc2}). Based on the intervention factors $\alpha^w$ and $\alpha^u$, \scc adjusts the weights of these conforming tuples (lines~\ref{ln:weight1}--\ref{ln:weight2}). Recall that the presented pseudo-code makes the assumption that the majority part of the data skews toward positive labels and that the minority part of the data skews toward negative labels.
Thus, to achieve balanced predictive accuracy across groups, \scc increases the weights of majority-negative-conforming tuples by $\alpha^w$, and the weights of minority-positive-conforming tuples by $\alpha^u$. 
Note that this assumption is made here for readability. The approach can easily generalize to the labels with multiple classes. And the skew of groups toward the labels can be easily estimated from the data, which can guide the tuning of the intervention factors (\eg increase weights for the minority group with positive labels or vice-versa). Finally, \scc returns the weight-augmented data to build an ML model. 

Similar to Algorithm~\ref{alg:driftcc}, the run-time complexity of Algorithm~\ref{alg:faircc} is also bounded by the derivation of conformance constraints, which takes $O(q^3)$ with $q$ numerical attributes for computing projections and $O(nm^2)$ with $n$ tuples and $m$ attributes for deriving the constraints~\cite{DBLP:conf/sigmod/Fariha0RGM21}. 

\subsection{Optimizing the derivation of CCs}
\label{sec:method_density}
The effectiveness of conformance constraints is affected by the variance of attributes in the input. 
A set of constraints learned from data with high variance has low discriminative power: most tuples will have high conformance with broad, permissive constraints.
Such weak constraints can critically impact the effectiveness of our methods. 
In this section, we propose a pre-processing optimization step that filters the input data $\mathcal{D}$ using density estimation, leading to stronger sets of conformance constraints, and by implication, increased effectiveness for \mcc and \scc.

We present our optimization in Algorithm~\ref{alg:density}. The algorithm processes each target class separately (lines~\ref{ln:class1}--\ref{ln:class2}), and uses density estimation on the majority and minority sets within the target class (lines~\ref{ln:dense1}--\ref{ln:dense2}). 
In our implementation, we employ a state-of-art, tree-based, non-parametric kernel density estimator~\cite{DBLP:journals/cacm/Bentley75}, implemented in the scikit-learn library~\cite{scikit-learn}.
Other kernel density estimators can also work for this step~\cite{wasserman2006all,silverman2018density,muller2004nonparametric}.
Algorithm~\ref{alg:density} proceeds to sort the sets $\mathbf{W}_i$ and $\mathbf{U}_i$ based on the density functions and select the first $k$ tuples to add to $\mathcal{D}'$ (lines~\ref{ln:sort1}--\ref{ln:sort2}).

\begin{algorithm}[t]
	\caption{Optimization for stronger conformance constraints}
	\begin{algorithmic}[1]
		\Require Dataset $\mathcal{D}$ with attributes $\mathbf{X}$, a target attribute $Y$, a mapping function $g$, and a density threshold $k$.
		\Ensure Dataset $\mathcal{D}'\subset\mathcal{D}$
		\State $\mathcal{D}' = \emptyset$
		\For {$i \leftarrow 1, \dots c$} \Comment{process each class in target attribute $Y$.}
            \label{ln:class1}
		\State $\mathbf{W}_i = \{\mathbf{t}| \mathbf{t}.Y=i, g(\mathbf{t})=0, \mathbf{t} \in \mathbf{D}\}$
        \State $\mathbf{U}_i = \{\mathbf{t}| \mathbf{t}.Y=i, g(\mathbf{t})=1, \mathbf{t} \in \mathbf{D}\}$
           \label{ln:class2}
		\State $d^w\leftarrow EstimateDensity(\mathbf{W}_i)$ \label{ln:dense1}
		\State $d^u\leftarrow EstimateDensity(\mathbf{U}_i)$ \label{ln:dense2}
		\State Sort $\mathbf{W}_i$ in the descending order of $d^w$ \label{ln:sort1}
		\State Sort $\mathbf{U}_i$ in the descending order of $d^u$ 
		\State $\mathcal{D}' \leftarrow $ $\mathcal{D}' \cup\{$ first $k$ tuples in $\mathbf{W}_i\}$
		\State $\mathcal{D}' \leftarrow $ $\mathcal{D}' \cup\{$ first $k$ tuples in $\mathbf{U}_i\}$\label{ln:sort2}
		\EndFor
		\Return $\mathcal{D}'$
	\end{algorithmic}
	\label{alg:density}
\end{algorithm}

The run-time complexity of Algorithm~\ref{alg:density} is bounded by the density estimation, which takes $O(mn^2)$ with $n$ tuples and $m$ attributes. 
This run-time can be improved to $O(m\log(n))$ using optimized data structures such as KD-Tree~\cite{DBLP:journals/cacm/Bentley75} or Ball Tree~\cite{omohundro1989five} for input data in higher dimensions (\eg $m > 20$).

%% file: exp_supp.tex

\section{Experimental Evaluation}
\label{sec:exp}

We evaluate \mcc and \scc against a breadth of datasets and methods.
Our experiments demonstrate that: (1)~\scc outperforms prior art in improving the fairness of models, while maintaining high accuracy (Section~\ref{sec:exp:scc});
(2)~\mcc can be a better solution compared to \scc for scenarios, where it is difficult to derive a single conforming model (Section~\ref{sec:exp:mcc});
(3) our optimization for deriving stronger conformance constraints is essential (Section~\ref{sec:exp:opt});
(4) our approach shows a reasonable run-time compared to prior art (Section~\ref{sec:exp:time}). 
We describe each experimental component below.

\smallskip
\noindent\textbf{Datasets.} 
We experiment with 7 real-world datasets that include people's demographics and information collected from various domains such as financial and health-related services. These datasets have been frequently used in fairness literature. 
We provide a summary description of the major aspects and statistics of each dataset in Figure~\ref{tab:datasets}. For more detail, we refer the reader to Bellamy \etal~\cite{aif360-oct-2018} for the \dmp and \dlg datasets, Kaggle~\cite{kaggle_give_me_some_credit} for the \dcr dataset, and Ding \etal~\cite{ding2021retiring} for the American Community Survey (ACS) datasets. We employ four predictive tasks using the ACS datasets, which pertain to people's health insurance (\dap and \dah), employment (\dae), and income (\dai). 
We choose to skip other frequently-used datasets such as Adult Income~\cite{adult} and COMPAS~\cite{barocas2016big} because these datasets include very few numerical attributes (\eg no more than 2) for deriving conformance constraints.

\input{figs/tab-datasets}

\smallskip
\noindent\textbf{Models.} 
As our methods do not depend on nor intervene with the learners, one could apply them in combination with any learning algorithm.
We experiment with two types of learners: Logistic Regression (LR) and XGBoost tree (XGB) from the scikit-learn library~\cite{scikit-learn}.

\smallskip
\noindent\textbf{Methods.}
We briefly describe each approach in our evaluation.

\noindent
$\triangleright$ \base.
This first baseline trains a model over the input without applying any fairness intervention. The goal of all other methods is to achieve improvement in the fairness metrics against this baseline, while maintaining comparable utility.

\noindent
$\triangleright$ \mbase is a simple model-splitting baseline. It partitions the input data into groups by a mapping function $g$, builds separate models for different groups, and deploys these group-dependent models based on the function $g$.

\noindent
$\triangleright$ \mcc (Section~\ref{sec:method_driftcc}) augments \mbase, by using conformance constraints in the model deployment, rather than rely on group membership (or the mapping function $g$), and is thus more flexible and robust to inaccuracies in membership data.

\noindent
$\triangleright$ \scc (Section~\ref{sec:method_faircc}) is a single-model reweighing strategy, which derives weights for training tuples, based on their conformance to the CCs in each subgroup.

\noindent
$\triangleright$ \textsc{KAM-CAL} (\kam)~\cite{DBLP:journals/kais/KamiranC11} is a reweighing method (like \scc). It assigns weights to achieve statistical independence between the demographic attributes (defining the groups) and labels. Relying on groups' statistics in the input, \kam does not support adjusting the level of interventions.

\noindent
$\triangleright$ \textsc{OmniFair} (\omn)~\cite{zhang2021omnifair} is another reweighing method that assigns weights to achieve fairness with respect to a given metric. 
We use a variant of \omn that optimizes for Disparate Impact, as this is the metric that \scc inherently optimizes as well (discussed in more detail in the evaluation metrics below).

\noindent
$\triangleright$ \textsc{CAPUCHIN} (\capu)~\cite{SalimiRHS2019} is an \emph{invasive} fairness intervention that modifies the input data to ensure that certain constraints hold over its outputs. Since \capu is designed for categorical data, we evaluate this method using the XGB models, which are a better fit for categorical input.

\smallskip
\noindent\textbf{Metrics.} 
We evaluate all methods with respect to the fairness and utility of the produced models. We use balanced accuracy (\ebacc) as the utility metric, which has been used extensively in literature to evaluate fairness interventions~\cite{aif360-oct-2018}. It is computed as $\frac{TPR + TNR}{2}$, where TPR and TNR are the True Positive Rate (or sensitivity) and True Negative Rate (or specificity), respectively. \ebacc is similar to the Area Under the ROC curve (AUC); both utility metrics are sensitive to the poor representation of minority groups. \ebacc is in the range of $[0,1]$, with higher values corresponding to greater utility.

We report two fairness metrics that are frequently used in the literature: Disparate Impact (\edi)~\cite{feldman2015certifying} and Average Odds Difference (\ead)~\cite{DBLP:conf/nips/HardtPNS16}. 
In our implementation, \scc targets \edi, but our results show that it performs well for both metrics.
\edi is computed as $\frac{SR_{\mathbf{U}}}{SR_{\mathbf{W}}}$ where  $SR_{\mathbf{U}}=\frac{|\{\mathbf{t}|\mathbf{t}.\hat{Y}=1, \mathbf{t} \in \mathbf{U}\}|}{|\mathbf{U}|}$ and $SR_{\mathbf{W}}=\frac{|\{\mathbf{t}|\mathbf{t}.\hat{Y}=1, \mathbf{t} \in \mathbf{W}\}|}{|\mathbf{W}|}$ are the selection rates for the minority $\mathbf{U}$ and majority $\mathbf{W}$, respectively. \edi takes values from 0 to $\infty$ with 1 being the optimum, \ie the two groups have the same rate in receiving a positive prediction. Values greater than 1 indicate bias favoring the minority group, which may in fact be reasonable or even desirable in some applications where minorities have suffered from historical disadvantages. 
\scc implicitly optimizes \edi, as it increases the weights of tuples with positive labels within the minority group, thus leading to an improvement of the selection rate for the minority. 

\ead is \rev{a generalized version of Equalized Odds~\cite{DBLP:conf/nips/HardtPNS16}, which is} computed as $\frac{(FPR_{\mathbf{U}}-FPR_{\mathbf{W}})+(TPR_{\mathbf{U}}-TPR_{\mathbf{W}})}{2}$ where $FPR_{\mathbf{U}}$ and $FPR_{\mathbf{W}}$ are the False Positive Rates for the minority $\mathbf{U}$ and majority $\mathbf{W}$, respectively, and $TPR_{\mathbf{U}}$ and $TPR_{\mathbf{W}}$ are the TPRs for these two groups.
\ead ranges from 0 to 1, with 0 indicating an optimal case where there is no difference in how a model makes positive predictions for the two groups. 
\ead captures a different aspect of model behavior compared to \edi; even though \scc is not designed to optimize \ead, our experiments will show that its fairness remains robust under this metric.

For ease of interpretation, we report simple transformations of these metrics, so that higher values correspond to better outcomes. 
Specifically, we report $\mathit{DI}^* = \min(\mathit{DI}, \frac{1}{\mathit{DI}})$, where unfairness ($\mathit{DI}\rightarrow 0$ or $\mathit{DI}\rightarrow \infty$) is mapped to a low value of $\mathit{DI}^*$. 
We report $\mathit{AOD}^*=1-abs($\textit{AOD}$)$ such that higher values of \ead represent improved fairness. 
For the remainder of this section, we simply use \edi and \ead to refer to $\mathit{DI}^*$ and $\mathit{AOD}^*$, respectively.

\smallskip
\noindent\textbf{Experimental steps.} 
We first prepare our data for training. For the datasets \dmp and \dlg, we use the same preprocessing steps as in the IBM AI Fairness
360 toolkit~\cite{aif360-oct-2018}. Similarly, we preprocess the other data by removing null values, normalizing numerical attributes, and one-hot encoding categorical attributes. 
We split the processed data into training (70\%), validation (15\%), and test (15\%) sets. For both multi-model and single-model settings, we tune hyperparameters on the validation set and evaluate model performance on the test set. To eliminate the effect of randomness, we repeat the previous process 20 times and report the average results in our evaluation.  

\smallskip
\noindent\textbf{Algorithm parameters.}
For the intervention factors $\alpha^w$ and $\alpha^u$ in Algorithm~\ref{alg:faircc}, we automatically search for the optimal value of $\alpha^u$ (\ie the intervention degree for the minority group) over each real-world dataset (on its validation part) and set the value of $\alpha^w$ (\ie the degree for the majority group) as $\alpha^u/2$. The tuning of intervention degrees in \scc implicitly optimizes \edi (\ie brings it closer to 1). For the density threshold $k$ in Algorithm~\ref{alg:density}, we use $k=0.2*n$ for all the datasets. 

\smallskip
\noindent\textbf{Implementation.} We implemented \mcc and \scc in Python 3.7.0, and ran experiments on a computing cluster with 9 nodes (2.40 GHz processor and 256 GB RAM). Our code is open-sourced~\cite{supp}.  

\begin{figure*}
	\centering
	\subfloat[Disparate Impact (\edi), LR models]{\includegraphics[width=0.33\linewidth]{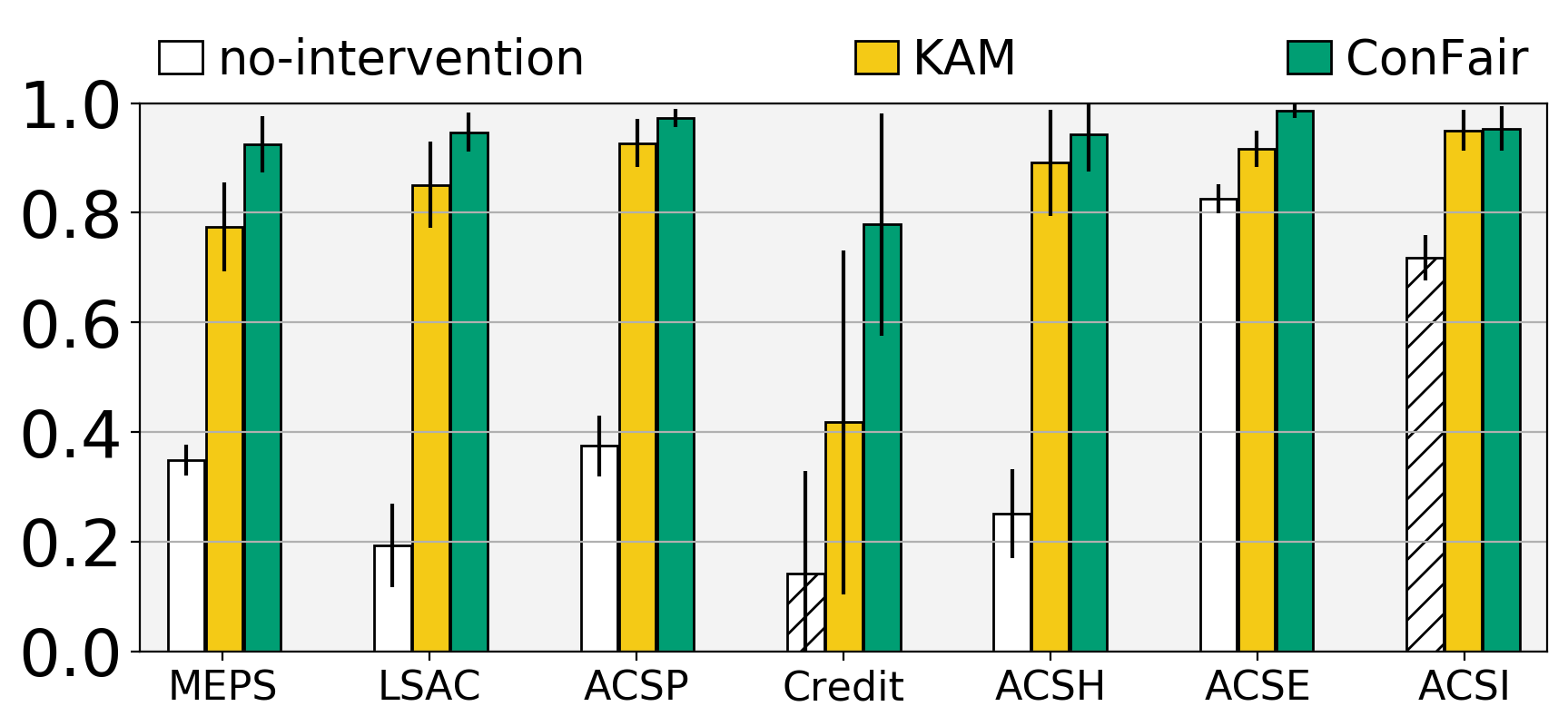}
	\label{fig:scc_kam_lr_di}
	}
	\subfloat[Average Odds Difference (\ead), LR models]{\includegraphics[width=0.33\linewidth]{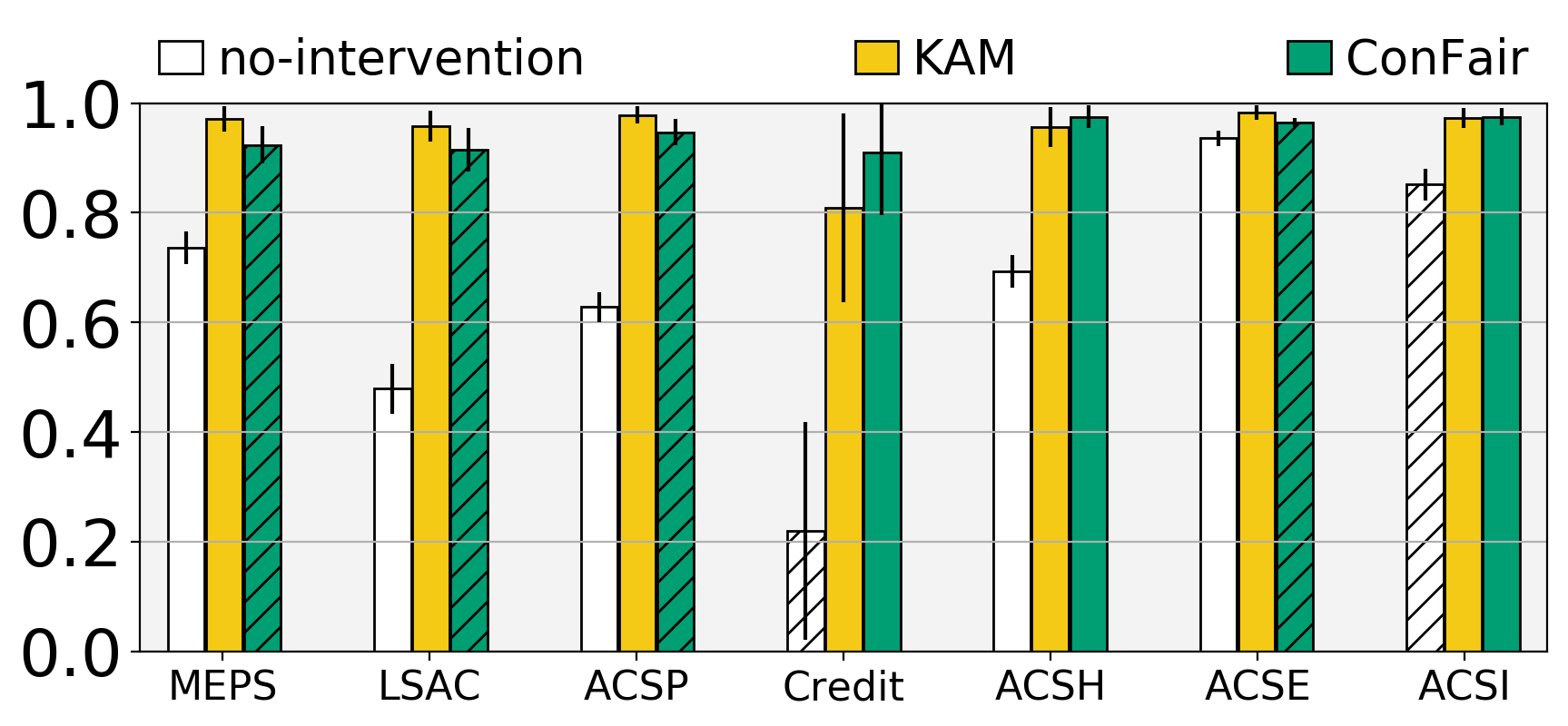}
	\label{fig:scc_kam_lr_aod}
	}
        \subfloat[Balanced Accuracy (\ebacc), LR models]{\includegraphics[width=0.33\linewidth]{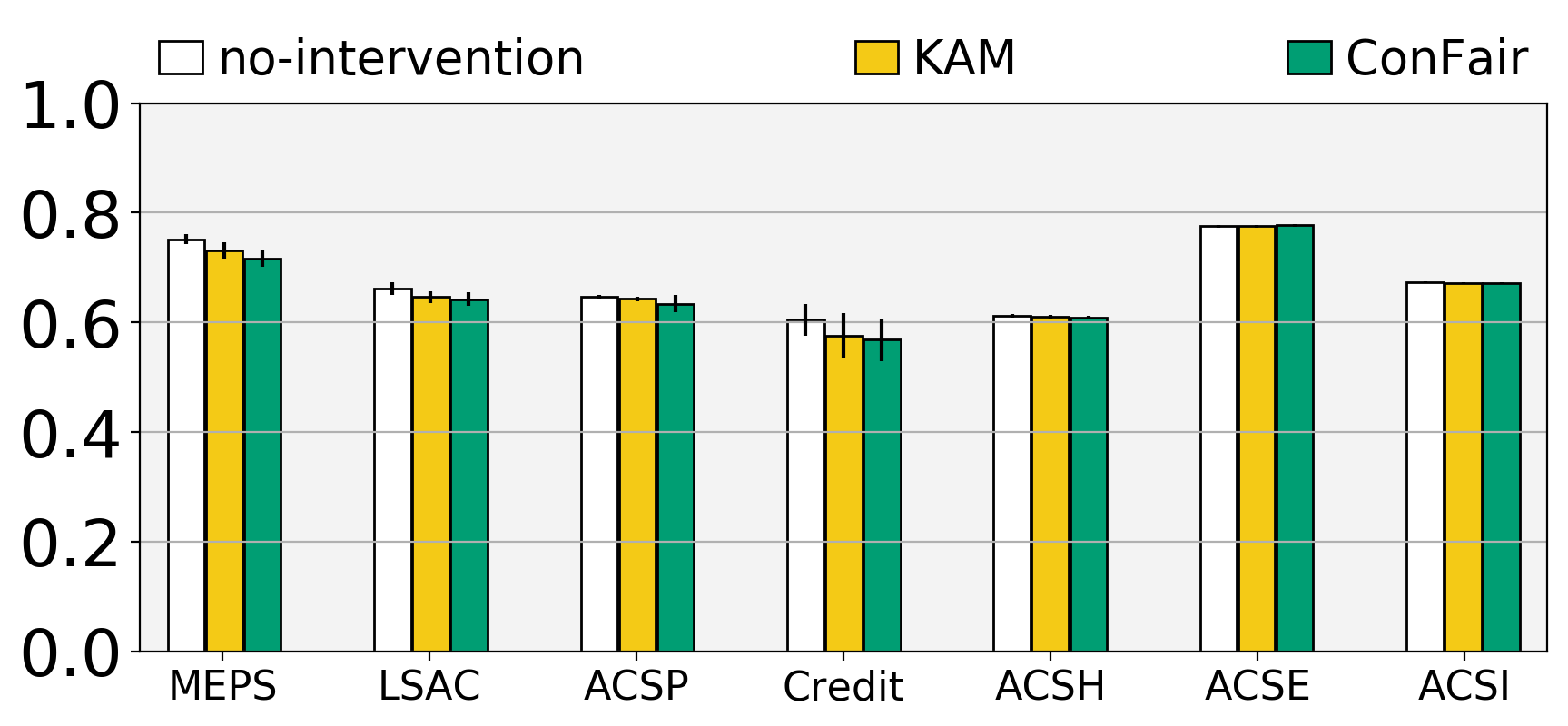}
	\label{fig:scc_kam_lr_balacc}
	}
        \\
        \subfloat[Disparate Impact (\edi), XGB models]{\includegraphics[width=0.33\linewidth]{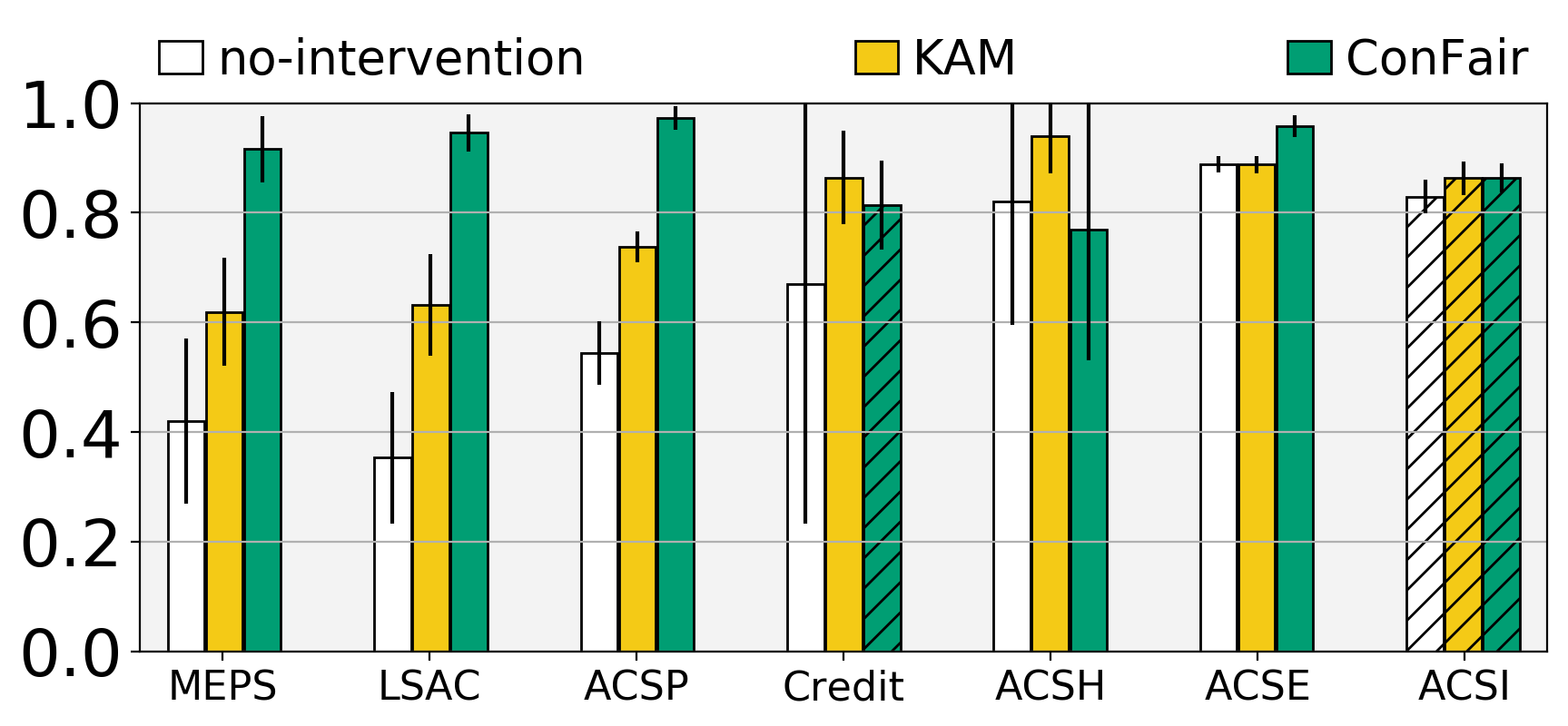}
	\label{fig:scc_kam_tr_di}
	}
	\subfloat[Average Odds Difference (\ead), XGB models]         
        {\includegraphics[width=0.33\linewidth]{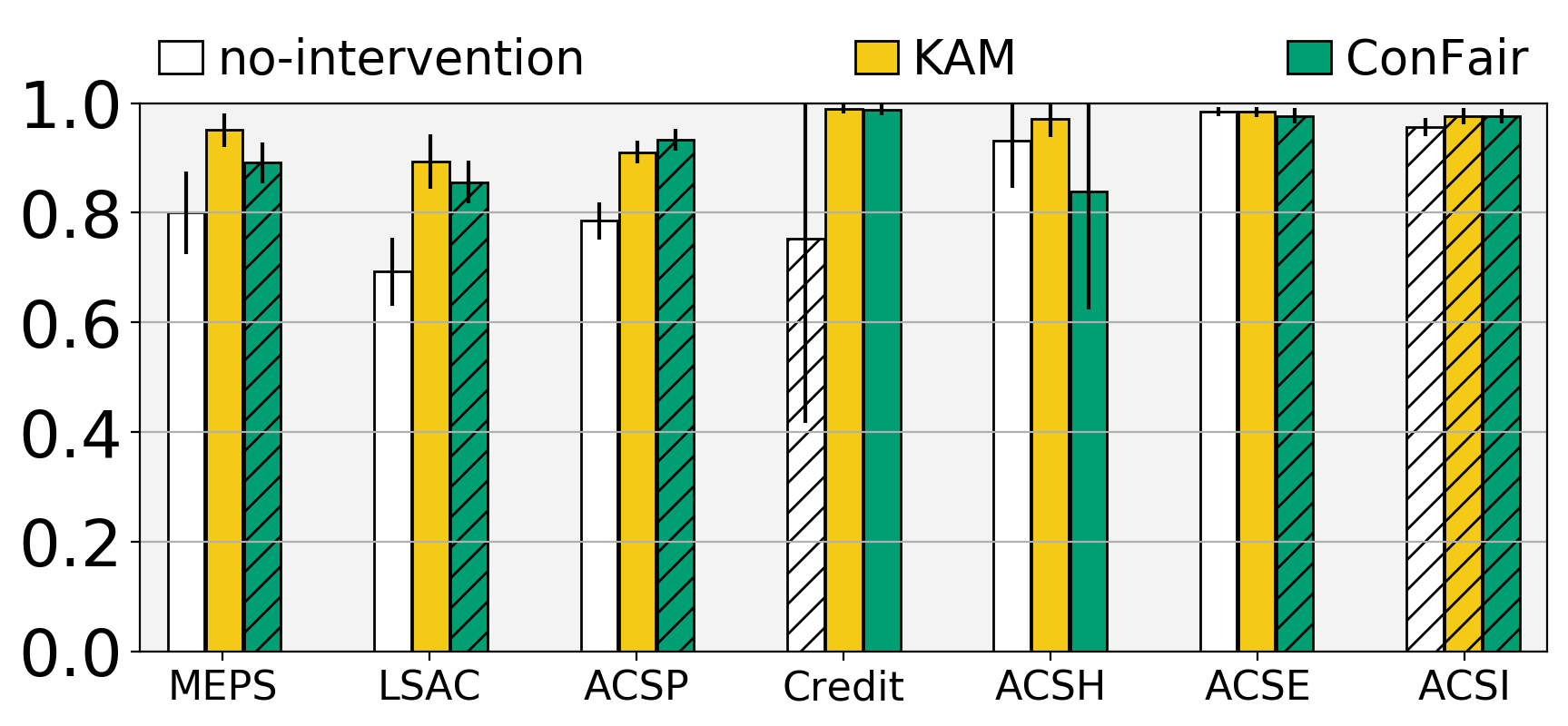}
	\label{fig:scc_kam_tr_aod}
	}
        \subfloat[Balanced Accuracy (\ebacc), XGB models]{\includegraphics[width=0.33\linewidth]{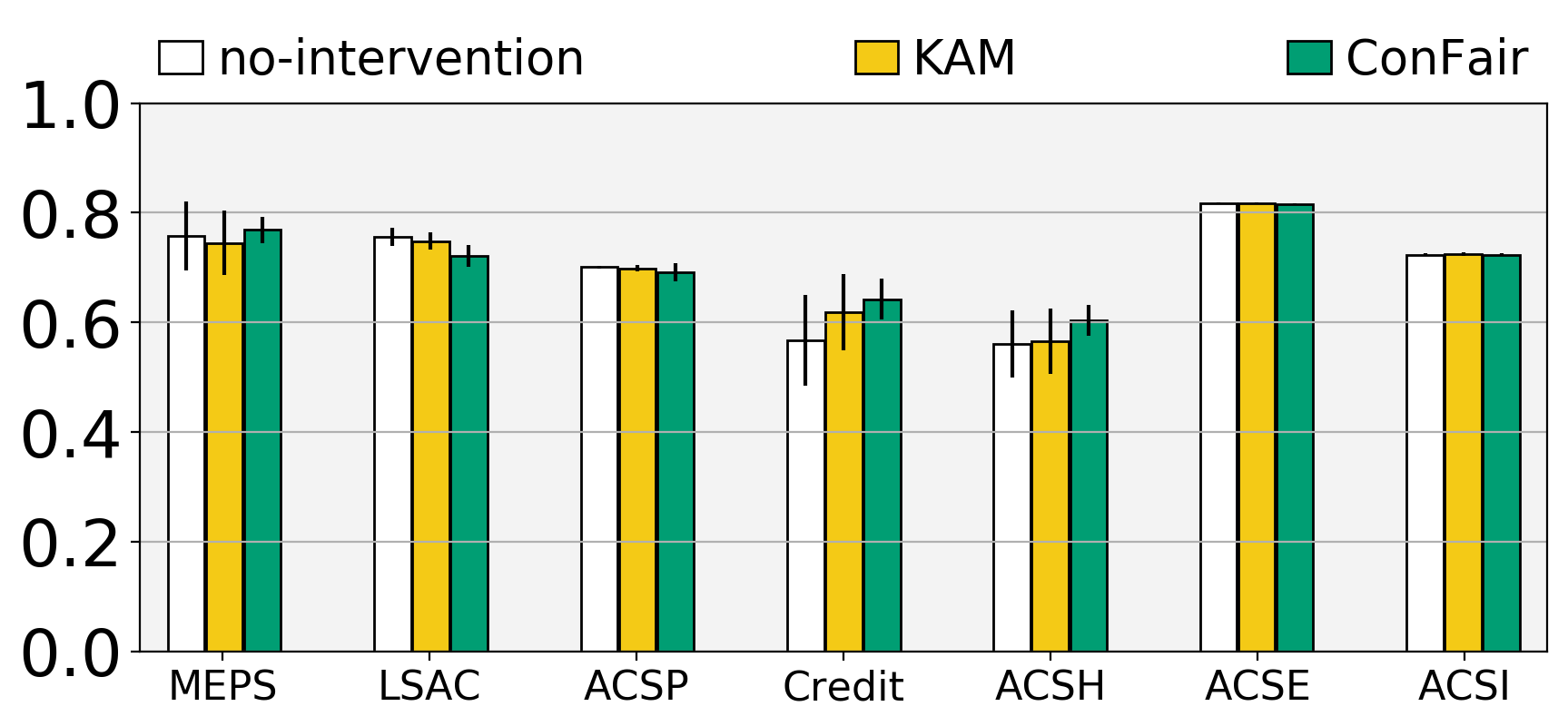}
	\label{fig:scc_kam_tr_balacc}
	}
        \vspace{-0.3em}
	\caption{Comparing \scc to \kam in improving fairness (measured by \edi and \ead) and maintaining utility (measured by \ebacc). Striped bars (\protect\inlinegraphics{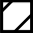}) represent bias favoring minorities.} 
        \vspace{-1em}
\label{fig:scc_kam}
\end{figure*}

\begin{figure*}
	\centering
	\subfloat[Disparate Impact (\edi), LR models]{\includegraphics[width=0.33\linewidth]{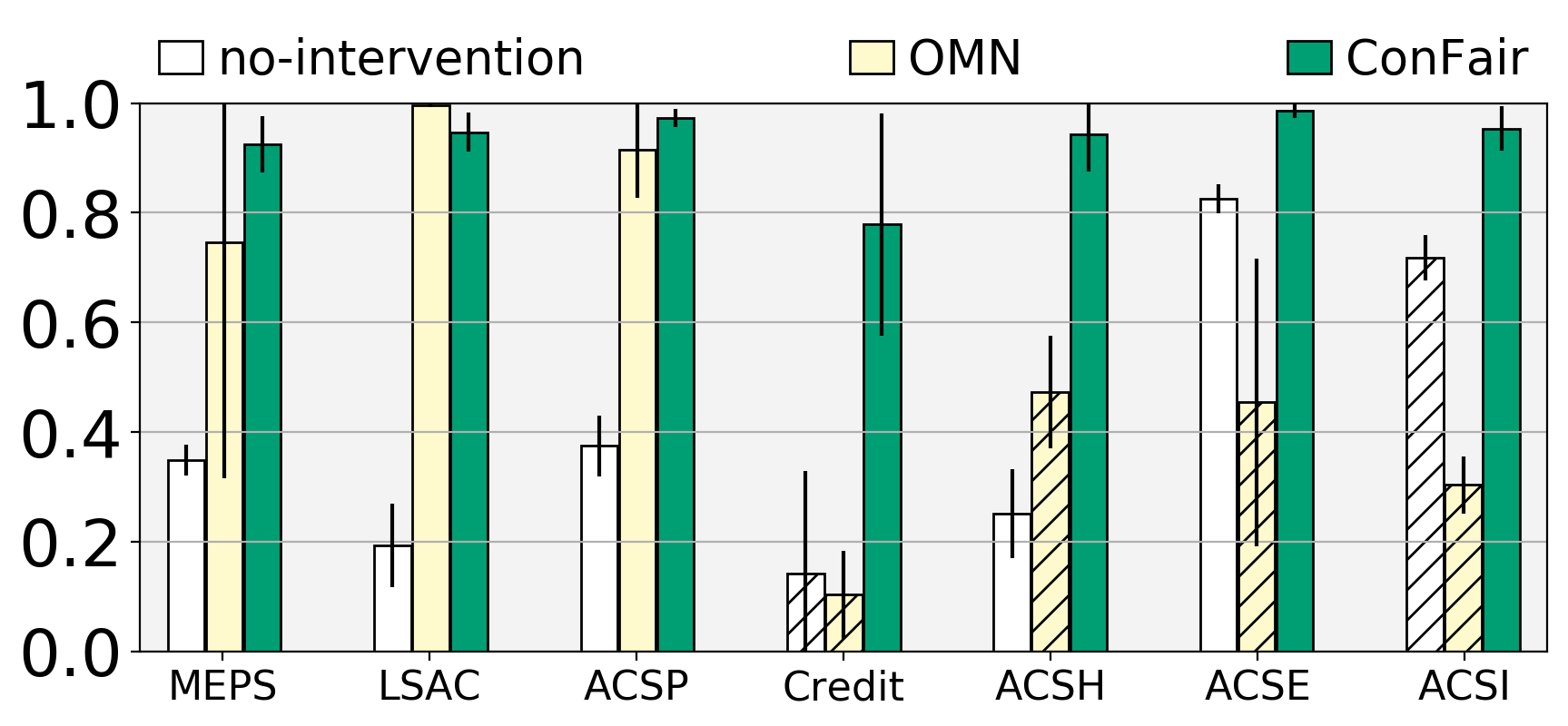}
	\label{fig:scc_other_lr_di}
	}
	\subfloat[Average Odds Difference (\ead), LR models]{\includegraphics[width=0.33\linewidth]{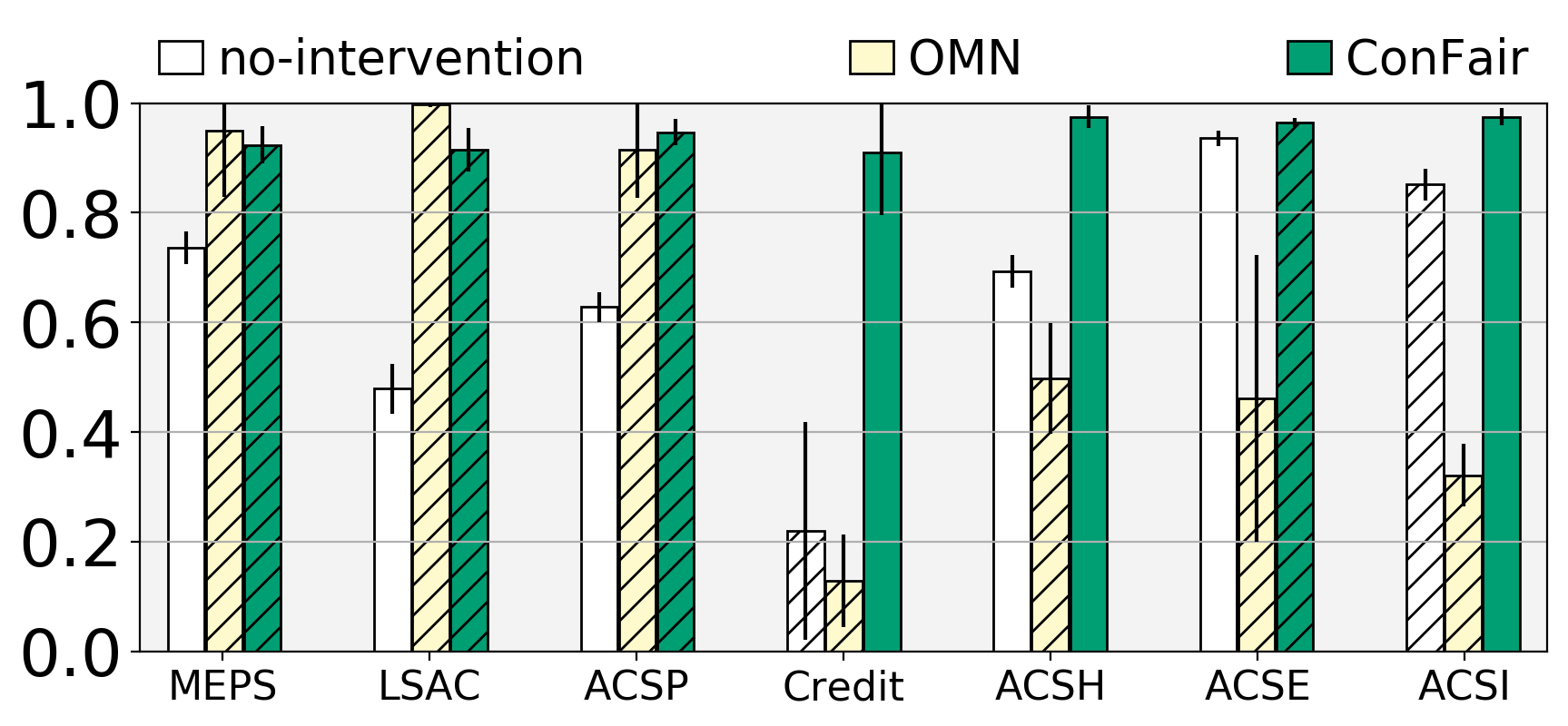}
	\label{fig:scc_other_lr_aod}
	}
        \subfloat[Balanced Accuracy (\ebacc), LR models]{\includegraphics[width=0.33\linewidth]{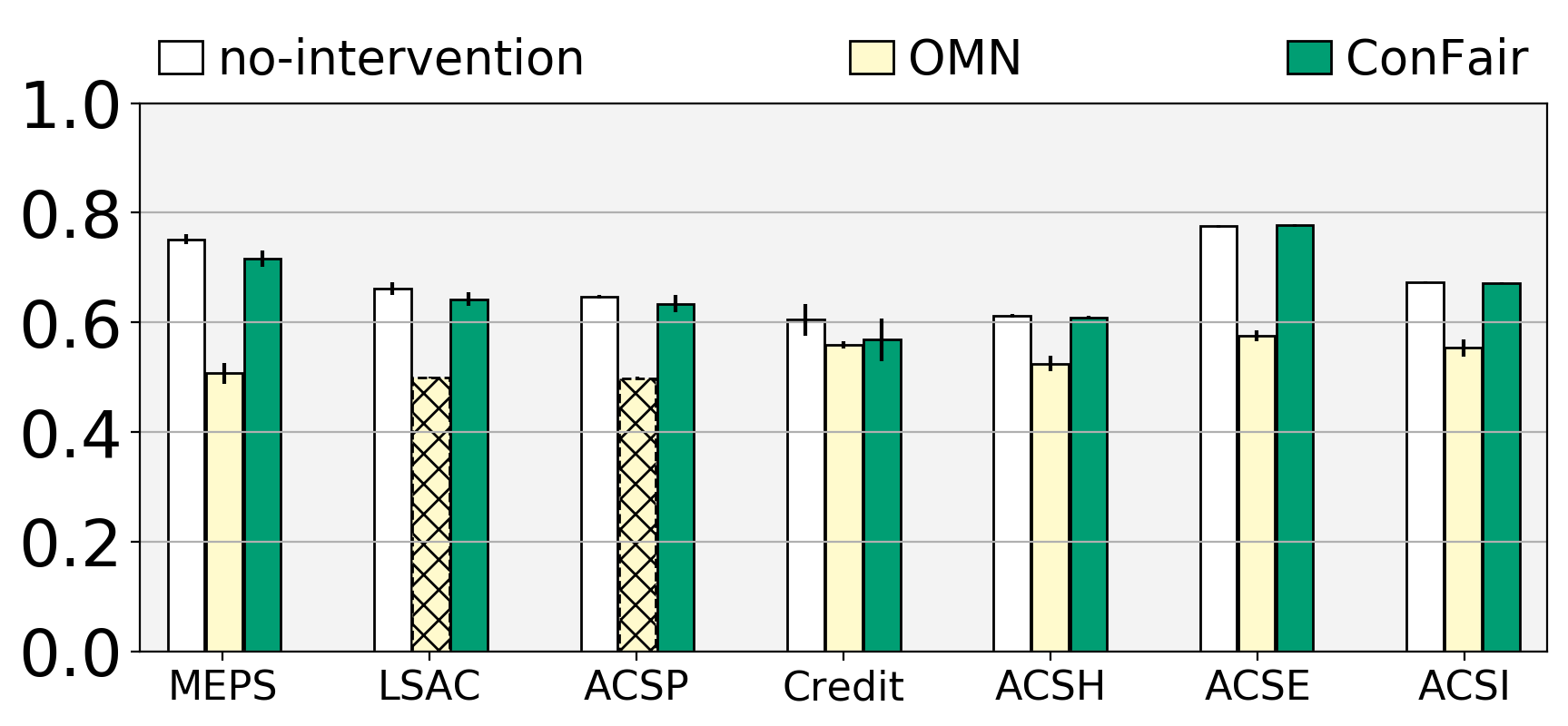}
	\label{fig:scc_other_lr_balacc}
	}
        \\
        \subfloat[Disparate Impact (\edi), XGB models]{\includegraphics[width=0.33\linewidth]{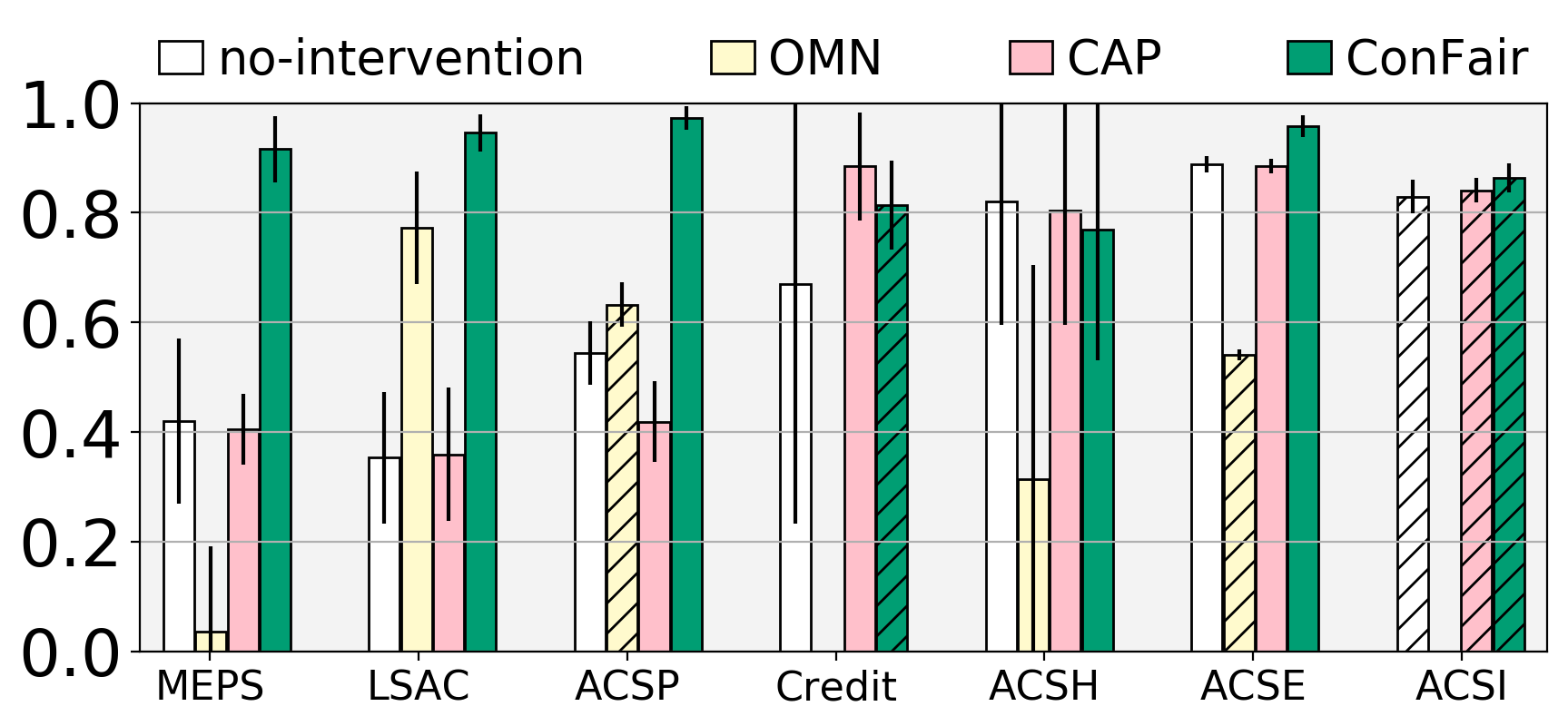}
	\label{fig:scc_other_tr_di}
	}
	\subfloat[Average Odds Difference (\ead), XGB models]         
        {\includegraphics[width=0.33\linewidth]{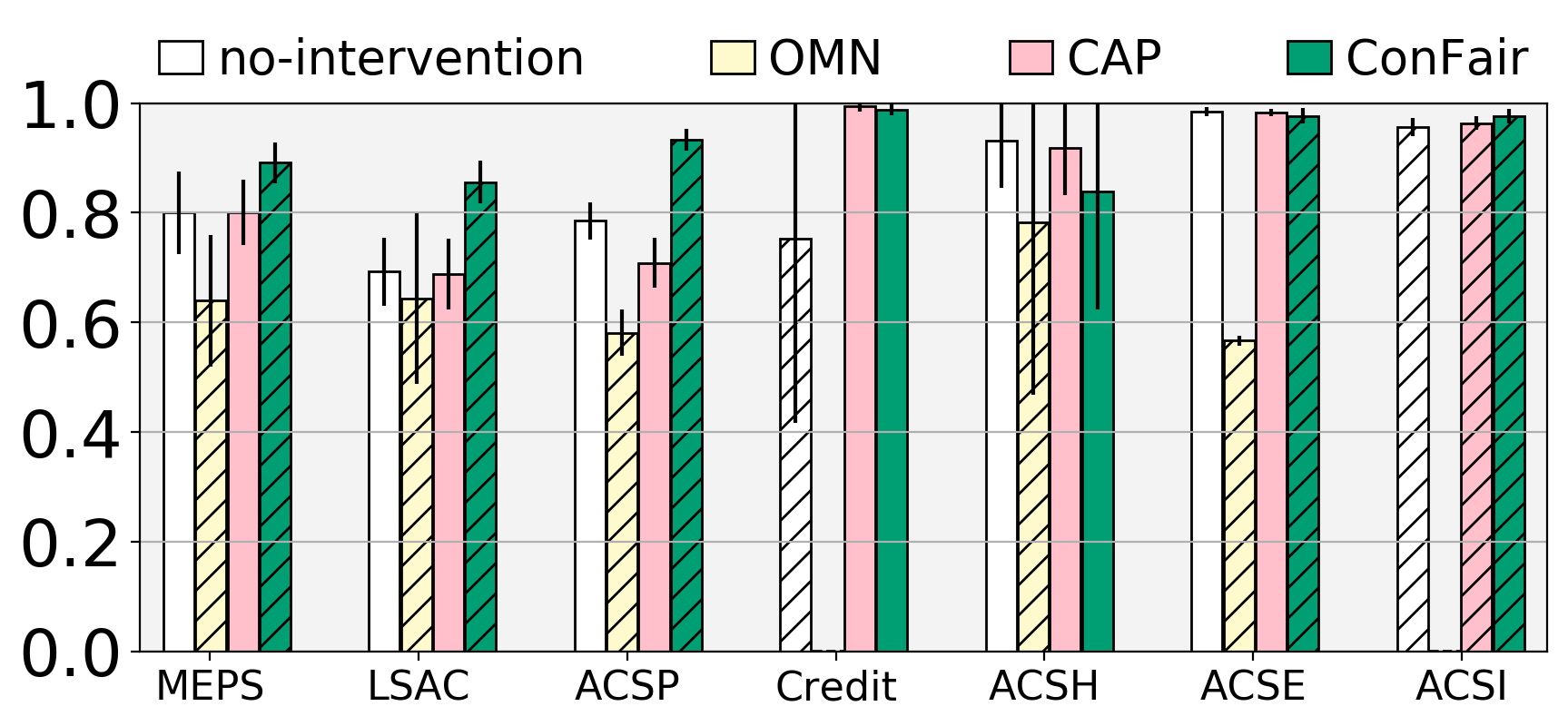}
	\label{fig:scc_other_tr_aod}
	}
        \subfloat[Balanced Accuracy (\ebacc), XGB models]{\includegraphics[width=0.33\linewidth]{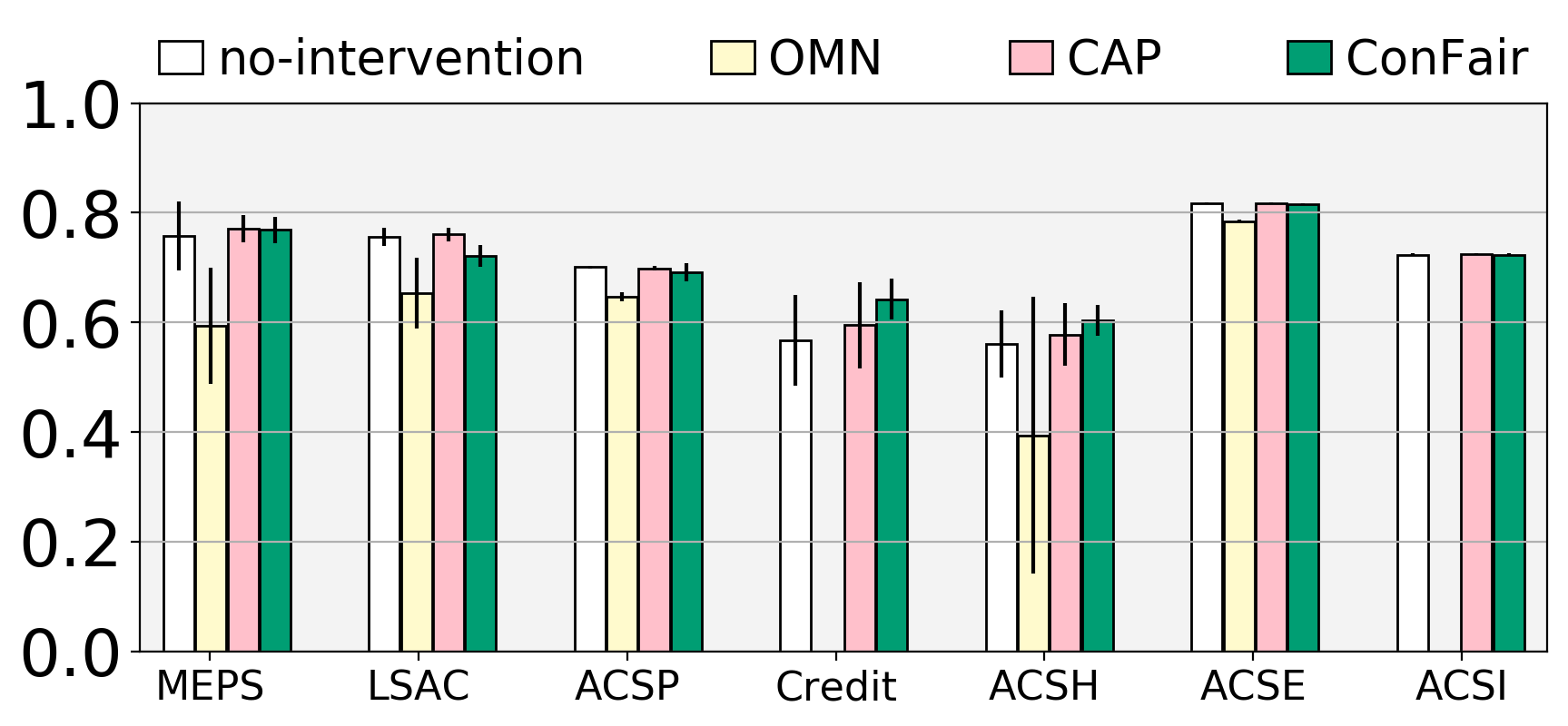}
	\label{fig:scc_other_tr_balacc}
	}
         \vspace{-0.3em}
	\caption{Comparing \scc to \omn and \capu in improving fairness (measured by \edi and \ead) and maintaining utility (measured by \ebacc). Crisscross bars (\protect\inlinegraphics{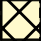}) indicate models that have devolved to useless predictions (e.g., predicting only one class).} 
        \vspace{-1em}
\label{fig:scc_other}
\end{figure*}

\begin{figure*}
	\centering
	\subfloat[Disparate Impact (\edi), LR models]{\includegraphics[width=0.33\linewidth]{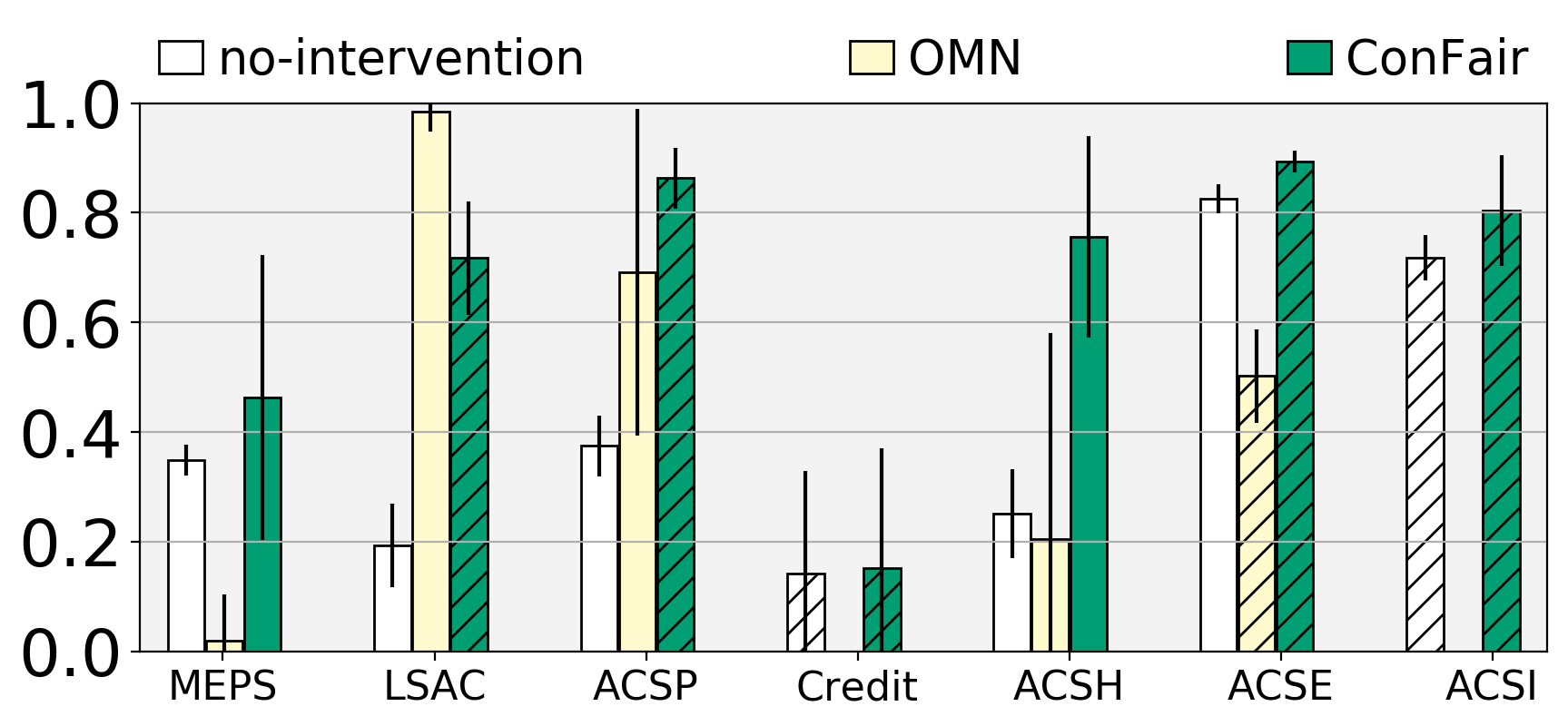}
	\label{fig:scc_aware_lr_di}
	}
	\subfloat[Average Odds Difference (\ead), LR models]{\includegraphics[width=0.33\linewidth]{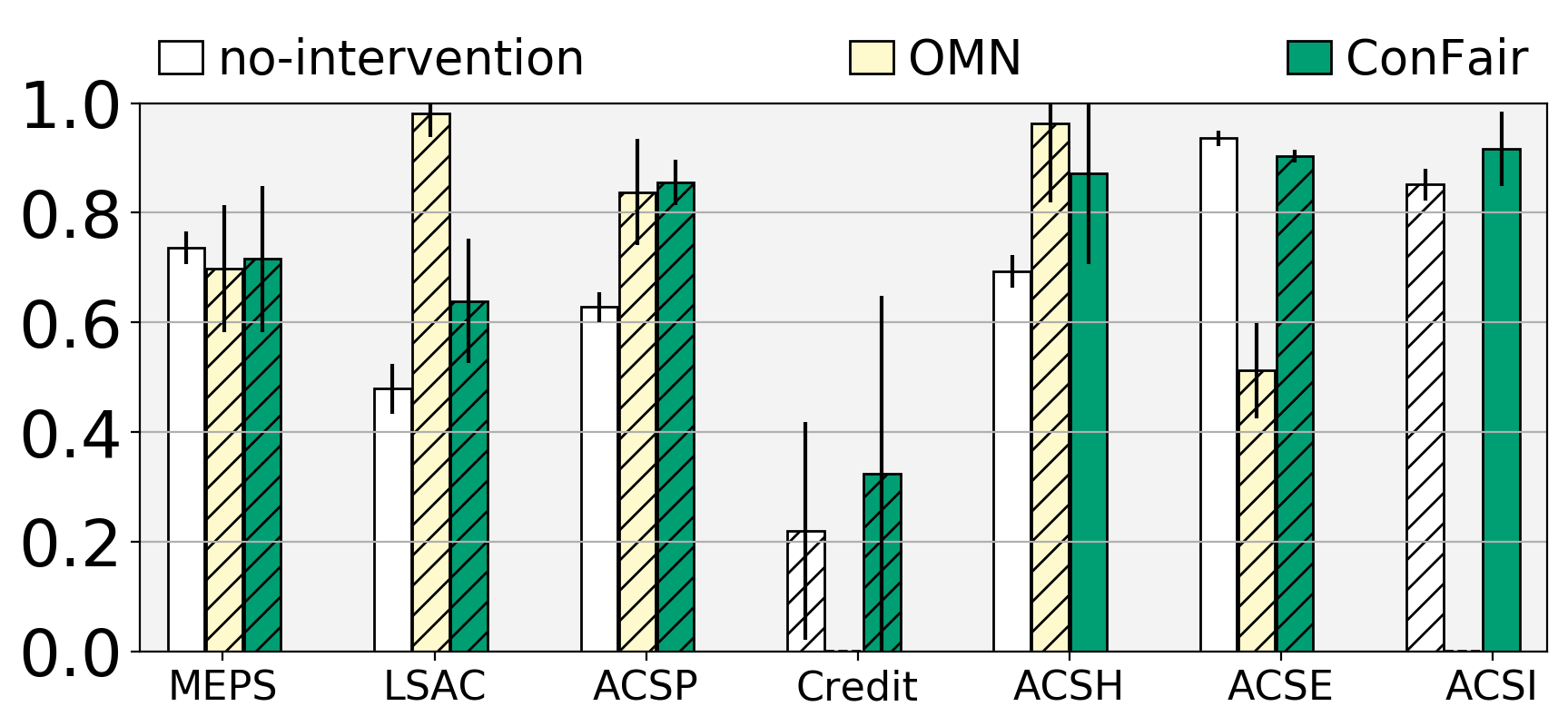}
	\label{fig:scc_aware_lr_aod}
	}
        \subfloat[Balanced Accuracy (\ebacc), LR models]{\includegraphics[width=0.33\linewidth]{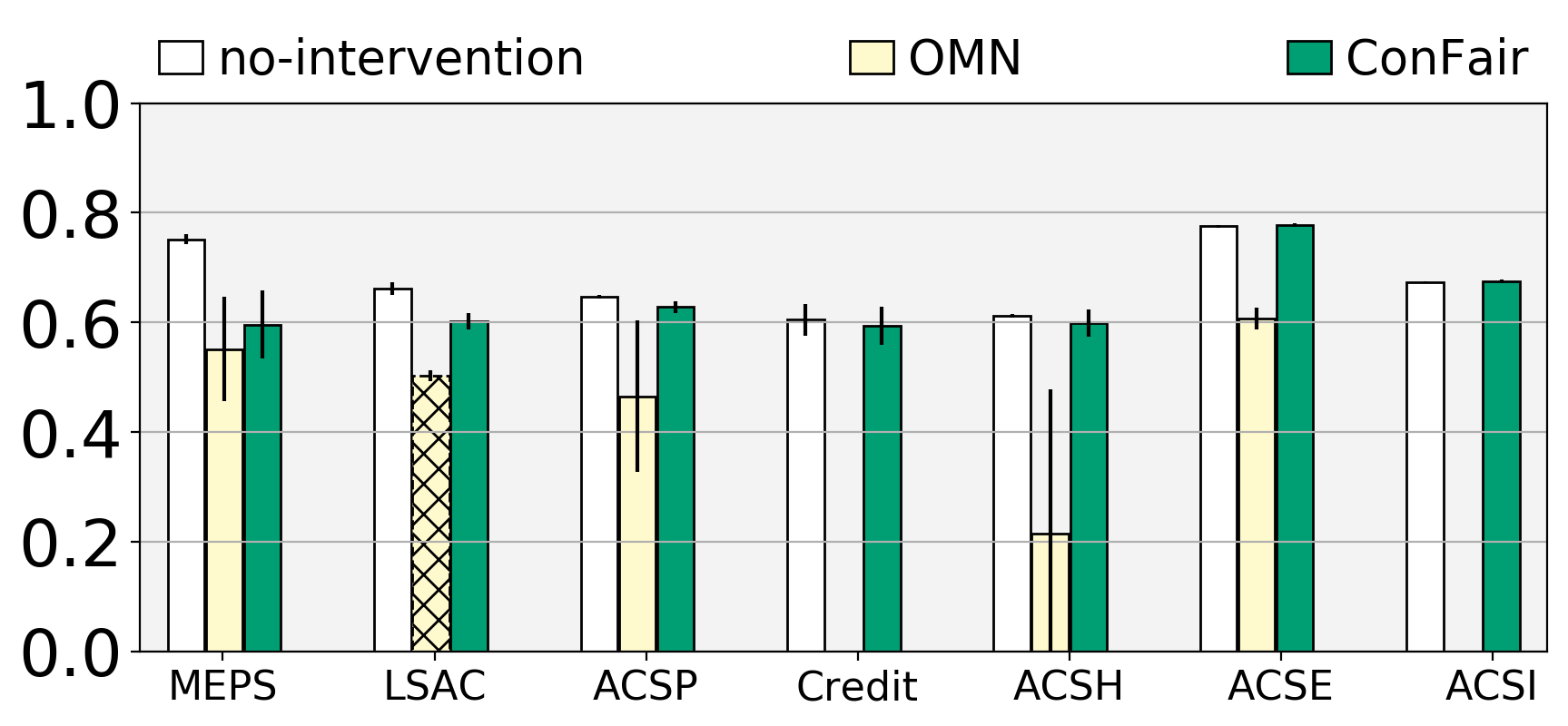}
	\label{fig:scc_aware_lr_balacc}
	}
        \\
        \subfloat[Disparate Impact (\edi), XGB models]{\includegraphics[width=0.33\linewidth]{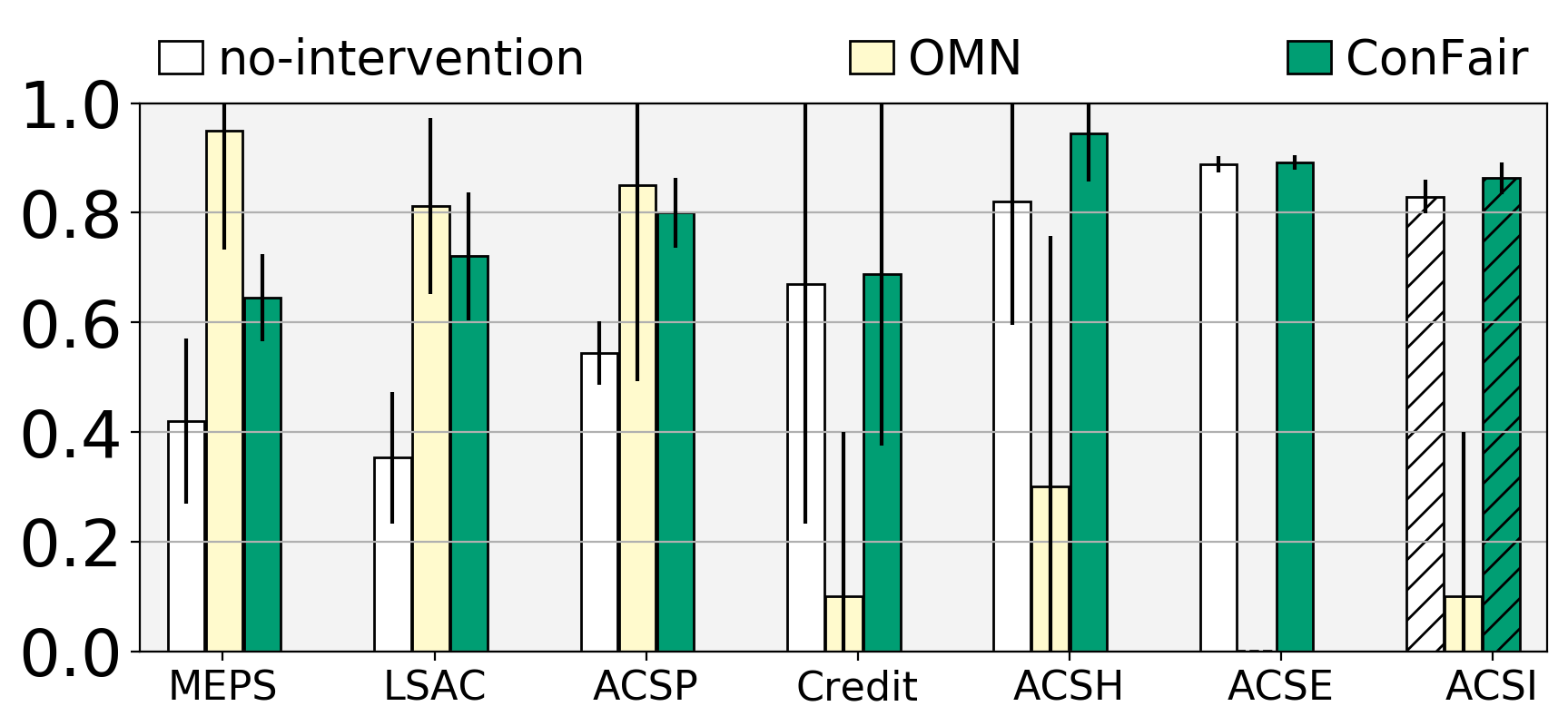}
	\label{fig:scc_aware_tr_di}
	}
	\subfloat[Average Odds Difference (\ead), XGB models]         
        {\includegraphics[width=0.33\linewidth]{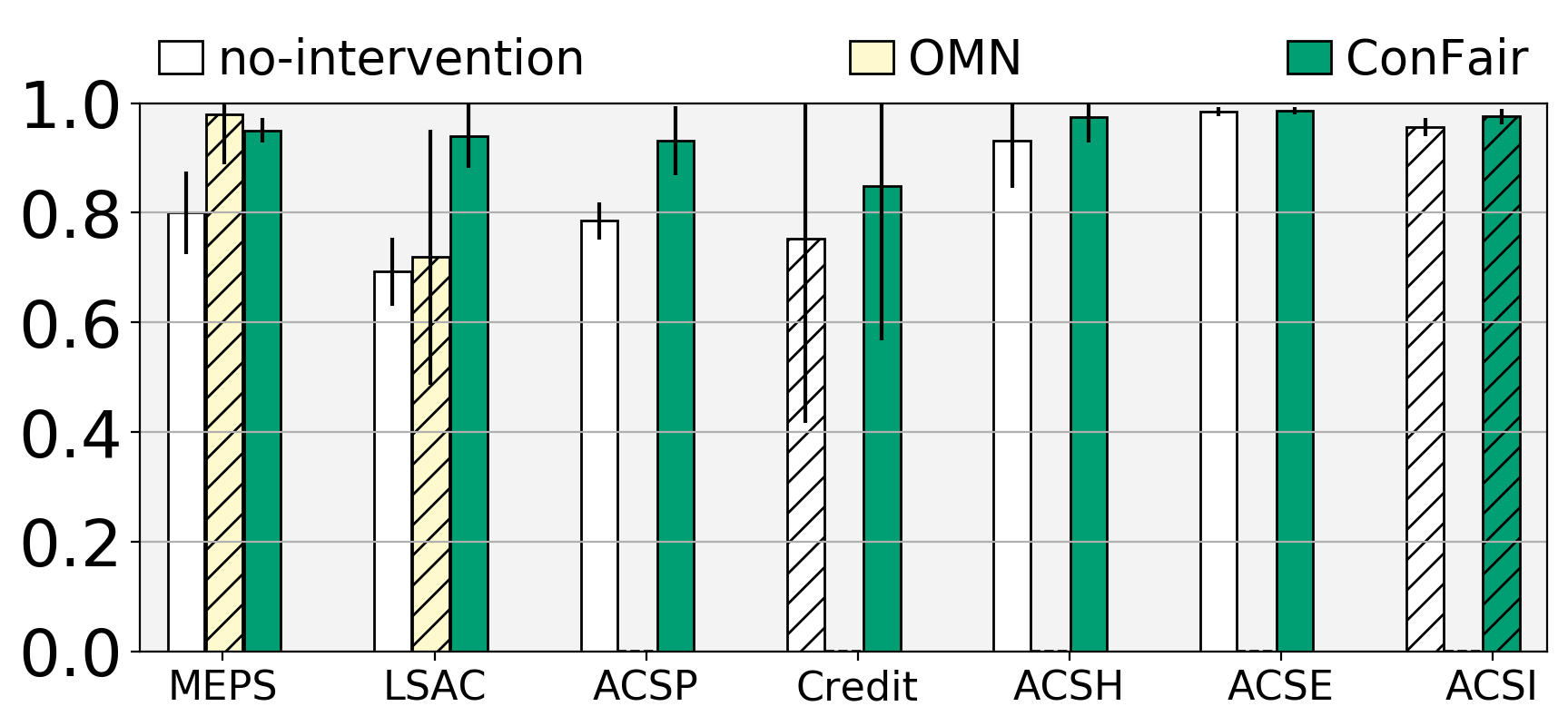}
	\label{fig:scc_aware_tr_aod}
	}
        \subfloat[Balanced Accuracy (\ebacc), XGB models]{\includegraphics[width=0.33\linewidth]{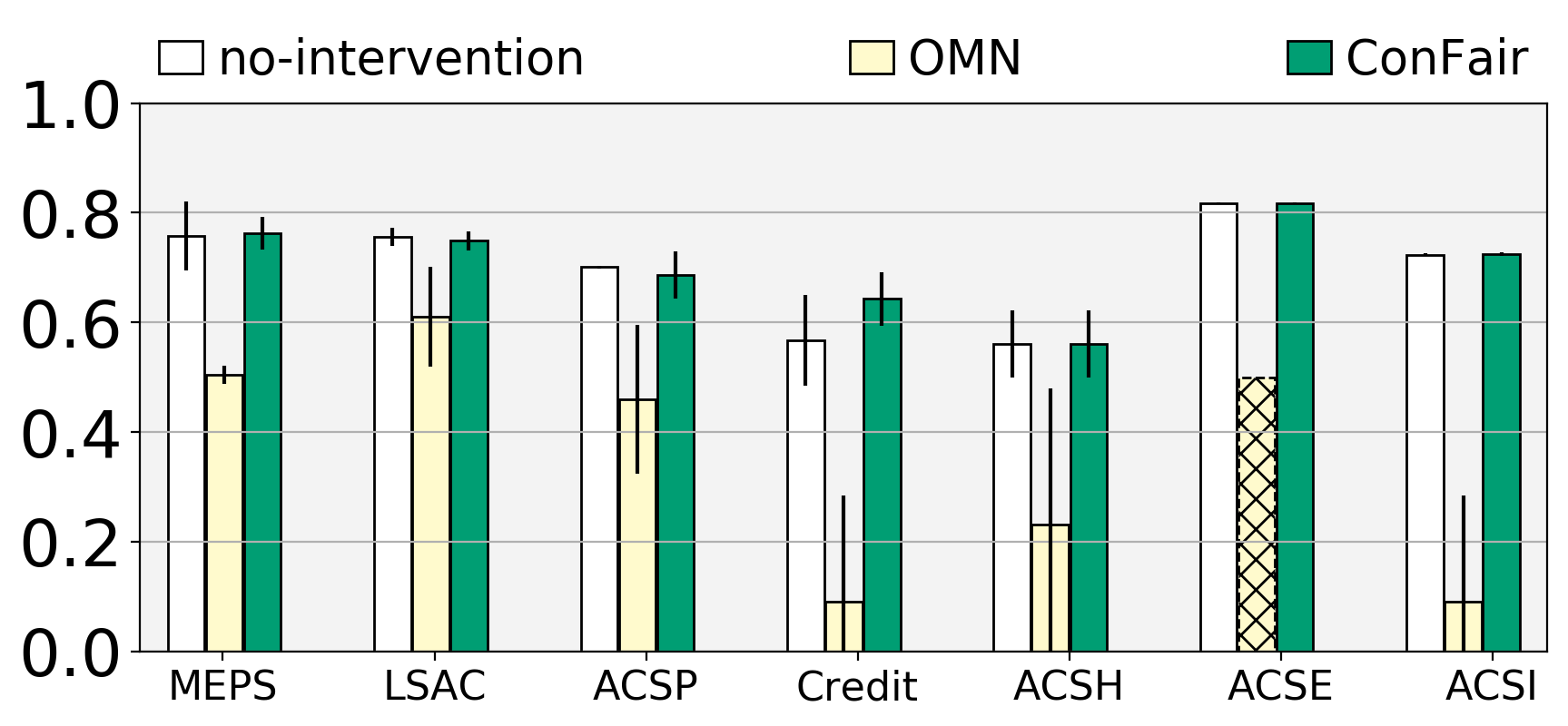}
	\label{fig:scc_aware_tr_balacc}
	}
         \vspace{-0.3em}
	\caption{Comparing \scc to \omn when models are derived using weights that are not tuned for them. In Fig.~\ref{fig:scc_aware_lr_di},~\ref{fig:scc_aware_lr_aod}, and~\ref{fig:scc_aware_lr_balacc}, both methods train an LR model using weights tuned for an XGB model. In Fig.~\ref{fig:scc_aware_tr_di},~\ref{fig:scc_aware_tr_aod}, and~\ref{fig:scc_aware_tr_balacc}, the setting is reversed. } 
        \vspace{-1.5em}
\label{fig:scc_aware}
\end{figure*}

\subsection{Evaluation of \scc}
\label{sec:exp:scc}

We start with the evaluation of \scc, which is our primary fairness-improvement strategy.  (As we will see in Section~\ref{sec:exp:mcc}, \mcc is strong in cases of significant drift, but loses to \scc in most practical settings.)  We compare \scc against three state-of-the-art methods: two reweighing strategies, \kam and \omn, and a data-invasive intervention, \capu. 
We do not compare against in-processing methods~\cite{hashimoto2018fairness,krasanakis2018adaptive,lahoti2020fairness,jiang2020identifying}, which alter the learners, or post-processing methods~\cite{DBLP:conf/icdm/KamiranKZ12,DBLP:conf/nips/HardtPNS16,DBLP:conf/nips/PleissRWKW17}, which alter predictions.
A broad evaluation against the existing extensive landscape of fairness interventions is in itself an independent research contribution~\cite{IslamFMS2022}.

\smallskip
\noindent\textbf{\scc vs \kam.} 
As we noted, prior methods have employed weighing strategies as a fairness intervention, but the novelty of \scc lies in the use of CCs to identify and increase the weights of tuples that conform to the densest areas of the input.  In contrast, prior art like \kam increases the weights of all tuples within a group, which may end up amplifying outliers and noise.  

Figure~\ref{fig:scc_kam} demonstrates a comparison between \scc and \kam across 7 datasets and two learning strategies.  The white bars in the graphs show the performance of the LR (Figures~\ref{fig:scc_kam_lr_di}, \ref{fig:scc_kam_lr_aod}, and~\ref{fig:scc_kam_lr_balacc}) and XGB models (Figures~\ref{fig:scc_kam_tr_di}, \ref{fig:scc_kam_tr_aod}, and~\ref{fig:scc_kam_tr_balacc}) before any fairness interventions are applied (\base).  We note that many of these results indicate significant bias (low fairness measures).  \scc and \kam both succeed at improving the fairness of predictions, without notable drops in accuracy (Figures~\ref{fig:scc_kam_lr_balacc} and~\ref{fig:scc_kam_tr_balacc}).

We note that \scc robustly outperforms \kam with respect to the \edi metric over the \dmp, \dlg, \dap, and \dae datasets, in the case of XGB models (Fig.~\ref{fig:scc_kam_tr_di}); its gains are clear, yet more modest, in the case of LR models (Fig.~\ref{fig:scc_kam_lr_di}).  Even though \scc does not directly target the \ead metric, it still achieves significant improvements over \base, and comparable performance to \kam (Figures~\ref{fig:scc_kam_lr_aod} and~\ref{fig:scc_kam_tr_aod}).  It is important also to highlight that, while the \ead values are similar between \scc and \kam, \scc more reliably favors the minority group (striped bars~\inlinegraphics{figs/Extras/stripes.png}).

For the datasets \dcr and \dai, \scc performs better against the LR models, concerning both \edi and \ead, and is closer to \kam against the XGB models.
For the \dah dataset, \scc outperforms \kam in improving the fairness (with respect to \edi) of the LR models but behaves worse than \kam with the XGB models. This is because the \dah dataset is very sensitive to the intervention factors of \scc, which are used to determine the increment of tuples' weights. We found that the fine-tuning of these factors for this dataset specifies that $\alpha^u=0.028$, while the optimum value of $\alpha^u$ for this dataset is 0.03.
Overall, \scc gains an edge over \kam as it is able to fine-tune weights for tuples within groups, which is not possible in \kam. 

\begin{figure*}
	\centering
	\subfloat[\scc targets \edi by Selection Rate]{\includegraphics[width=0.33\linewidth]{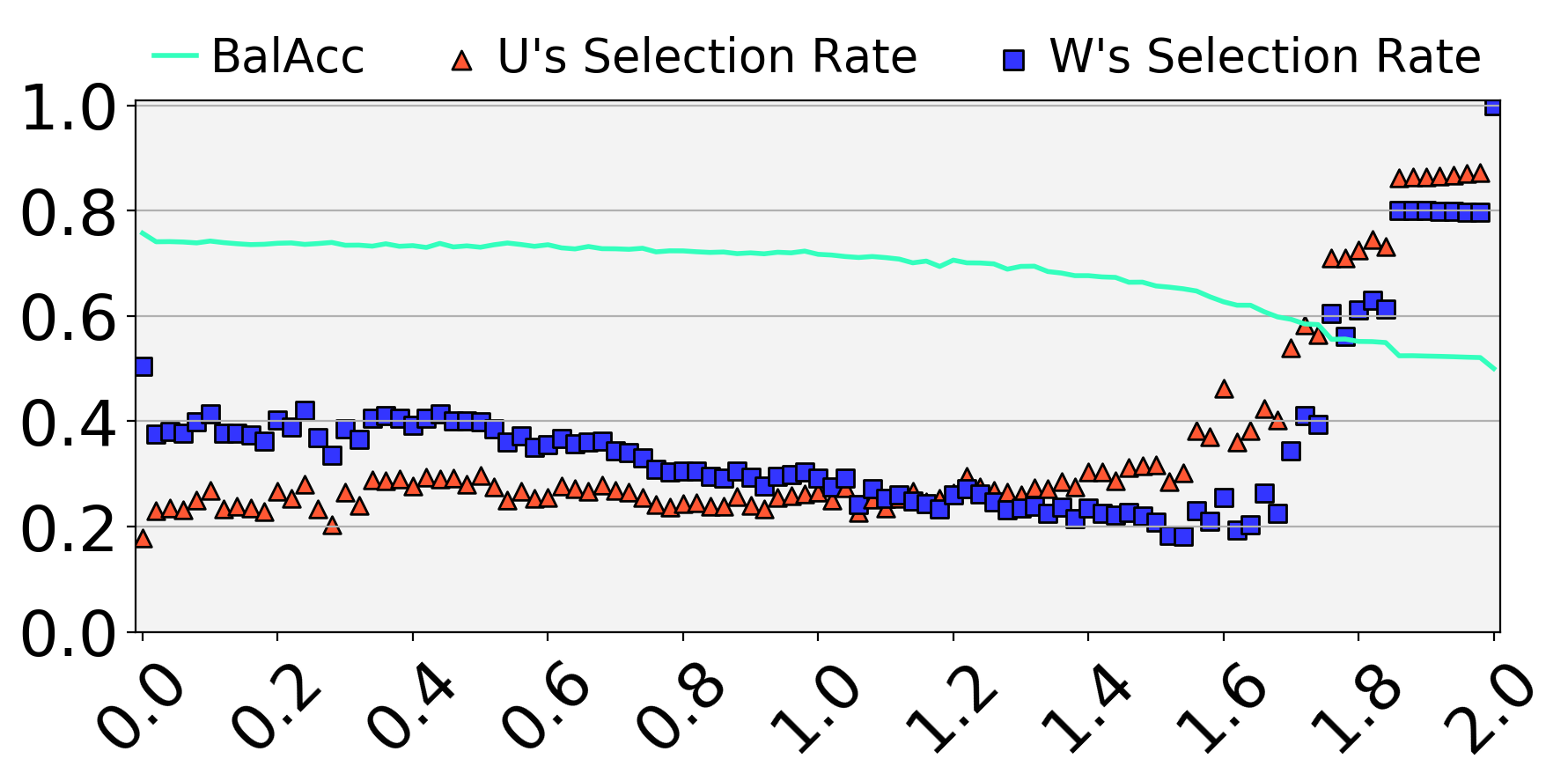}
	\label{fig:mp_scc_di}
	}
	\subfloat[\scc targets \eeo by FNR]         {\includegraphics[width=0.33\linewidth]{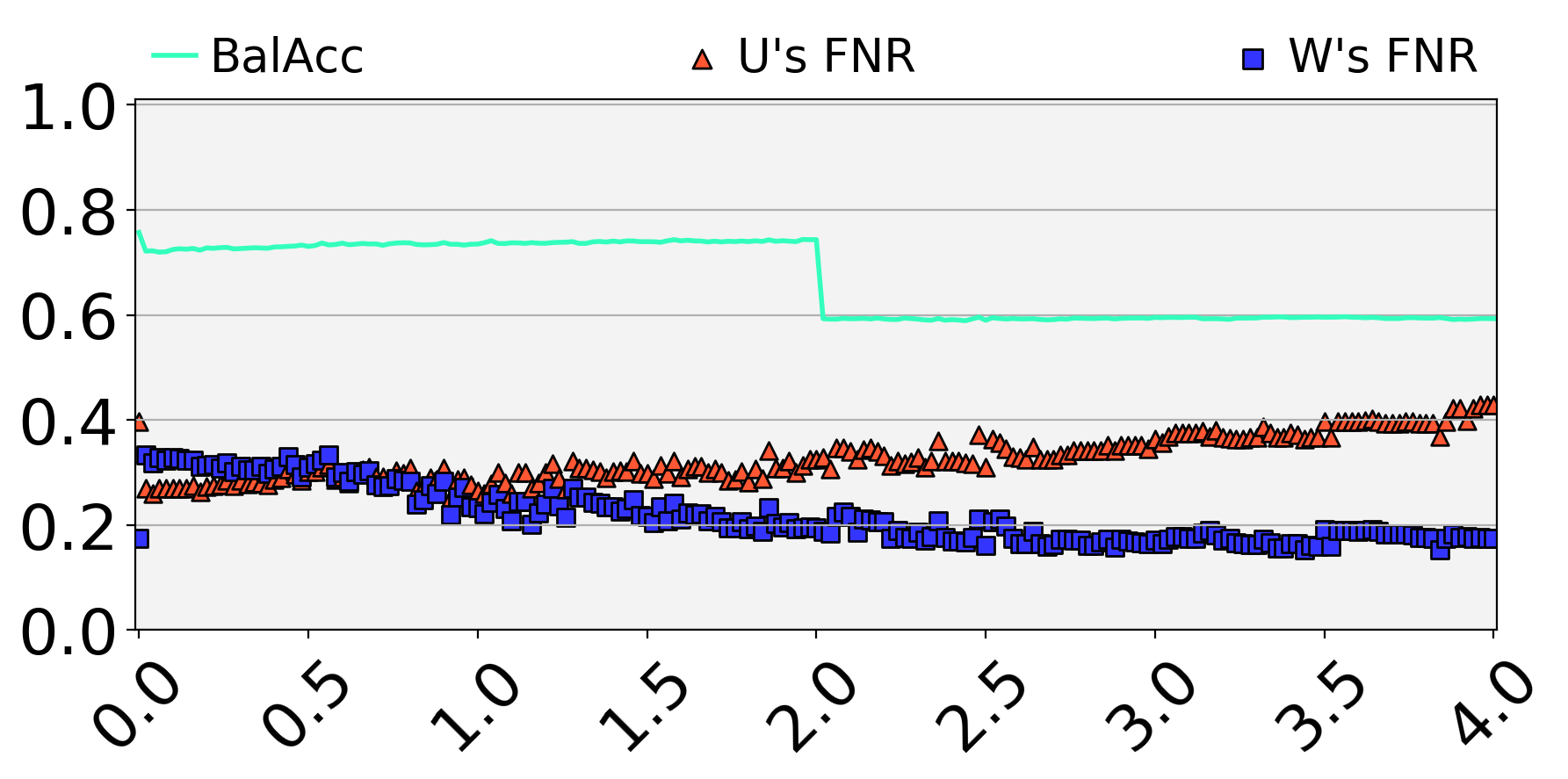}
	\label{fig:mp_scc_fnr}
	}
        \subfloat[\scc targets \eeo by FPR]{\includegraphics[width=0.33\linewidth]{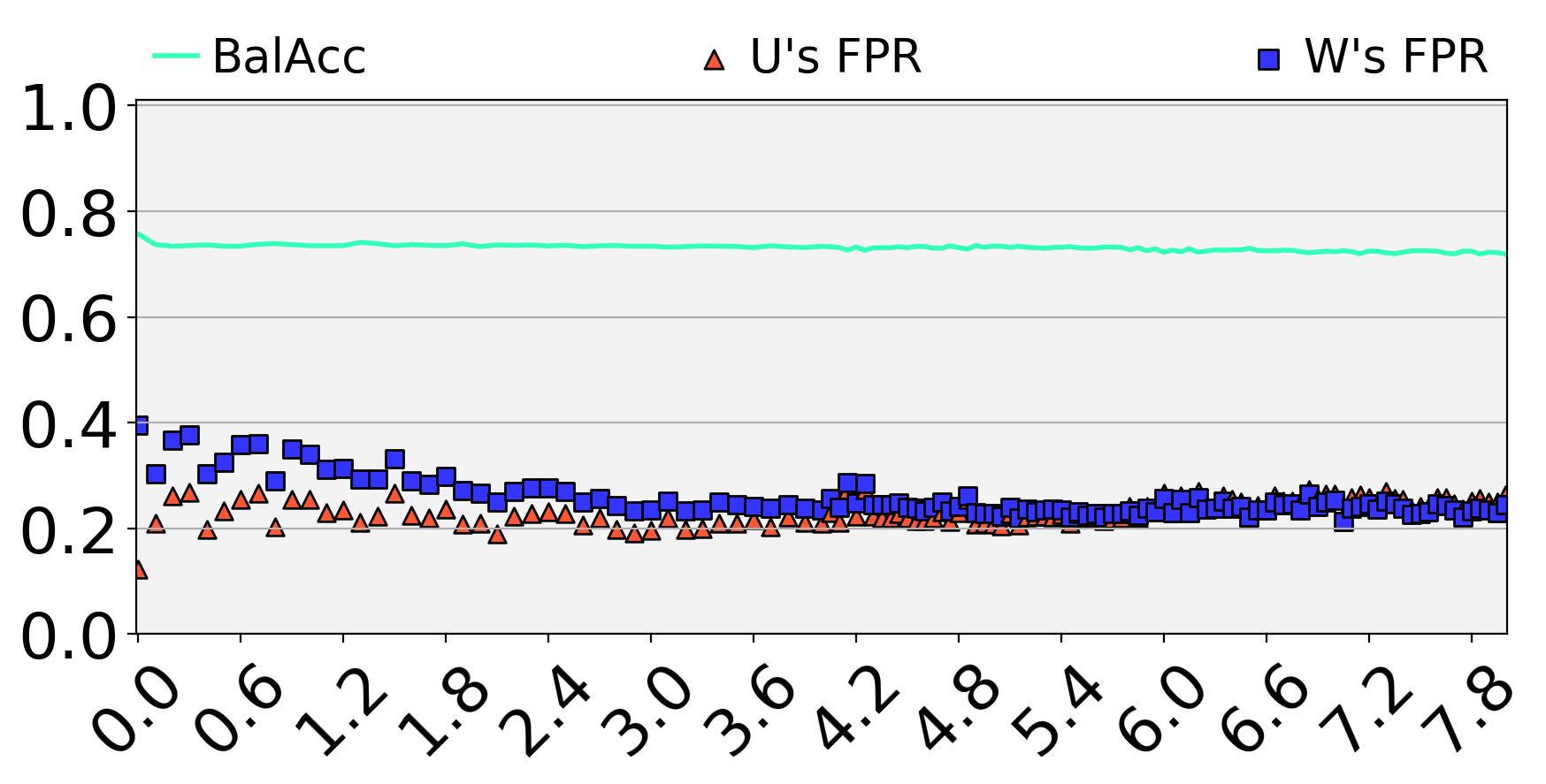}
	\label{fig:mp_scc_fpr}
	}
        \\
        \subfloat[\omn targets \edi by Selection Rate]{\includegraphics[width=0.33\linewidth]{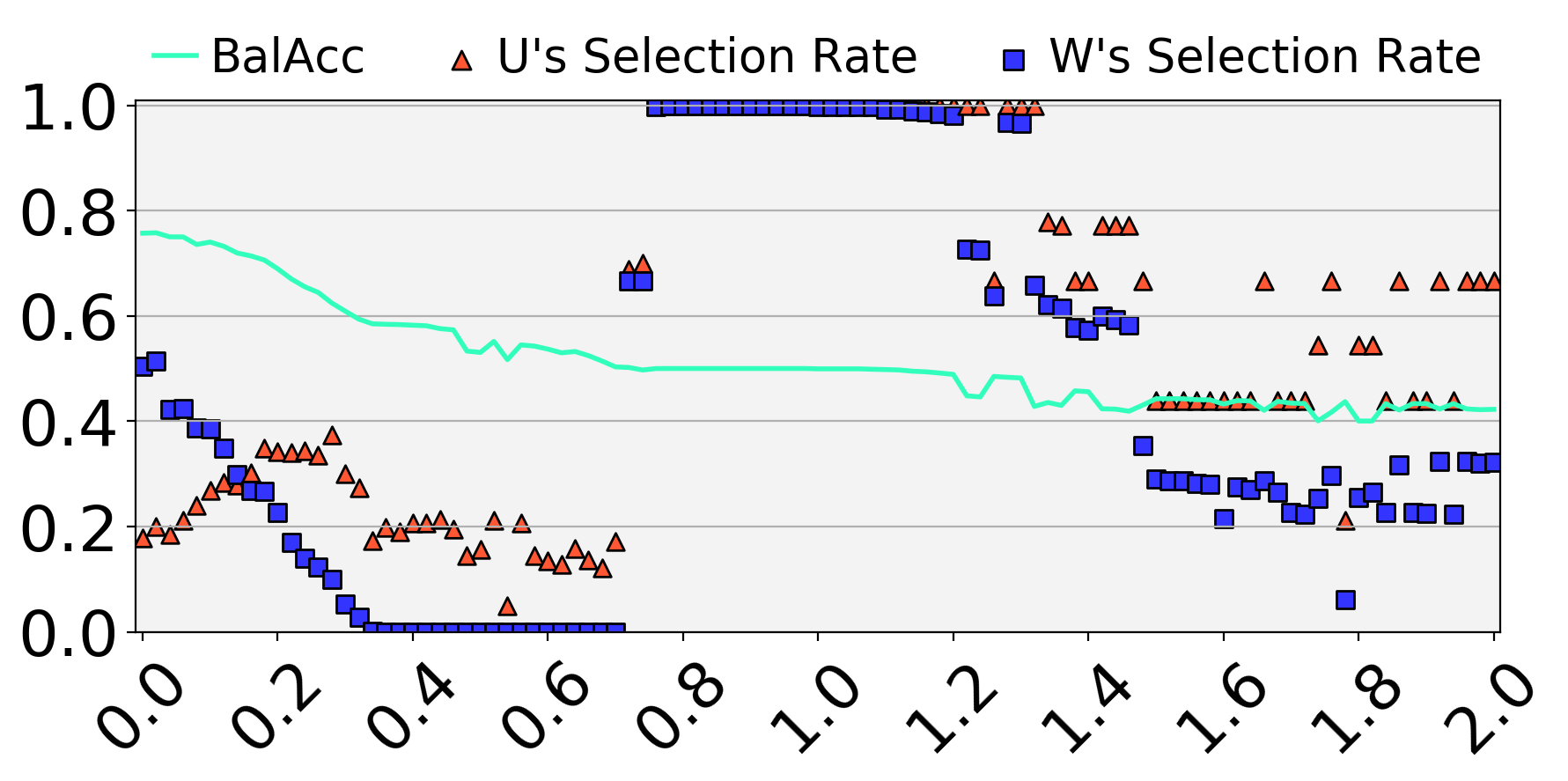}
	\label{fig:mp_omn_di}
	}
	\subfloat[\omn targets \eeo by FNR]         
        {\includegraphics[width=0.33\linewidth]{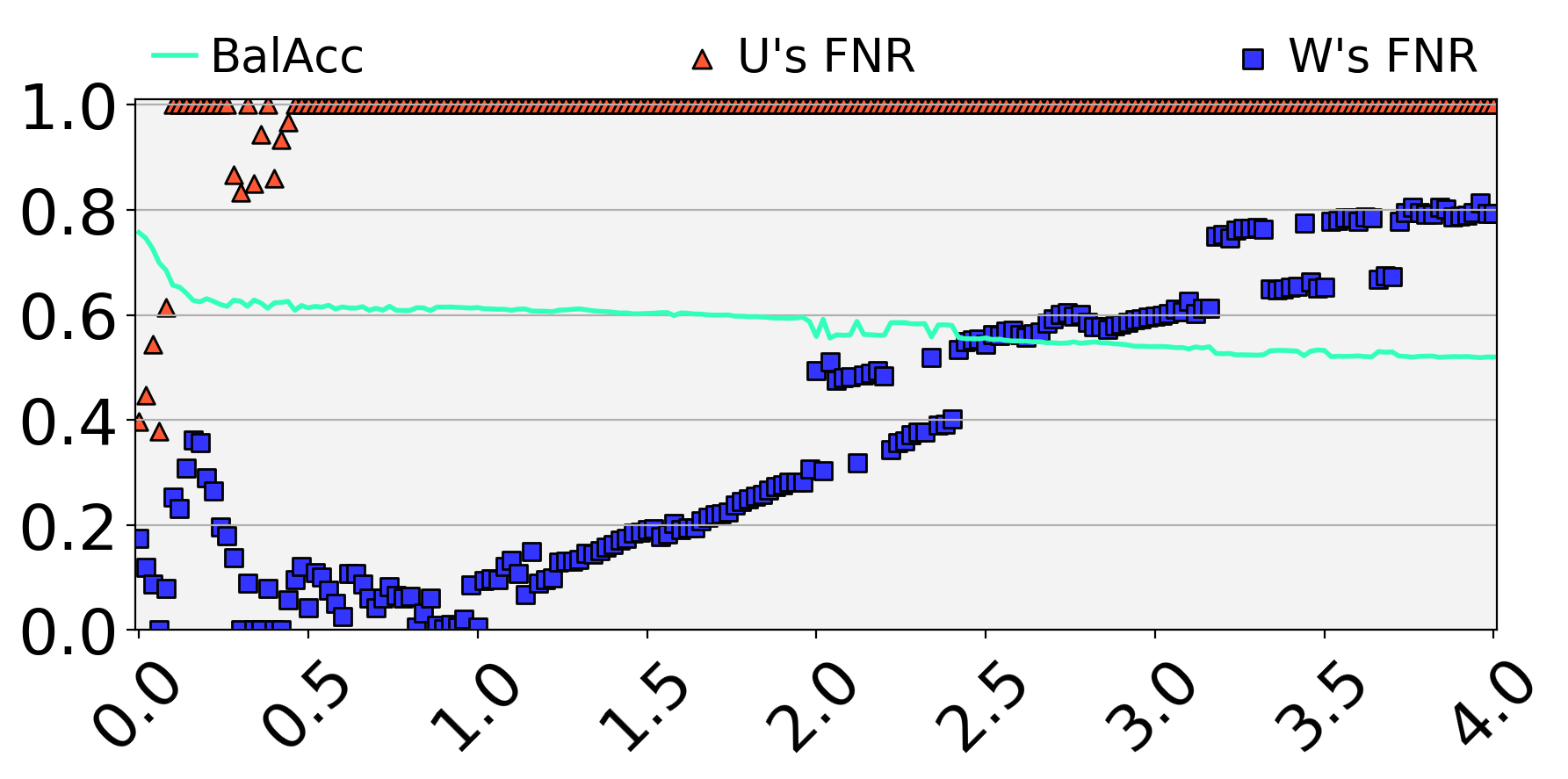}
	\label{fig:mp_omn_fnr}
	}
        \subfloat[\omn targets \eeo by FPR]{\includegraphics[width=0.33\linewidth]{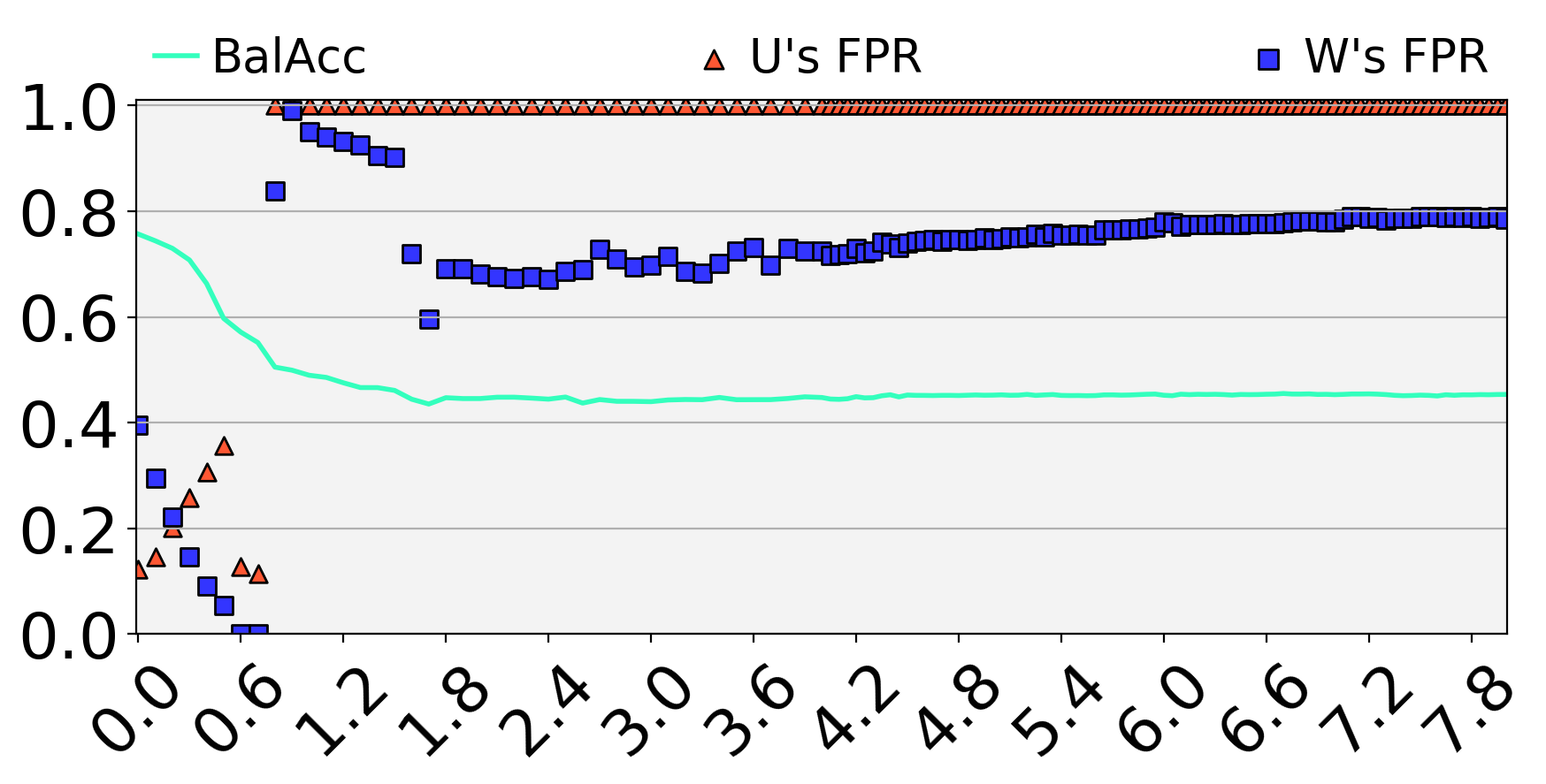}
	\label{fig:mp_omn_fpr}
	}
	\caption{Comparison of \scc and \omn with respect to the degree of intervention over the \dmp dataset. The values of the metrics for minority $\mathbf{U}$ and majority $\mathbf{W}$ groups are highlighted with red triangles (\protect\inlinegraphics{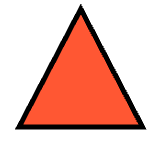}) and blue squares (\protect\inlinegraphics{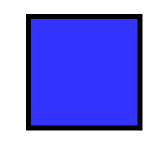}), respectively.  Fairness is measured as the difference in the corresponding metrics between the two groups, \ie perfect fairness is achieved when red triangles overlap with blue squares. The utility of ML models (measured by \ebacc) is highlighted with a green line.
    In Figures~\ref{fig:mp_scc_di},~\ref{fig:mp_scc_fnr}, and~\ref{fig:mp_scc_fpr}, \scc trains a LR model using the corresponding weights (\ie $alpha^u$ shown on the $X$-axis and $alpha^w=0$) and optimizes \edi and \eeo by FNR and FPR, respectively. The beginning of the x-axis ($x=0$) denotes the performance of models without any intervention. In Figures~\ref{fig:mp_omn_di},~\ref{fig:mp_omn_fnr}, and~\ref{fig:mp_omn_fpr}, \omn works under the same setting with its input parameter $\lambda$ shown on the $X$-axis.
    } 
\label{fig:relation_mp}
\end{figure*}

\begin{figure*}
	\centering
	\subfloat[\scc targets \edi by Selection Rate]{\includegraphics[width=0.33\linewidth]{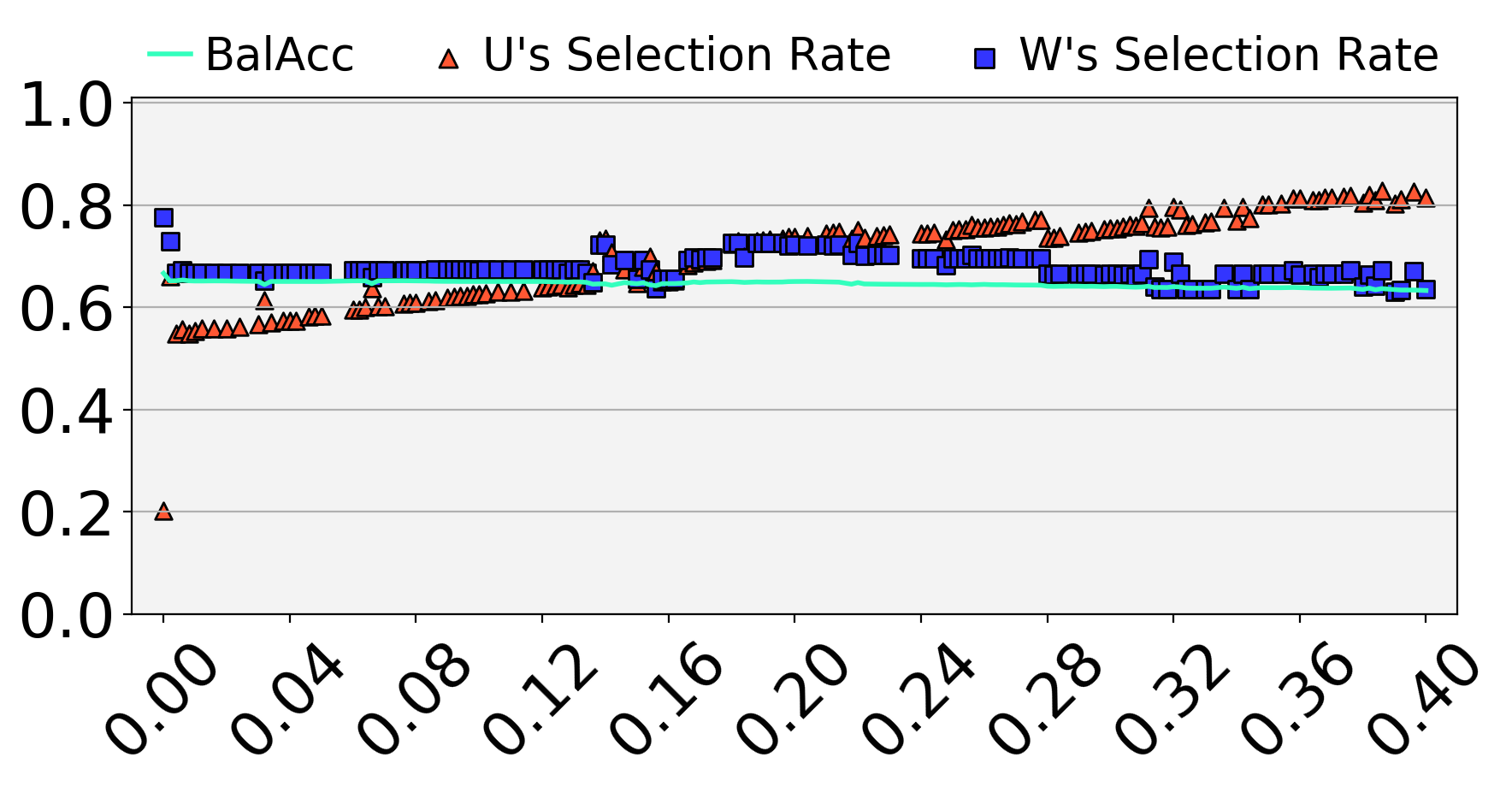}
	\label{fig:lg_scc_di}
	}
	\subfloat[\scc targets \eeo by FNR]{\includegraphics[width=0.33\linewidth]{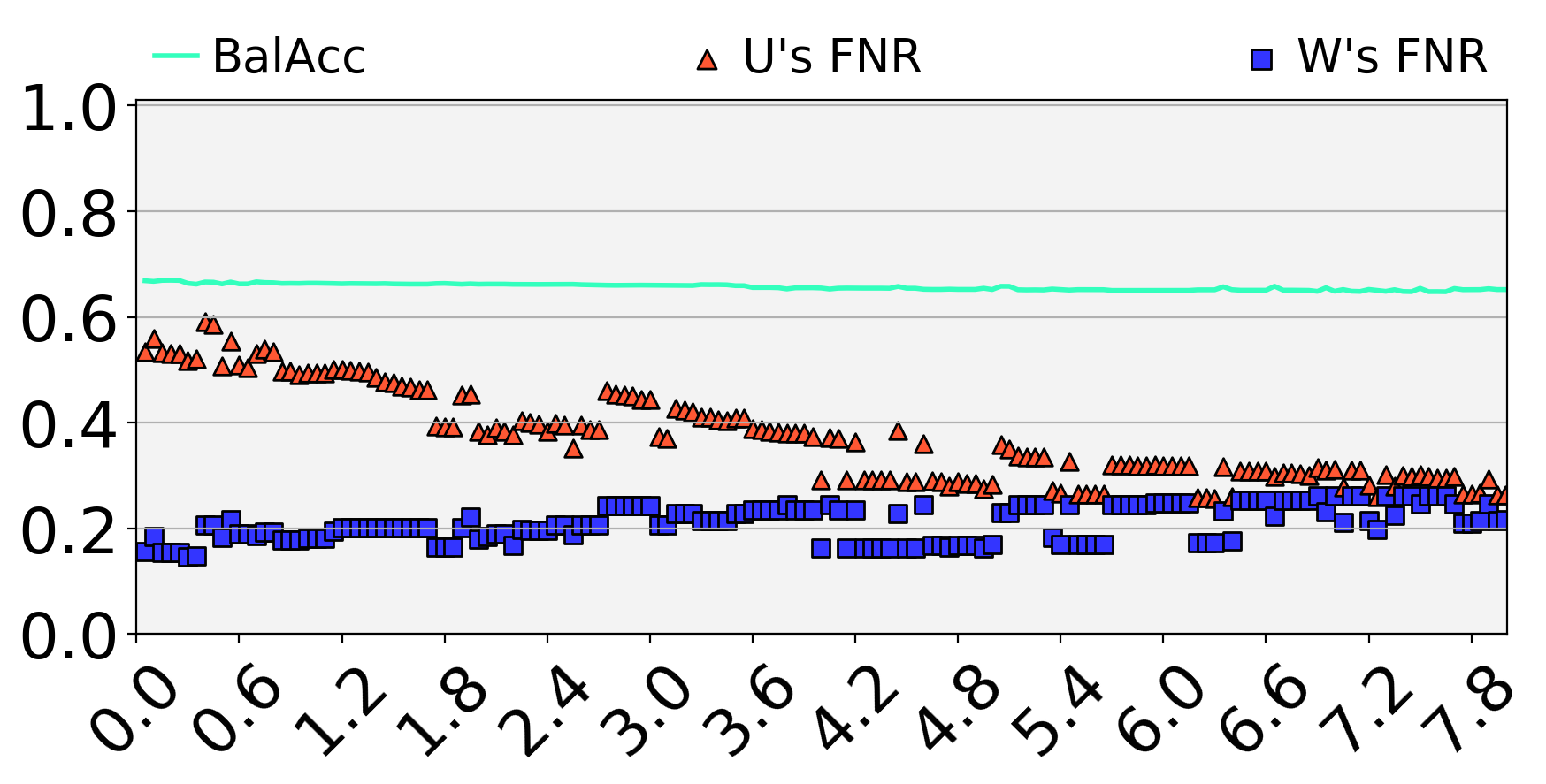}
	\label{fig:lg_scc_fnr}
	}
        \subfloat[\scc targets \eeo by FPR]{\includegraphics[width=0.33\linewidth]{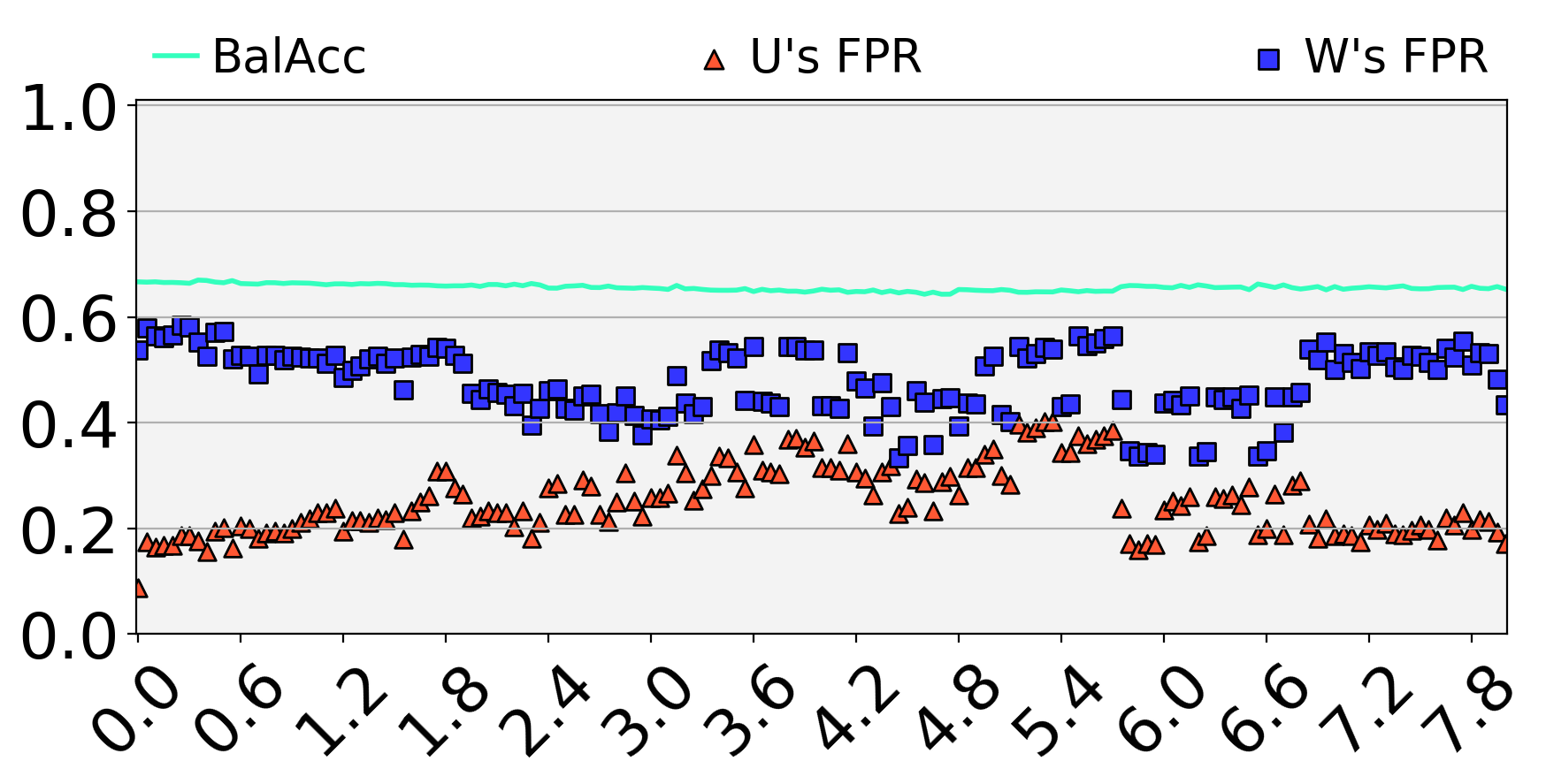}
	\label{fig:lg_scc_fpr}
	}
        \\
        \subfloat[\omn targets \edi by Selection Rate]{\includegraphics[width=0.33\linewidth]{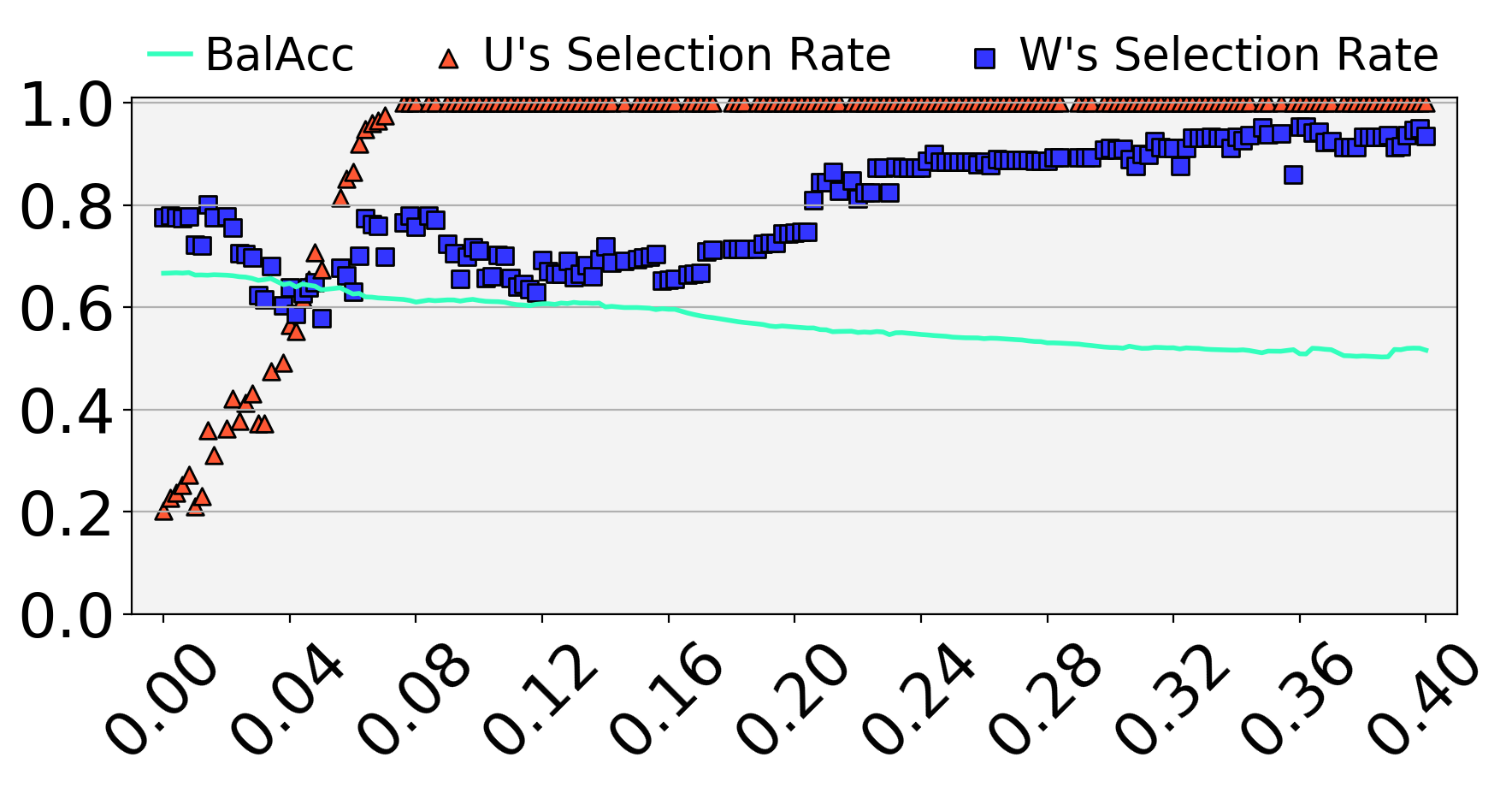}
	\label{fig:lg_omn_di}
	}
	\subfloat[\omn targets \eeo by FNR]         
        {\includegraphics[width=0.33\linewidth]{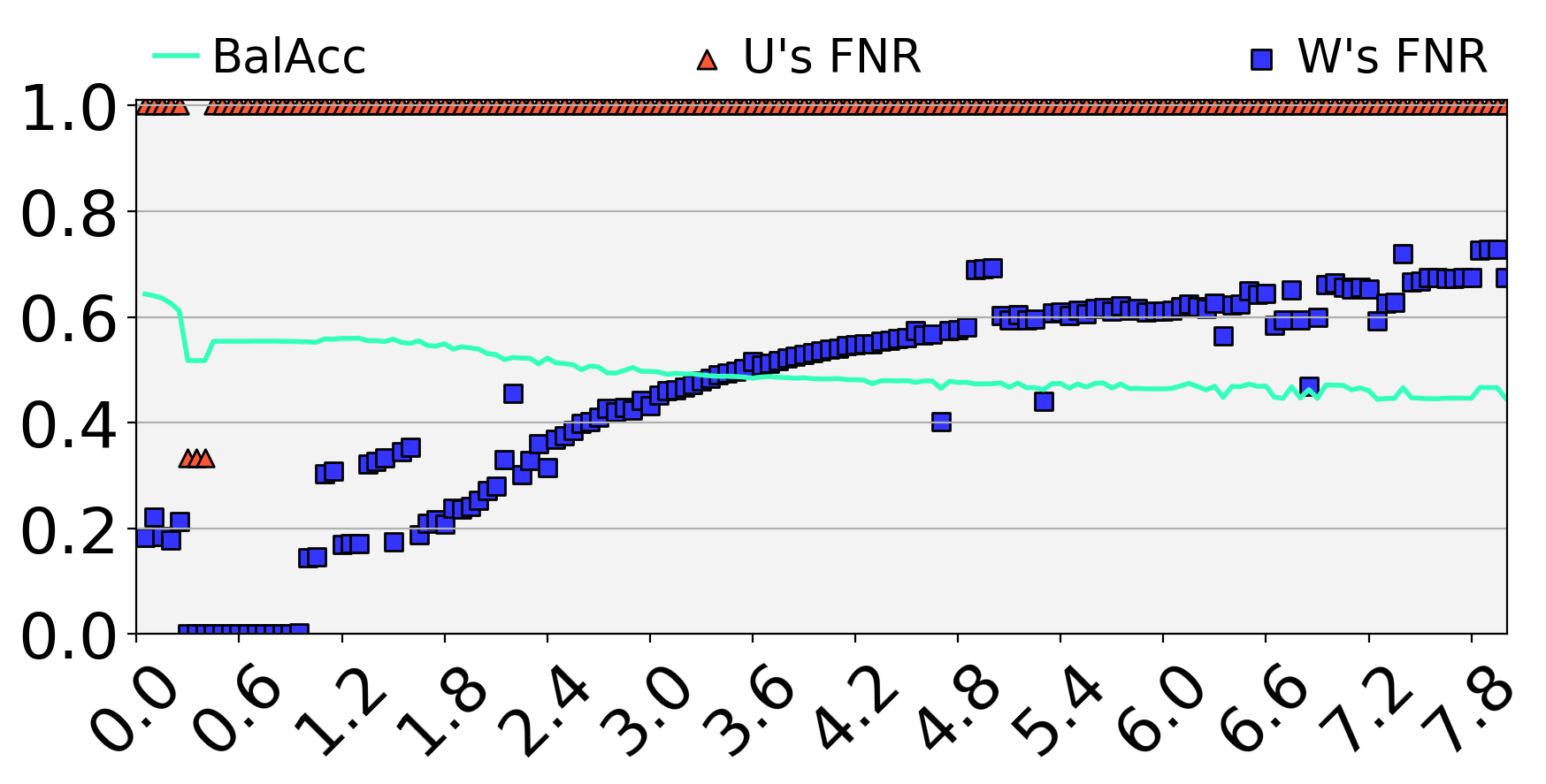}
	\label{fig:lg_omn_fnr}
	}
        \subfloat[\omn targets \eeo by FPR]{\includegraphics[width=0.33\linewidth]{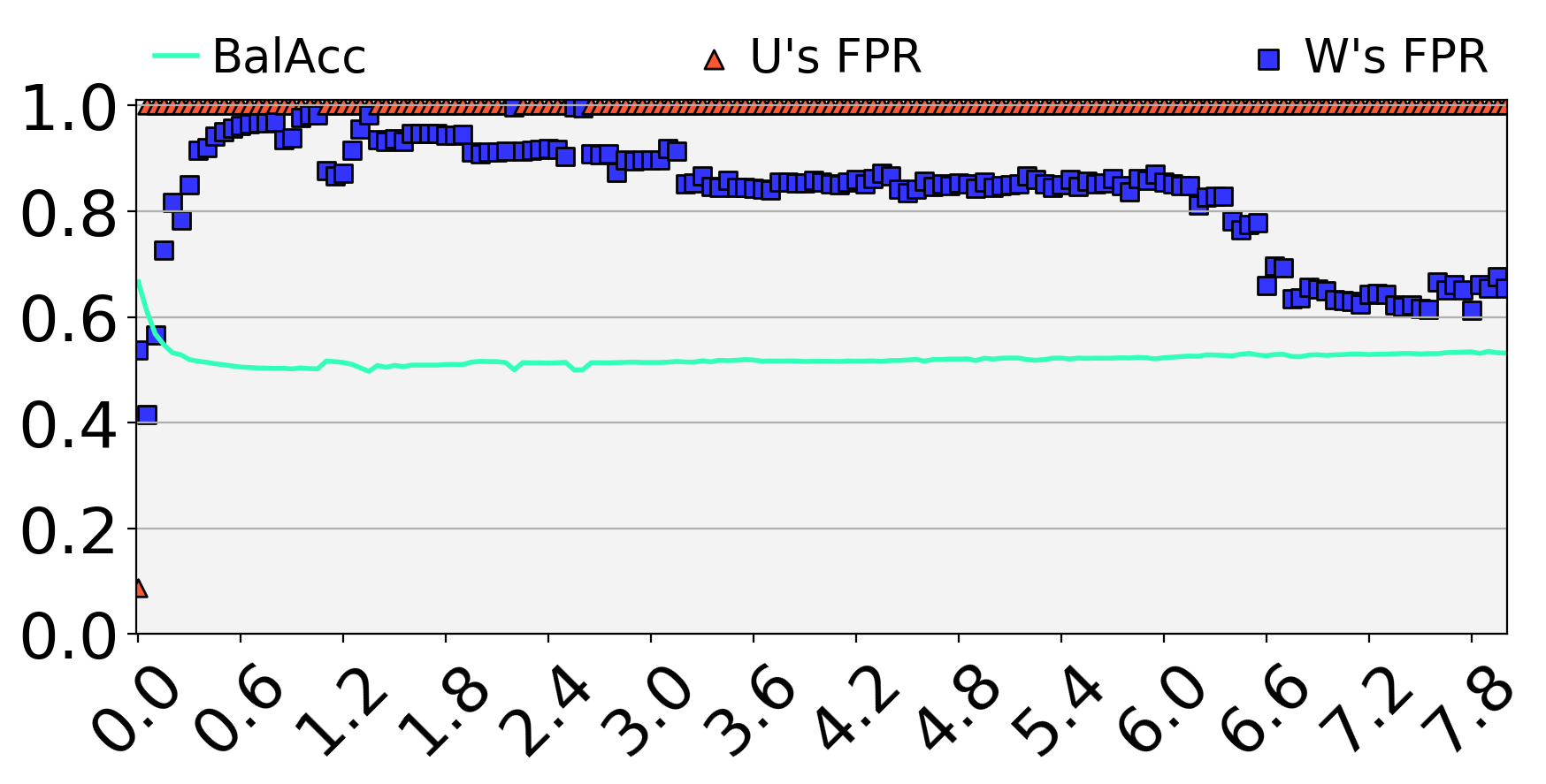}
	\label{fig:lg_omn_fpr}
	}
         \vspace{-0.3em}
	\caption{Comparison of \scc and \omn with respect to the degree of intervention over the \dlg dataset; the setting is otherwise the same as the one shown in Figure~\ref{fig:relation_mp}.} 
        \vspace{-1em}
\label{fig:relation_lg}
\end{figure*}
\noindent \textbf{\scc vs \omn.} 
Figure~\ref{fig:scc_other} repeats the experiment against an alternative state-of-the-art.  Similar to \scc, \omn reweighs input data with an adjustable degree of intervention, but its performance is much less reliable across different datasets.
We observe that \scc significantly outperforms \omn in terms of \edi across all datasets using the XGB models (Fig.~\ref{fig:scc_other_tr_di}). Note that \omn is unable to return a model for the \dcr and \dai datasets in this setting (the corresponding bars are missing in Figures~\ref{fig:scc_other_tr_di}--\ref{fig:scc_other_tr_balacc}).\footnote{The XGB learner fails to converge with the weights produced by \omn.}
Under the LR models, \omn does particularly poorly for the \dcr, \dah, \dae, and \dai datasets (Figures~\ref{fig:scc_other_lr_di} and~\ref{fig:scc_other_lr_aod}), while \scc demonstrates significant improvement over the fairness metrics.

Another important observation is that \omn interventions result in significant loss of accuracy, while \scc produces models with utility that remain on par with that before interventions (Figures~\ref{fig:scc_other_lr_balacc} and~\ref{fig:scc_other_tr_balacc}).
In particular, even cases with apparent improvements in fairness for \omn, come at a problematic utility loss.  For example, in the case of LR models over the \dlg and \dap datasets, \omn shows high improvements in \edi and \ead (Figures~\ref{fig:scc_other_lr_di} and~\ref{fig:scc_other_lr_aod}); however, the resulting models in these cases only predict one class (\ebacc=0.5 with TPR=1 or TNR=1), which renders the models useless.  We indicate these cases with crisscross bars (\inlinegraphics{figs/Extras/crisscross.png}) in the utility graphs (e.g., Figure~\ref{fig:scc_other_lr_balacc}).

We highlight that, despite both being reweighing strategies, \scc and \omn demonstrate vastly different performances. This is due to their distinct weighing methodologies: \omn adjusts its weights according to the model output, while \scc uses tuple conformance to fine-tune the weights, thus achieving a much more robust fairness-utility balance.

\scc and \omn are both \textit{practically} model-agnostic, in the sense that the weights they derive can be used by any learning algorithm.  However, both methods assume a particular model to calibrate their weights.  In the experiment of Fig.~\ref{fig:scc_other}, this calibration was done using the same learner (LR or XGB) as the corresponding experiment.  Figure~\ref{fig:scc_aware} repeats the experiment, but this time each method calibrates its weights assuming a different model than the one eventually trained.  For example, in Figures~\ref{fig:scc_aware_lr_di}--\ref{fig:scc_aware_lr_balacc}, we calibrate the weights of \scc and \omn over each dataset assuming a XGB model, but we subsequently use the weights to train a LR model.  Conversely, in Figures~\ref{fig:scc_aware_tr_di}--\ref{fig:scc_aware_tr_balacc} we assume a LR model for weight calibration, but then train XGB models.

The performance of \scc naturally drops compared to the previous experiments, however, it still maintains robust improvements in fairness across most datasets, while maintaining high utility.  In contrast, \omn becomes less reliable, with inconsistent performance across datasets, and more severe loss of accuracy.  For example, \omn is not able to improve the fairness at all for the XGB models  (Fig.~\ref{fig:scc_aware_tr_aod}) over the \dcr, \dah, \dae, and \dai datasets (corresponding yellow bars have zero height, indicating the maximum difference between groups). This is because the model is not well trained under the input weights (\eg only outputting one type of prediction), resulting in \ebacc lower than 0.5 (Fig.~\ref{fig:scc_aware_tr_balacc}).
Notably, cases that appear to demonstrate fairness gains for \omn (e.g., the \dlg dataset under LR models) come with unacceptable utility loss, producing a model that outputs only one class of predictions (\ebacc=0.5 in Fig.~\ref{fig:scc_aware_lr_balacc}).


\revs{
\scc supports intuitive tuning of the fairness intervention degree (\ie the $\alpha$ parameters), because of the monotonic relationship between the degree of intervention and the fairness adjustment.  
Recall that $\alpha^u$ and $\alpha^w$ represent the additional weights to be assigned to training tuples. \scc assigns such additional weights to the tuples with positive, negative, or both labels, which improves different evaluation metrics (\eg Selection Rate, FNR, and FPR) and optimizes fairness measures such as \edi (computed as the difference in Selection Rate) and \eeo (as the difference in FNR or FPR). 
As we noted in Section~\ref{sec:method_faircc}, \scc only augments the weights of tuples conforming with the identified CCs; this is in contrast to \omn, which augments the weights of all tuples within a group. The latter approach is susceptible to noise in the data, leading to a non-monotonic relationship between the intervention degree and the improvement of fairness.}

\revs{Figures~\ref{fig:relation_mp} and~\ref{fig:relation_lg} compare \scc and \omn while varying the degree of intervention (\ie parameters $\alpha^u$ for \scc and $\lambda$ for \omn).  We evaluate the methods across three fairness metrics (\edi, \eeo by FNR, and \eeo by FPR) and two real-world datasets, \dmp and \dlg. We note that, as the intervention degree increases, \scc reduces the difference between groups over the corresponding metrics, and, under the optimal intervention degree, no difference is observed (\ie blue squares and red triangles overlap). This monotonic behavior, which is consistent across datasets and fairness measures, allows for tuning the intervention degree effectively and efficiently.
In contrast, the behavior of \omn is more erratic with respect to its intervention parameter $\lambda$ for \edi, while the method does poorly for \eeo by FNR and FPR, \ie not minimizing the difference in the metrics between groups and not returning reasonable models for the minority group with most of the interventions (the red triangles equal to 1 in Figures~\ref{fig:mp_omn_fnr},~\ref{fig:mp_omn_fpr},~\ref{fig:lg_omn_fnr}, and~\ref{fig:lg_omn_fpr}).}


In summary, despite the premise of being model-agnostic, \omn demonstrates high model dependence, as its ability to impart fairness improvements is severely hindered when its weights are not calibrated using the \enquote{right} model. \revs{Further, its non-monotonic performance with respect to the degree of intervention makes tuning that parameter challenging.}
In contrast, \scc demonstrates robustness, as its weights are primarily driven by conformance over the training data and not the model output, \revs{while its consistent and monotonic behavior with respect to the degree of intervention allows for flexible and effective tuning of that parameter to meet application needs.}

\noindent \textbf{\scc vs \capu.} \capu is an invasive intervention that alters the input data to improve the fairness of ML models. 
Figures~\ref{fig:scc_other_tr_di}--\ref{fig:scc_other_tr_balacc} demonstrate the performance of \capu across all 7 datasets over XGB models.
We see that \scc significantly outperforms \capu in improving \edi over the \dmp, \dlg, and \dap datasets. These gains remain present but are more modest concerning \ead.  The two methods have similar performance across the rest of the datasets, and maintain similar high utility.  We need to highlight that \scc achieves performance on par with and often better than \capu, \emph{while remaining non-invasive.}  This distinction is significant, as invasive methods are naturally poised to achieve greater fairness improvements, simply due to the flexibility that data changes can afford them.  Nevertheless, \scc outperforms \capu, while also avoiding the potential issues of invasive approaches, such as introducing unintended drift in the data.

\smallskip
\begin{mdframed}[everyline=true,innertopmargin = 1mm,innerleftmargin = 1mm]
\textit{\textbf{Key takeaways:}} \scc outperforms prior art in improving the fairness of ML models, while maintaining high utility. It shows clear and consistent gains compared to other reweighing methods, and achieves on-par or better performance compared to invasive alternatives.  It further stays robust when using learners different from those used to calibrate its weights, thus being effectively model-agnostic.
\end{mdframed}

\input{exp_cond}

%% file: figs/tab-datasets.tex

\begin{figure*}
\resizebox{0.99\textwidth}{!}{
	\begin{tabular}{lccccccc}
	    \toprule
    	\textbf{dataset} &  \dmp & \dlg & \dcr & \textit{ACSP}  & \textit{ACSH} &  \textit{ACSE} & \textit{ACSI}   \\   
	    \midrule
     \rowcolor{lightgray}
	    \textbf{size} & 15,675 & 24,479 & 120,269 & 86, 600 & 250, 847 & 250, 847 & 250, 847 \\  
	    \makecell[l]{\textbf{\# of attributes}\\ \textbf{numerical / categorical}} & 6 \big/ 34 & 6 \big/ 4 & 6 \big/ 0 & 4 \big/ 14 & 4 \big/ 21  & 4 \big/ 11  & 6 \big/ 13   \\ 
     \rowcolor{lightgray}
      \textbf{minority group U}
      & non-White & African-American & age $<$\ 35 & African-American           & African-American          & African-American                         & African-American           \\ 
      \textbf{population of U}
      & 61.6\% & 7.7\% & 13.7\% & 9.2\%  & 7.3\% & 7.3\%  & 7.3\%   \\ 
      \rowcolor{lightgray}
      \textbf{\% positive labels in U}
      & 11.4\% & 56.6\% & 10.7\% & 48.3\% & 9.3\%   & 39.3\%   & 40.2\% \\ 
      \textbf{predictive task}
      & \makecell[c]{high hospital \\utilization} & passing bar exam & \makecell[c]{serious delay \\ in 2 years} & \makecell[c]{covered by private \\ insurance companies} & \makecell[c]{having health\\ insurance} & employment & \makecell[c]{income poverty \\ rate $<$ 250}  \\
		\bottomrule
	\end{tabular}}
     \vspace{-3pt}
	    \caption{Summary statistics and main aspects of the 7 real-world datasets used in our experiments. }
     \vspace{-1em}
        \label{tab:datasets}
\end{figure*}

%% file: exp_cond.tex

\subsection{Evaluation of \mcc}
\label{sec:exp:mcc}
In this section, we contrast \mcc and \scc, highlighting scenarios where \mcc is the preferable strategy.  Intuitively, as a single-model approach, \scc is more generally applicable, and, as we showed in Section~\ref{sec:exp:scc}, performs well across our real-world datasets.  By design, it can effectively address inter-group drift that may not be obvious in the data.  In contrast, \mcc can more effectively address cases of significant drift across groups, where a single-model strategy is unlikely to be able to derive an effective single model.  We simulate these scenarios with synthetic data to highlight this strength of \mcc. We proceed to describe the synthetic data generation next.

\begin{figure}[t]
    \centering
    \includegraphics[width=0.4\textwidth]{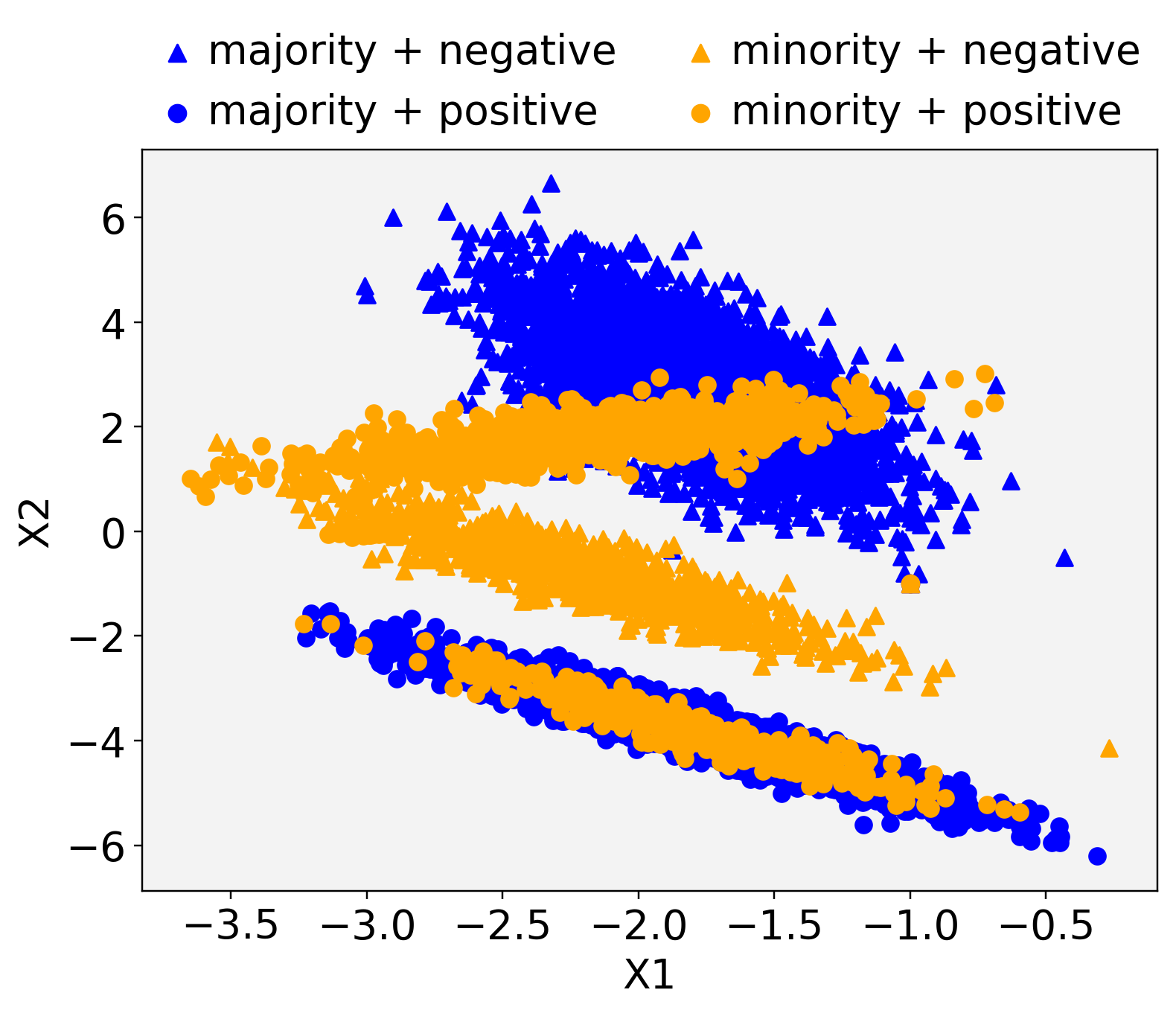}
    \caption{A synthetic dataset \textit{Syn1} containing two groups: majority and minority color-coded in blue and orange, respectively. The attributes $X_1$ and $X_2$ show dissimilar distributions, indicating a significant drift over groups.}
    \label{fig:syn_example}
\end{figure}

\rev{We generate synthetic data with $N=11,000$, with $8,000$ majority and $3,000$ minority elements, and $50\%$ positive and $50\%$ negative labels within each group.} 
\rev{We generate five synthetic datasets using the \textit{make\_classification} function 
from the scikit-learn library~\cite{scikit-learn}.
In the synthetic datasets, the majority and minority groups are distributed over similar areas of the space, with their positive and negative labels following dissimilar distributions, making the generation of a single model extremely challenging.} \revs{Figure~\ref{fig:syn_example} shows an example of our synthetic datasets.}


\begin{figure*}
	\centering
	\subfloat[Disparate Impact (\edi), LR models]{\includegraphics[width=0.33\linewidth]{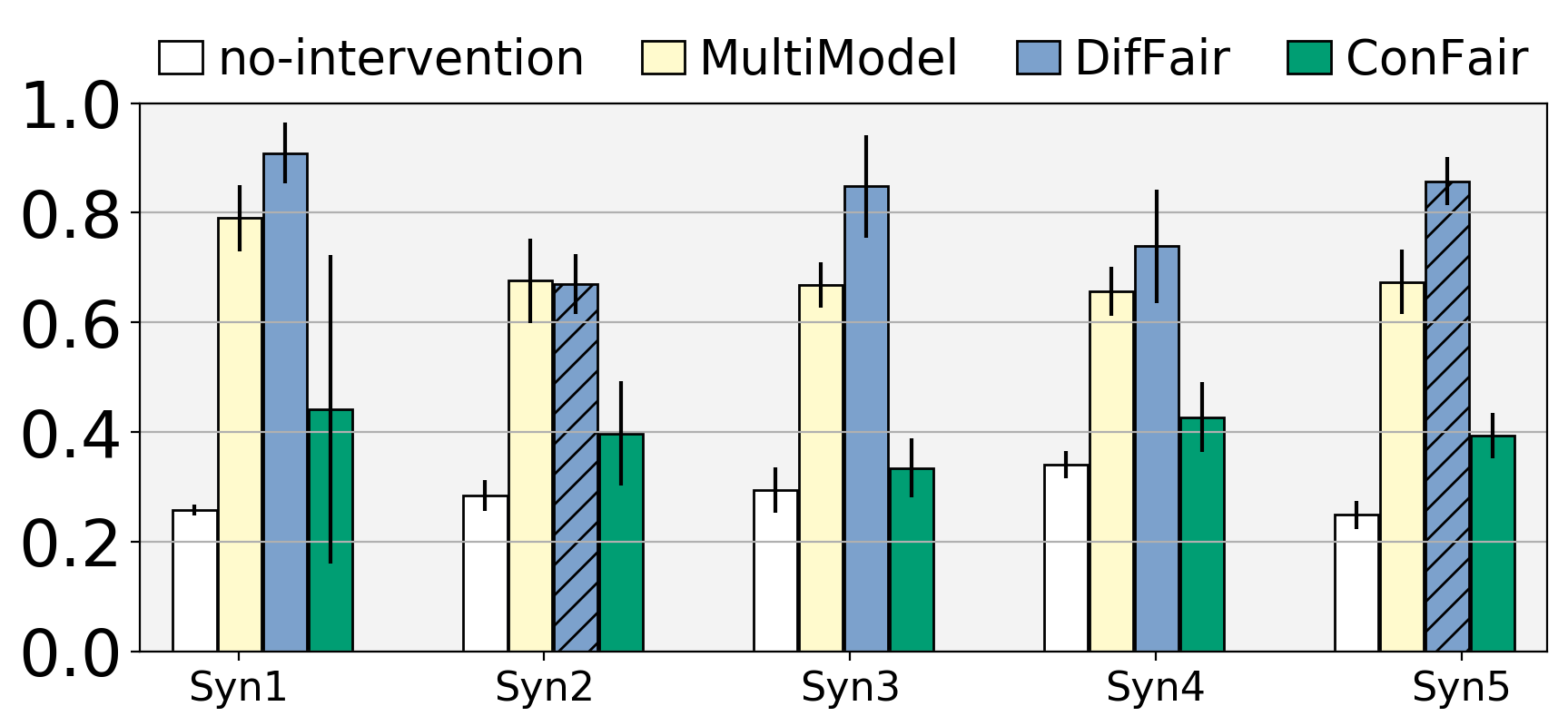}
	\label{fig:mcc_syn_lr_di}
	}
	\subfloat[Average Odds Difference (\ead), LR models]{\includegraphics[width=0.33\linewidth]{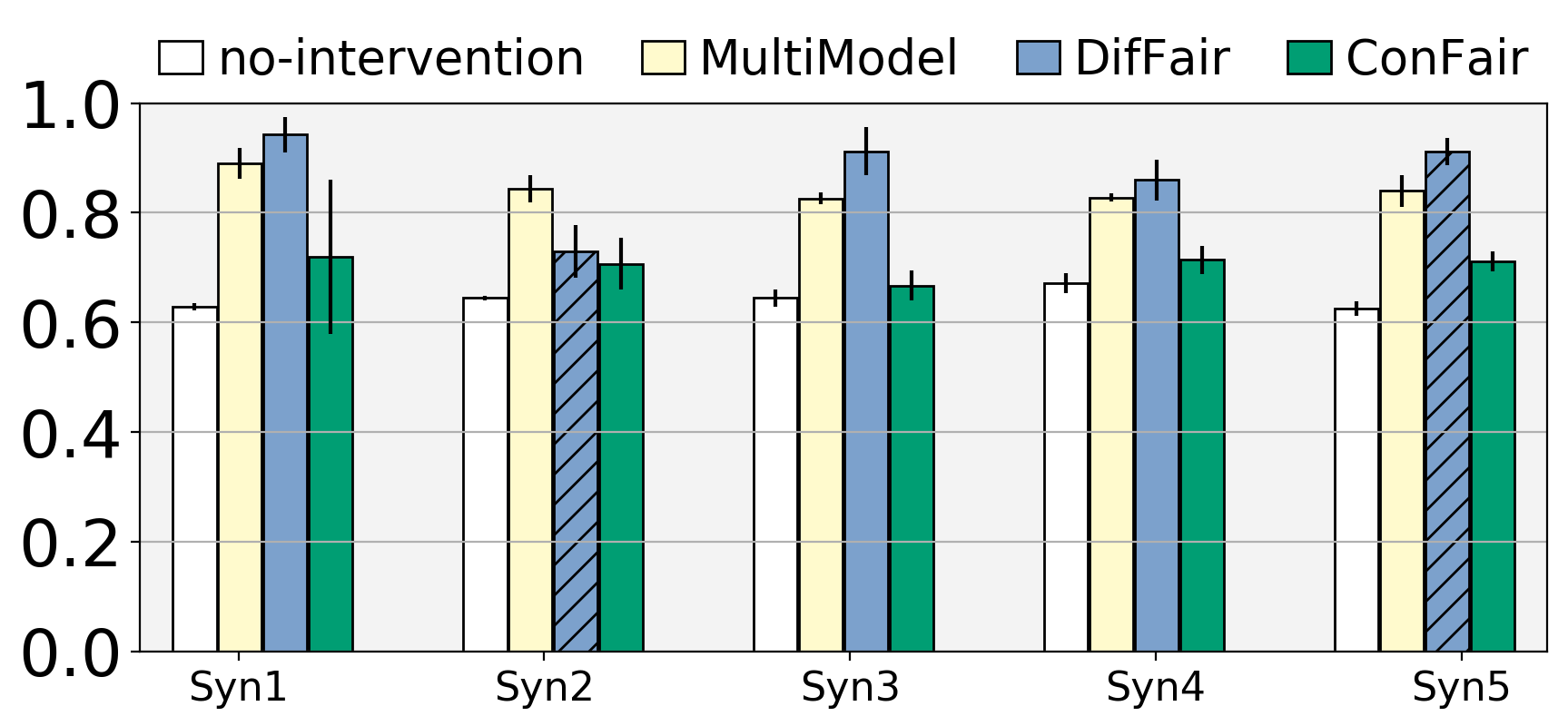}
	\label{fig:mcc_syn_lr_aod}
	}
        \subfloat[Balanced Accuracy (\ebacc), LR models]{\includegraphics[width=0.33\linewidth]{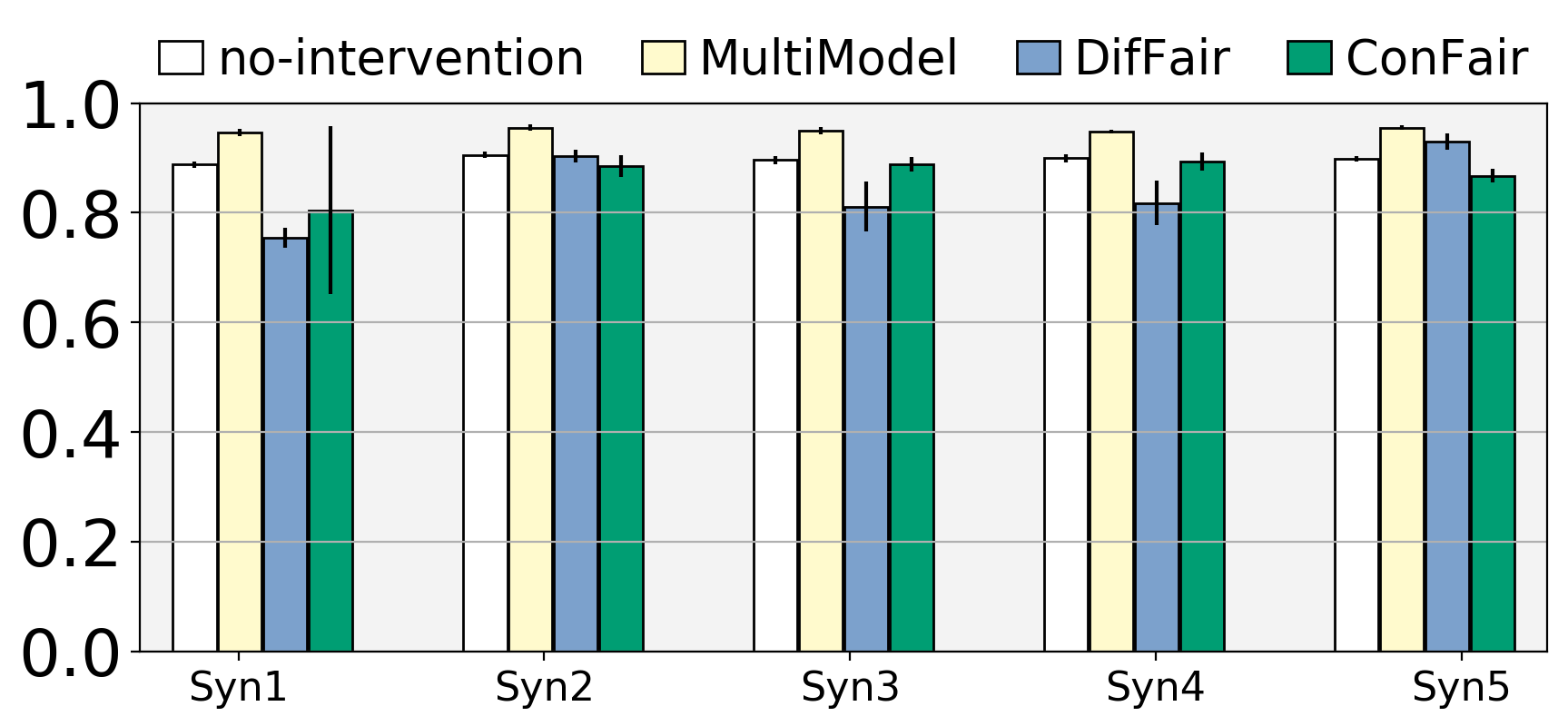}
	\label{fig:mcc_syn_lr_balacc}
	}
         \vspace{-0.3em}
	\caption{\mcc can produce stronger fairness outcomes compared to \scc in cases of significant drift that is simulated in synthetic datasets.} 
        \vspace{-2em}
\label{fig:mcc_syn}
\end{figure*}
\begin{figure*}
	\centering
	\subfloat[Disparate Impact (\edi), LR models]{\includegraphics[width=0.33\linewidth]{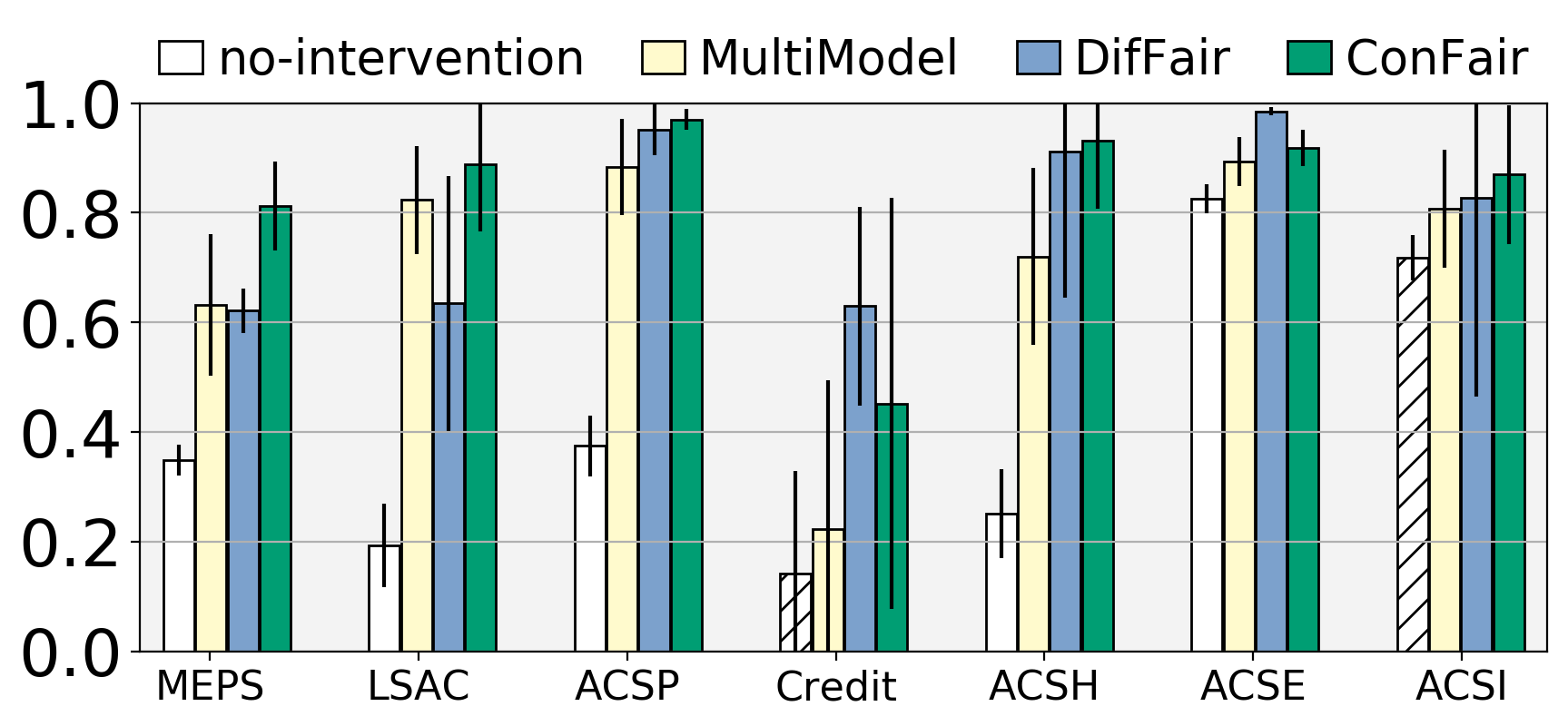}
	\label{fig:mcc_real_lr_di}
	}
	\subfloat[Average Odds Difference (\ead), LR models]{\includegraphics[width=0.33\linewidth]{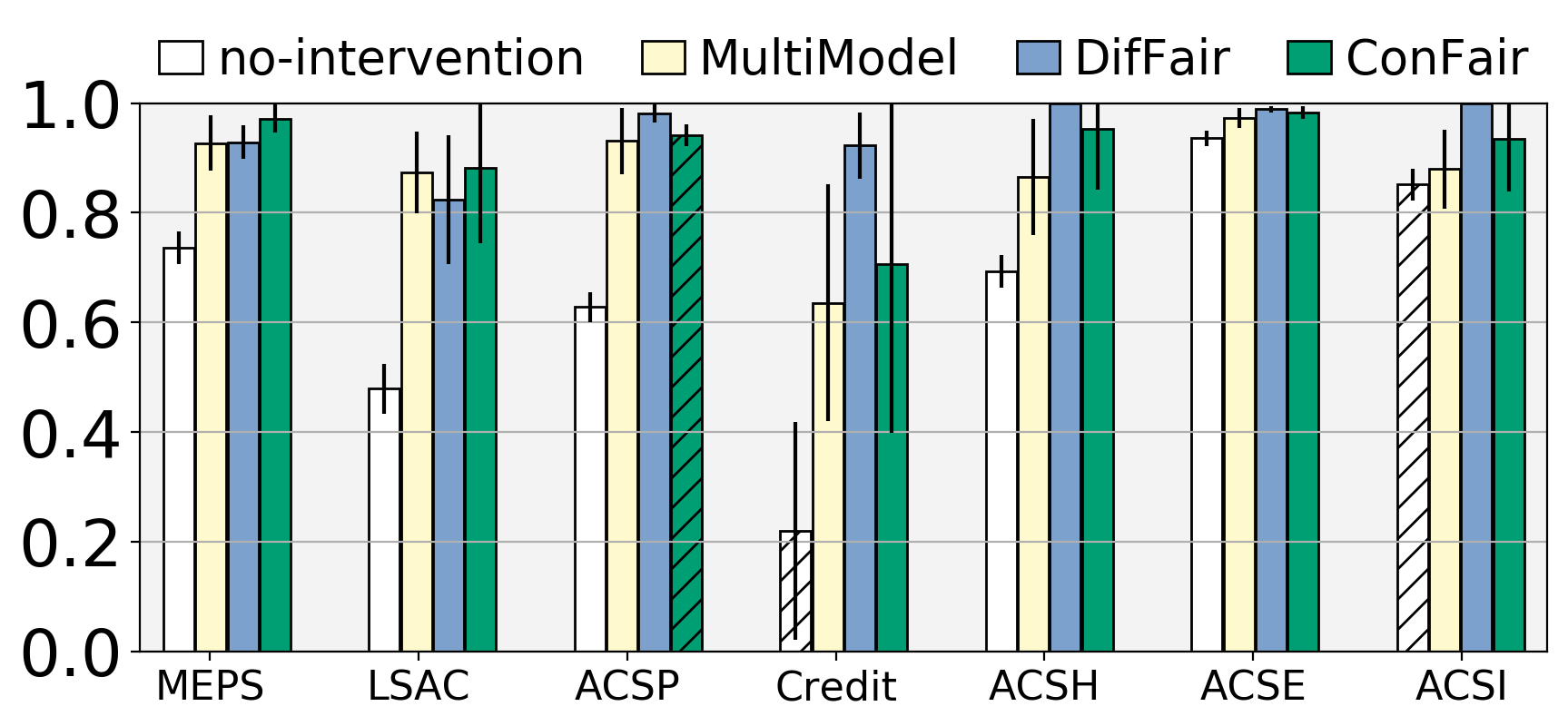}
	\label{fig:mcc_real_lr_aod}
	}
        \subfloat[Balanced Accuracy (\ebacc), LR models]{\includegraphics[width=0.33\linewidth]{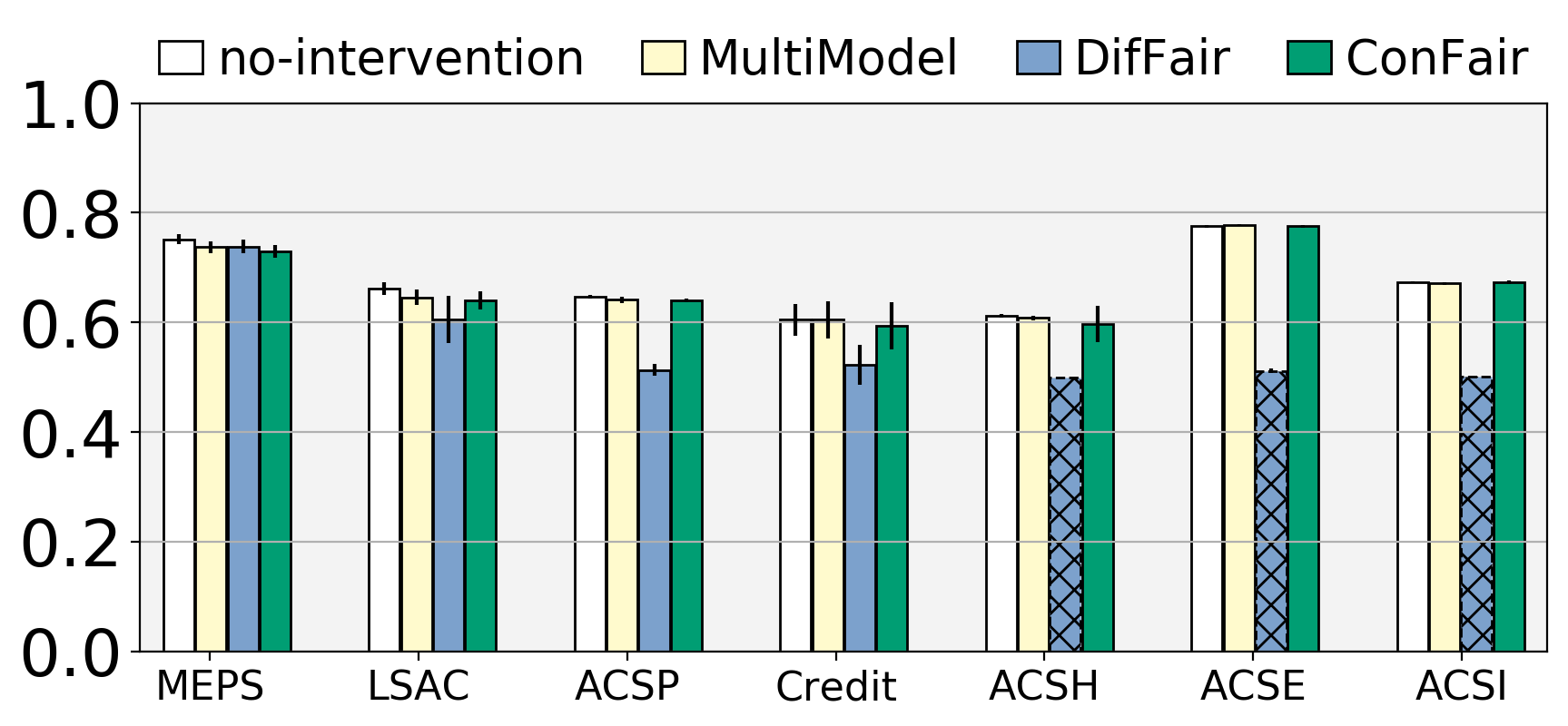}
	\label{fig:mcc_real_lr_balacc}
	}
        \\
        \subfloat[Disparate Impact (\edi), XGB models]{\includegraphics[width=0.33\linewidth]{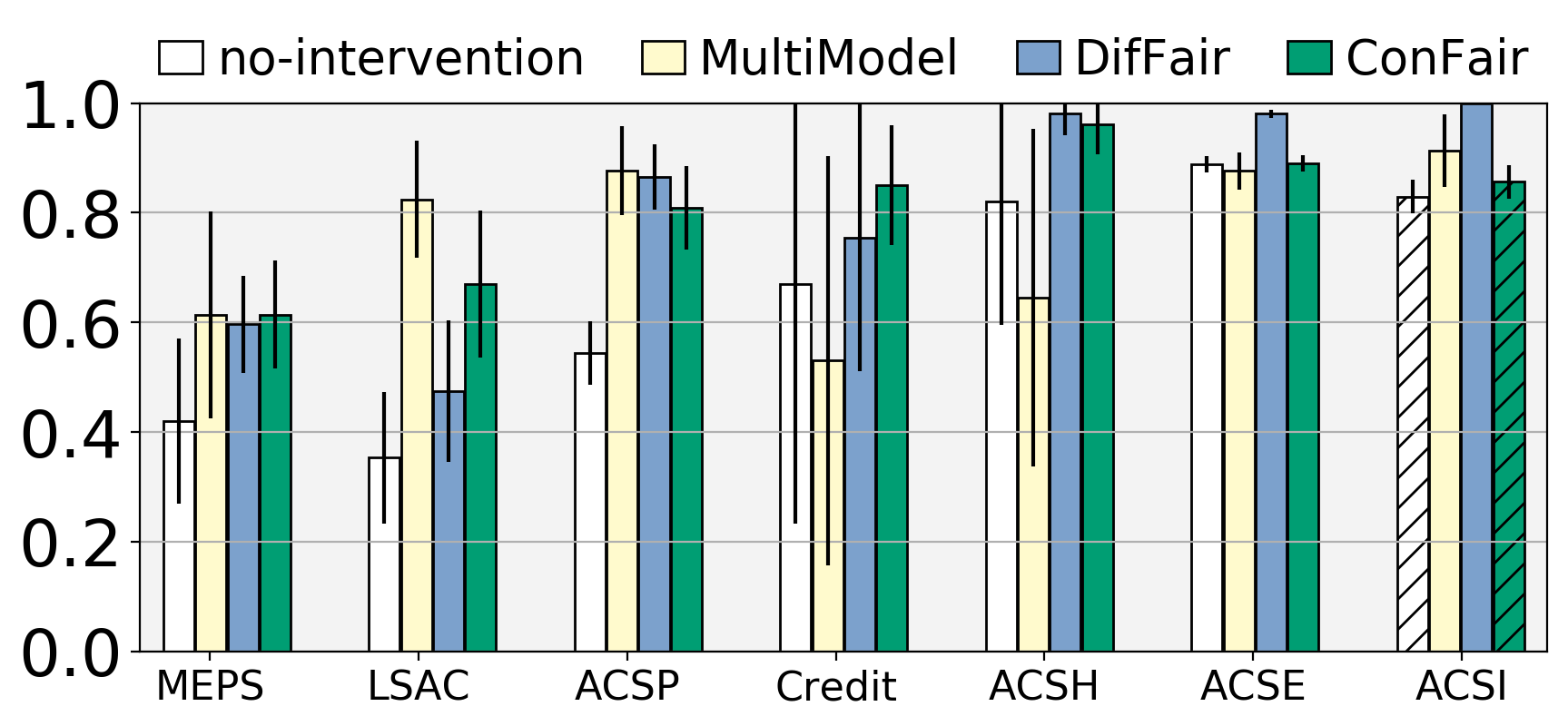}
	\label{fig:mcc_real_tr_di}
	}
	\subfloat[Average Odds Difference (\ead), XGB models]         
        {\includegraphics[width=0.33\linewidth]{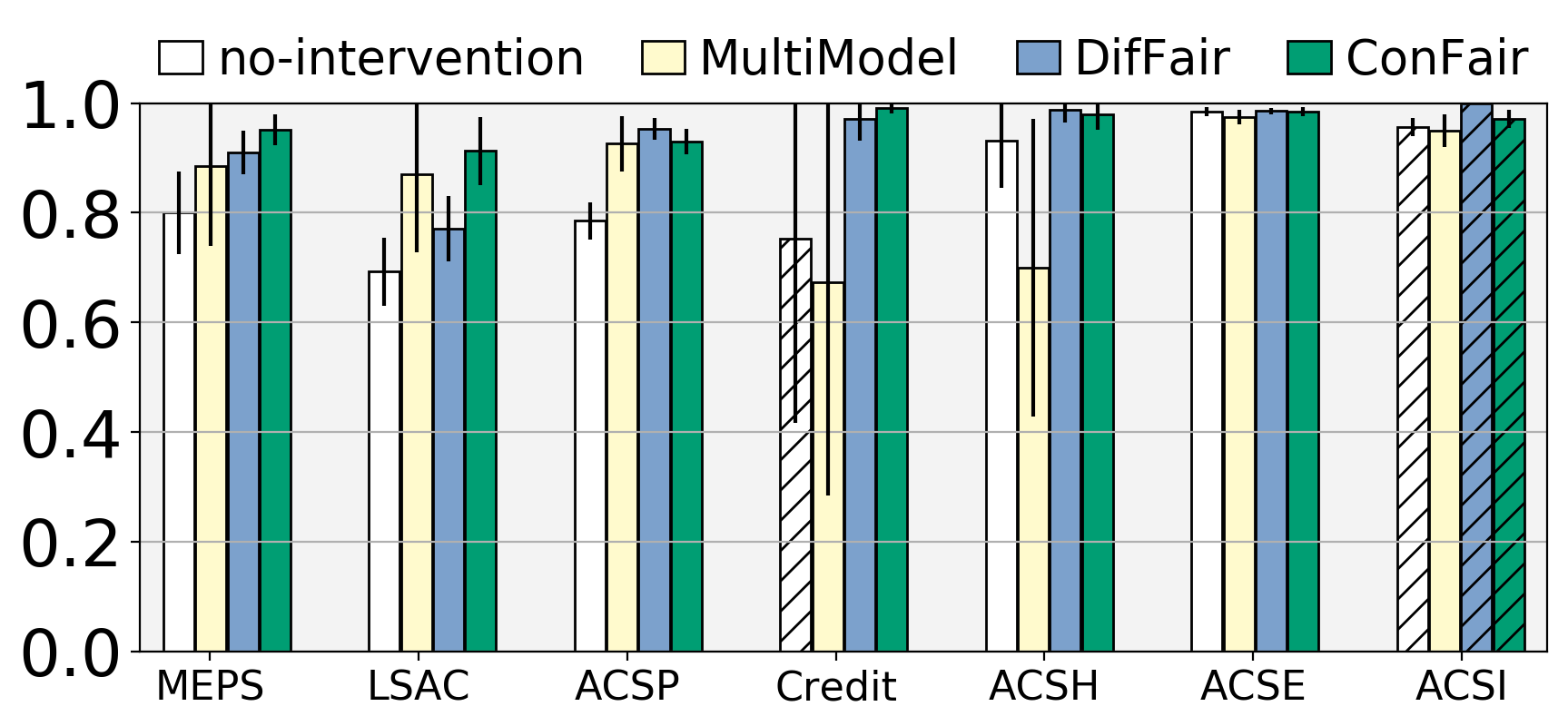}
	\label{fig:mcc_real_tr_aod}
	}
        \subfloat[Balanced Accuracy (\ebacc), XGB models]{\includegraphics[width=0.33\linewidth]{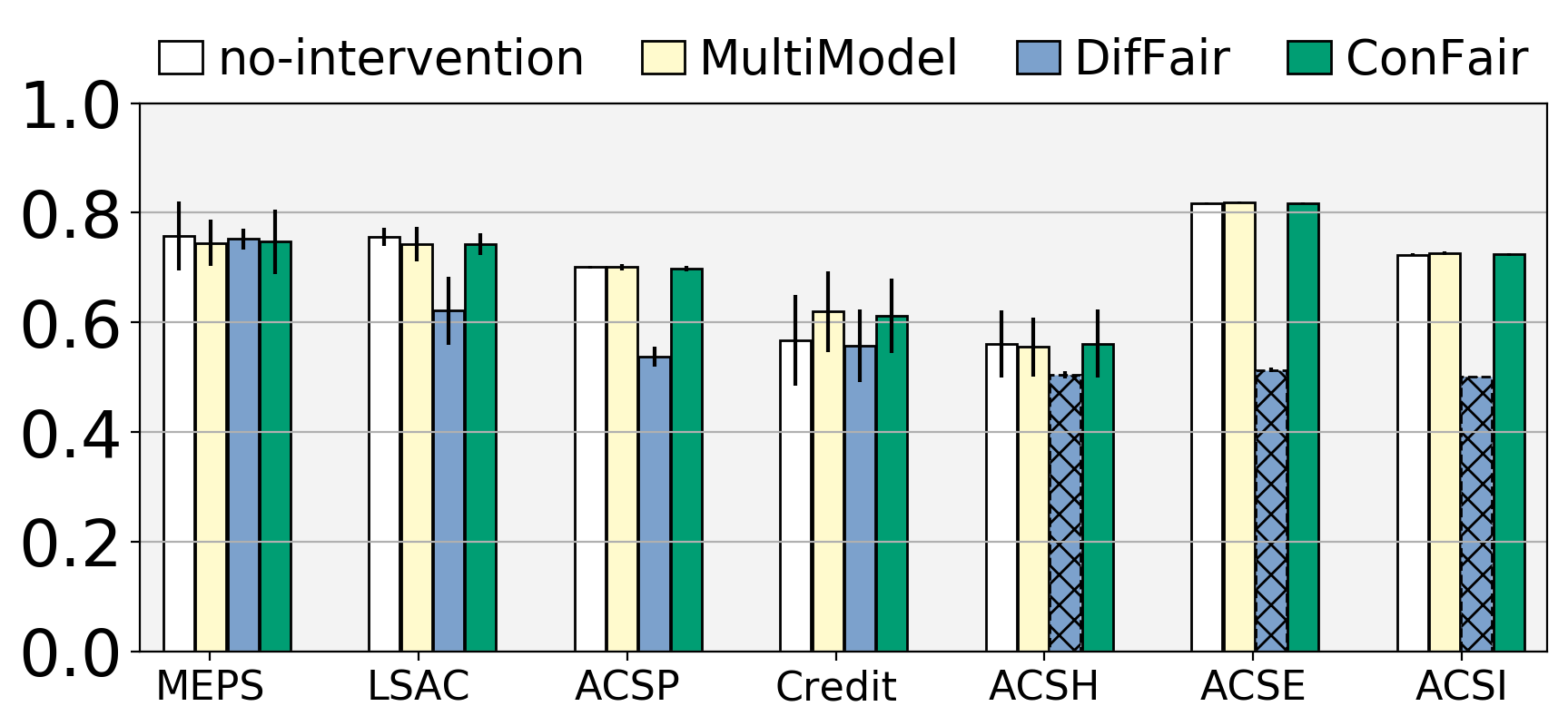}
	\label{fig:mcc_real_tr_balacc}
	}
         \vspace{-0.3em}
	\caption{\mcc is comparable to \scc in most real-world datasets.} 
        \vspace{-2em}
\label{fig:mcc_real}
\end{figure*}

\begin{figure*}
	\centering
	\subfloat[Disparate Impact (\edi), LR models]{\includegraphics[width=0.33\linewidth]{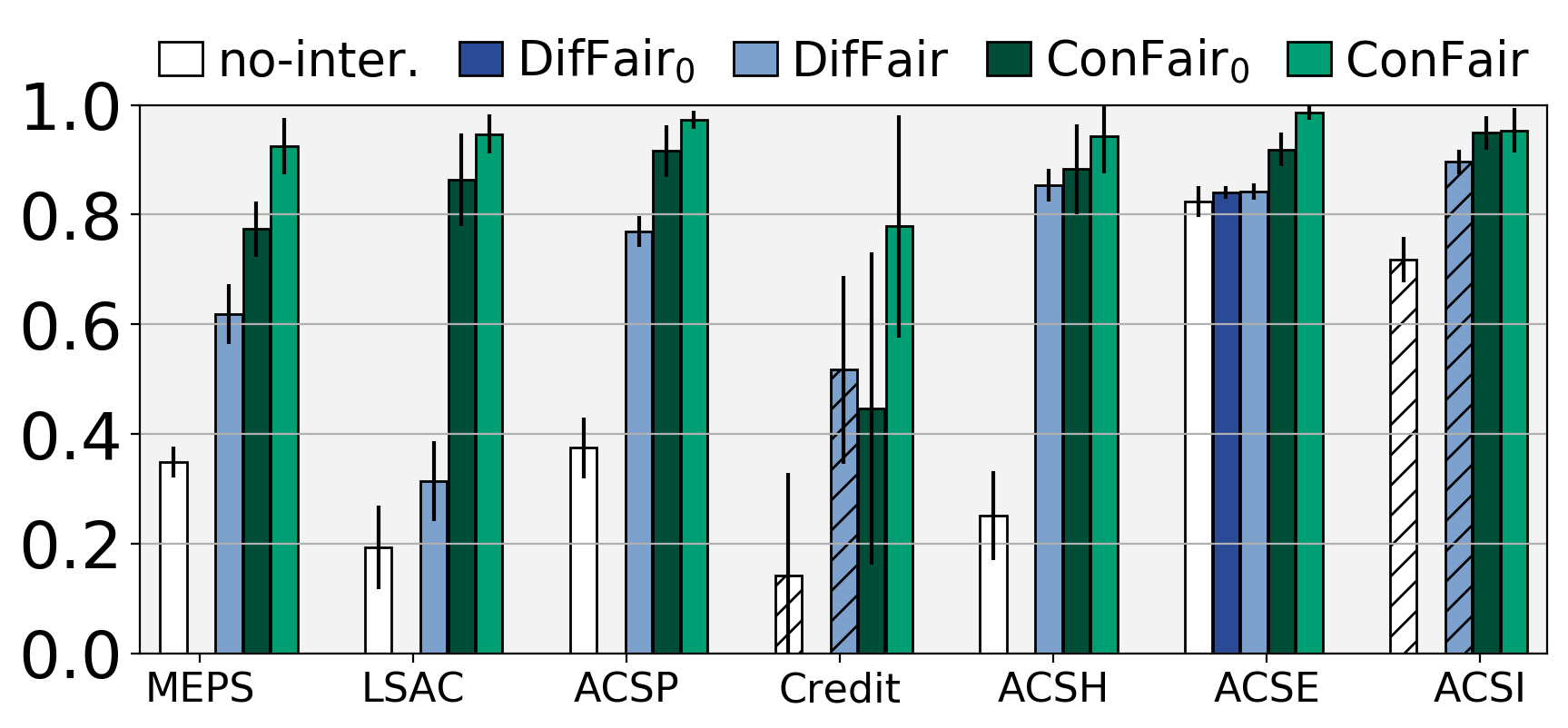}
	\label{fig:cc_opt_lr_di}
	}
	\subfloat[Average Odds Difference (\ead), LR models]{\includegraphics[width=0.33\linewidth]{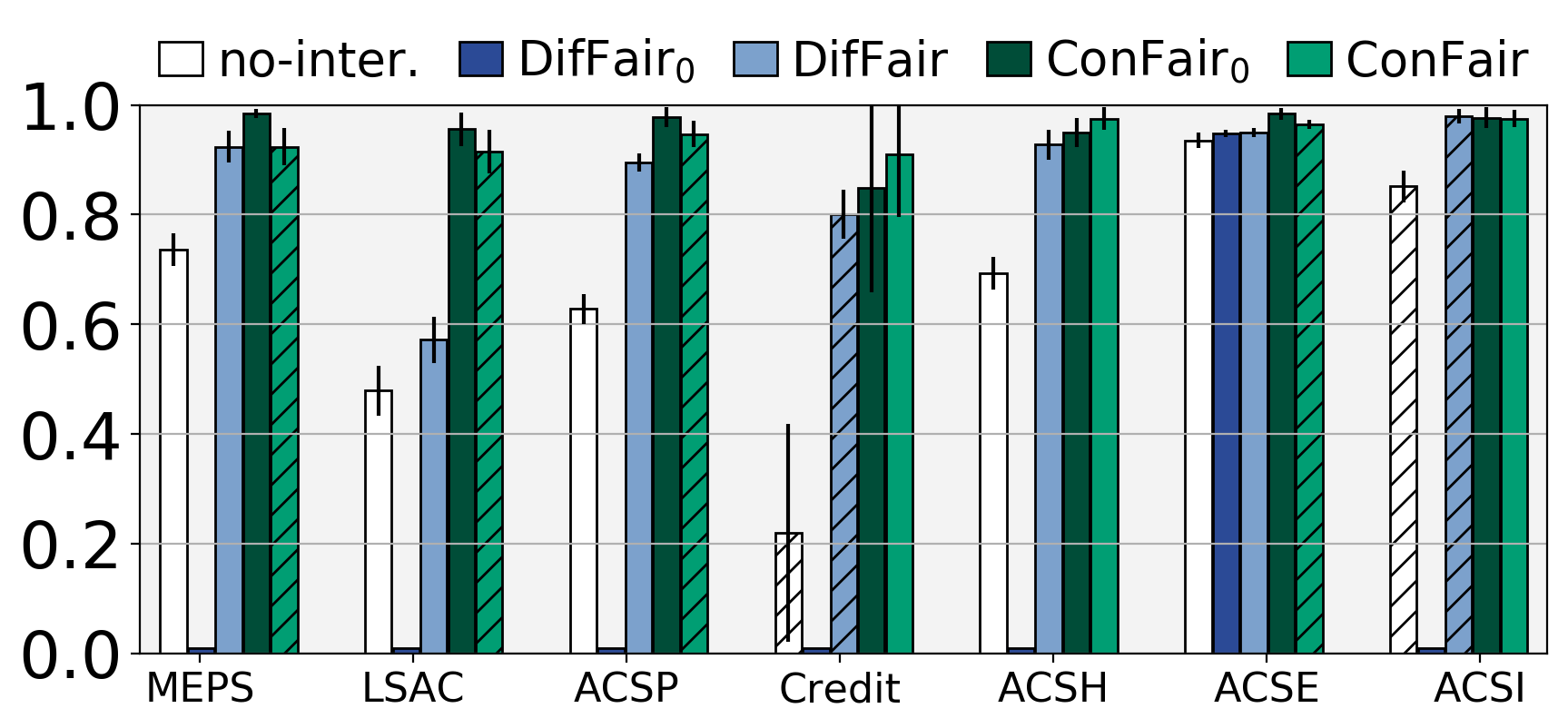}
	\label{fig:cc_opt_lr_aod}
	}
        \subfloat[Balanced Accuracy (\ebacc), LR models]{\includegraphics[width=0.33\linewidth]{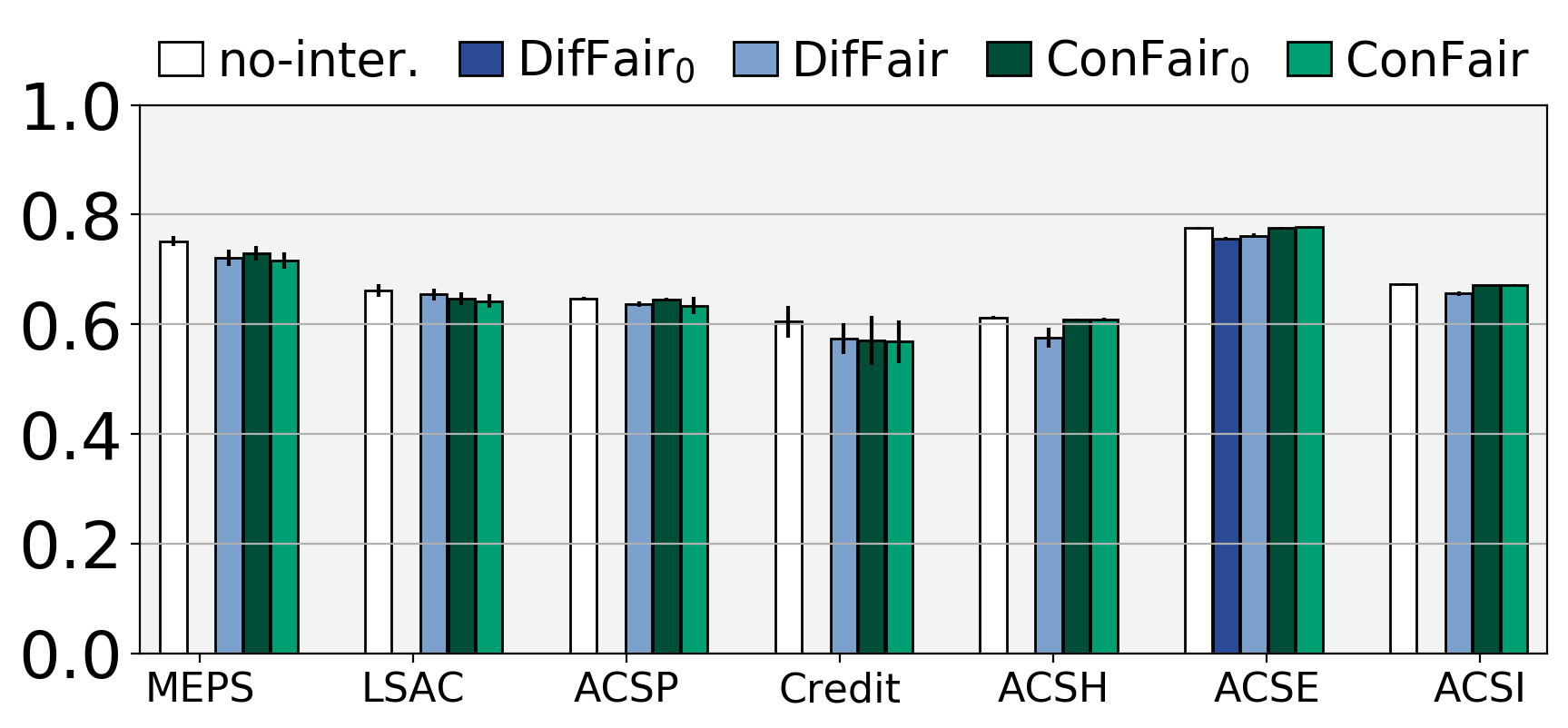}
	\label{fig:cc_opt_lr_balacc}
	}
        \\
        \subfloat[Disparate Impact (\edi), XGB models]{\includegraphics[width=0.33\linewidth]{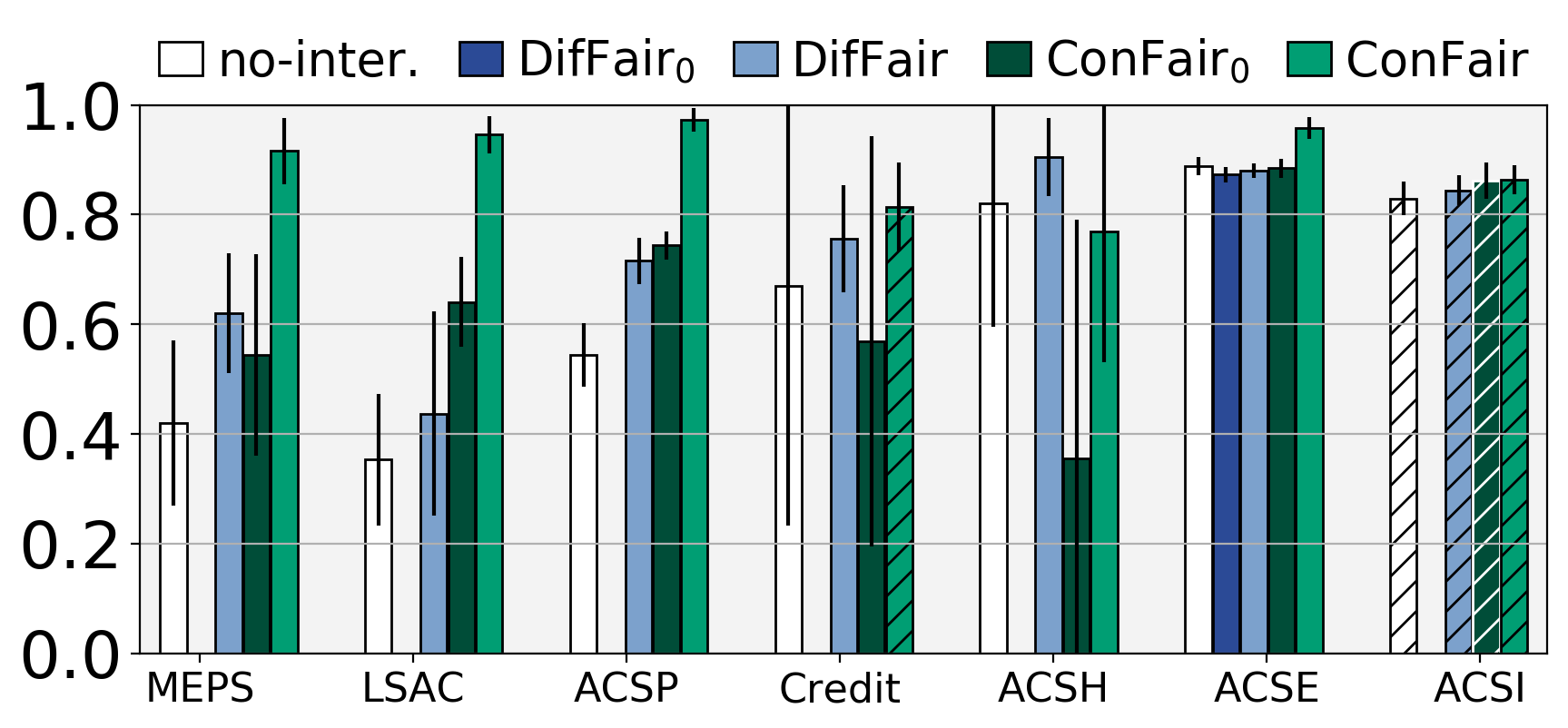}
	\label{fig:cc_opt_tr_di}
	}
	\subfloat[Average Odds Difference (\ead), XGB models]         
        {\includegraphics[width=0.33\linewidth]{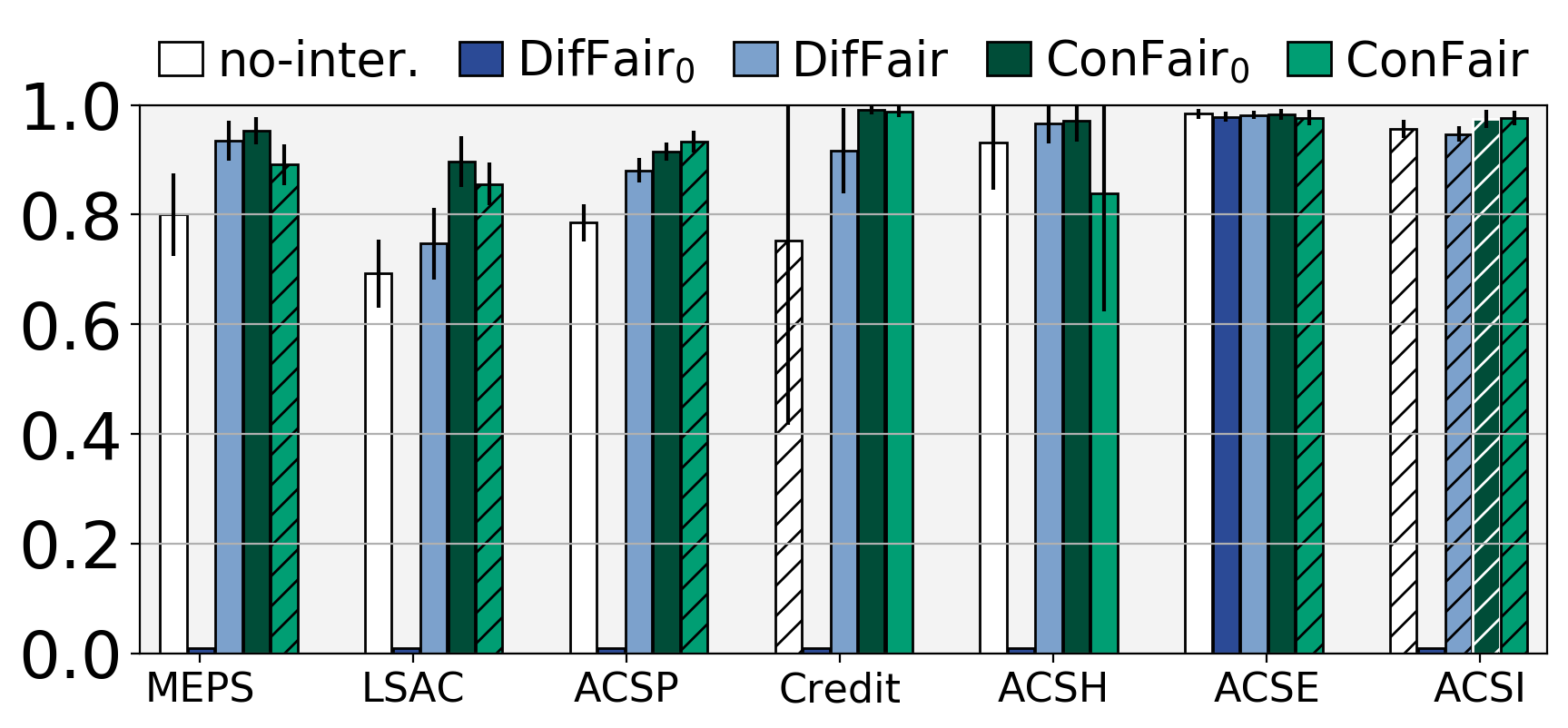}
	\label{fig:cc_opt_tr_aod}
	}
        \subfloat[Balanced Accuracy (\ebacc), XGB models]{\includegraphics[width=0.33\linewidth]{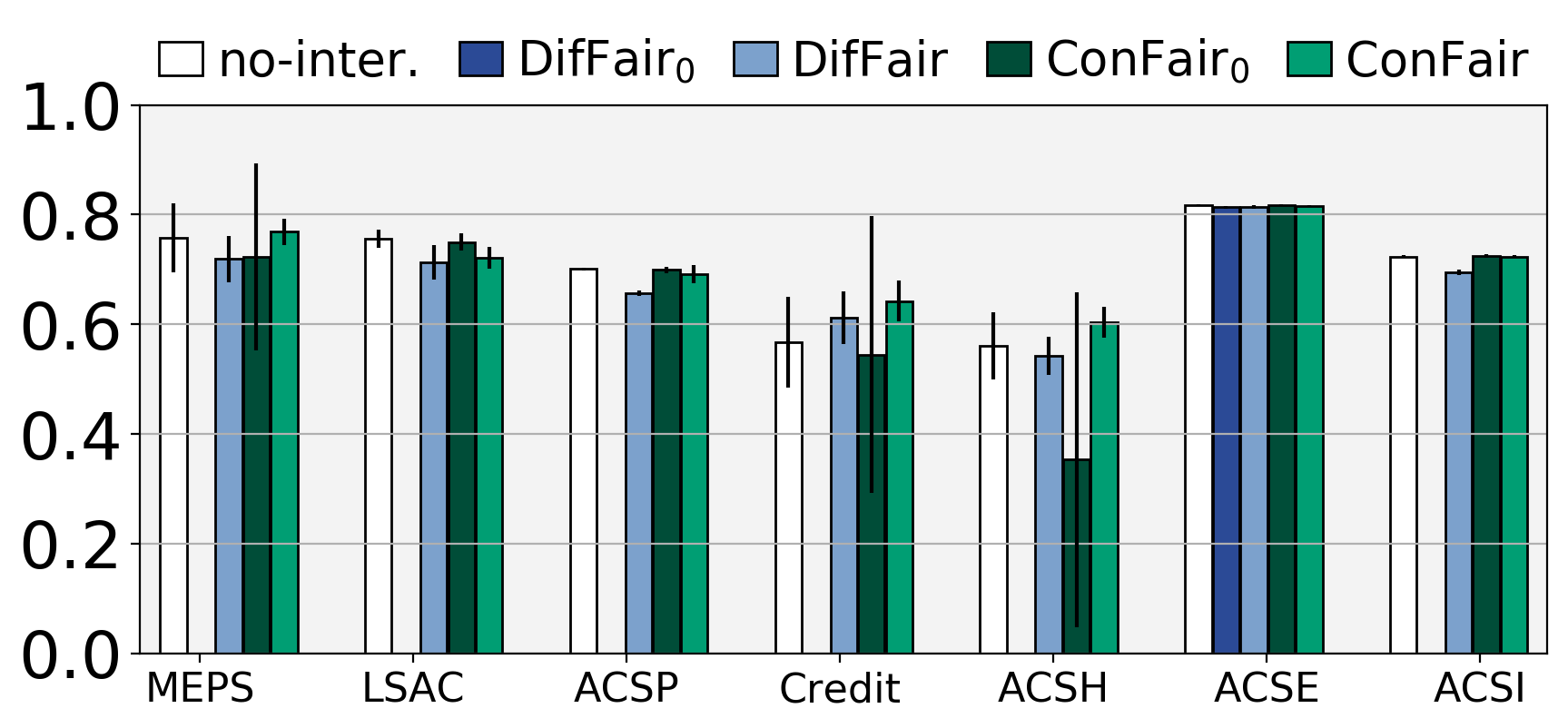}
	\label{fig:cc_opt_tr_balacc}
	}
         \vspace{-0.3em}
	\caption{The density-based optimization is essential in the performance of \mcc and \scc. Variants {\mcc}$_0$ and {\scc}$_0$ that don't optimize CCs have lower effectiveness.} 
        \vspace{-2em}
\label{fig:cc_opt}
\end{figure*}
\begin{figure}
    \centering
    \subfloat[run-time, LR models]{\includegraphics[width=0.9\linewidth]{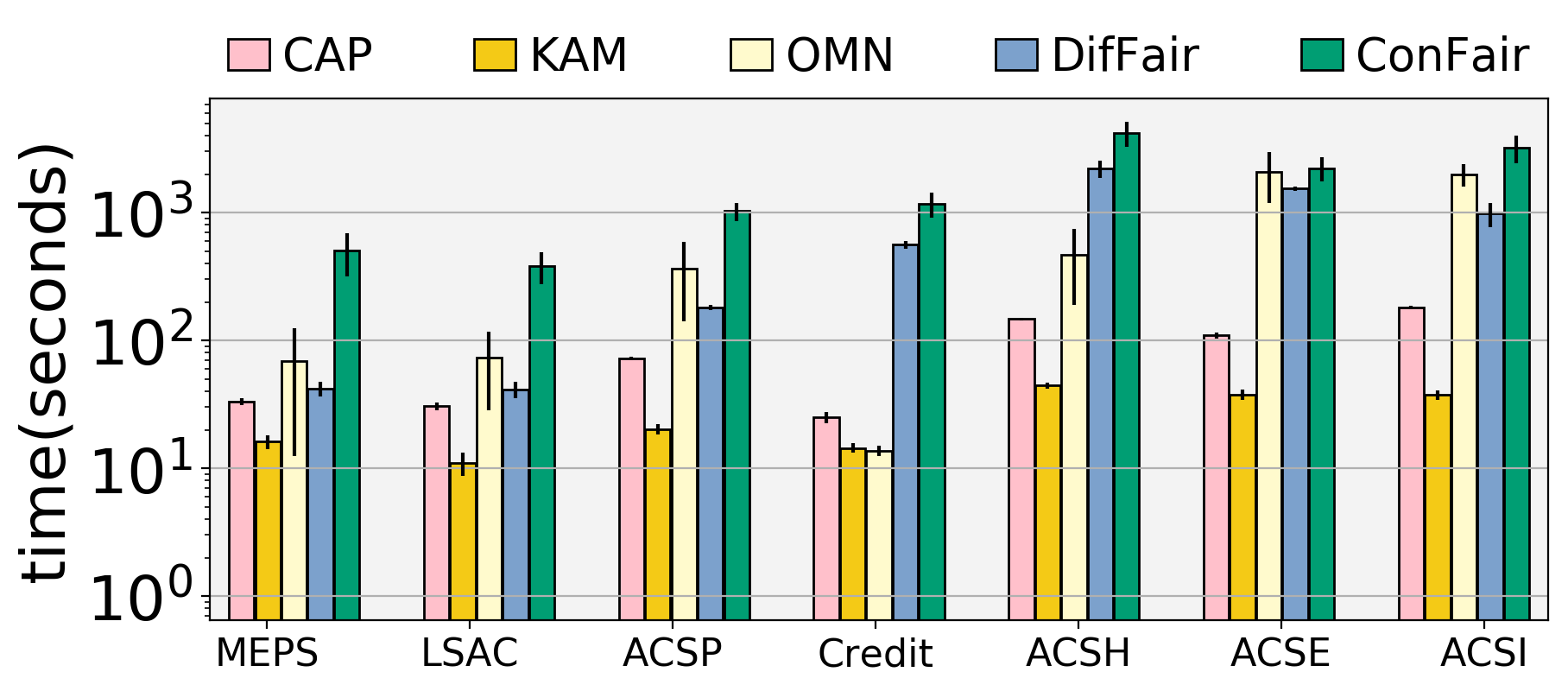}
    \label{fig:time_lr}}
    \\
    \subfloat[run-time, XGB models]{\includegraphics[width=0.9\linewidth]{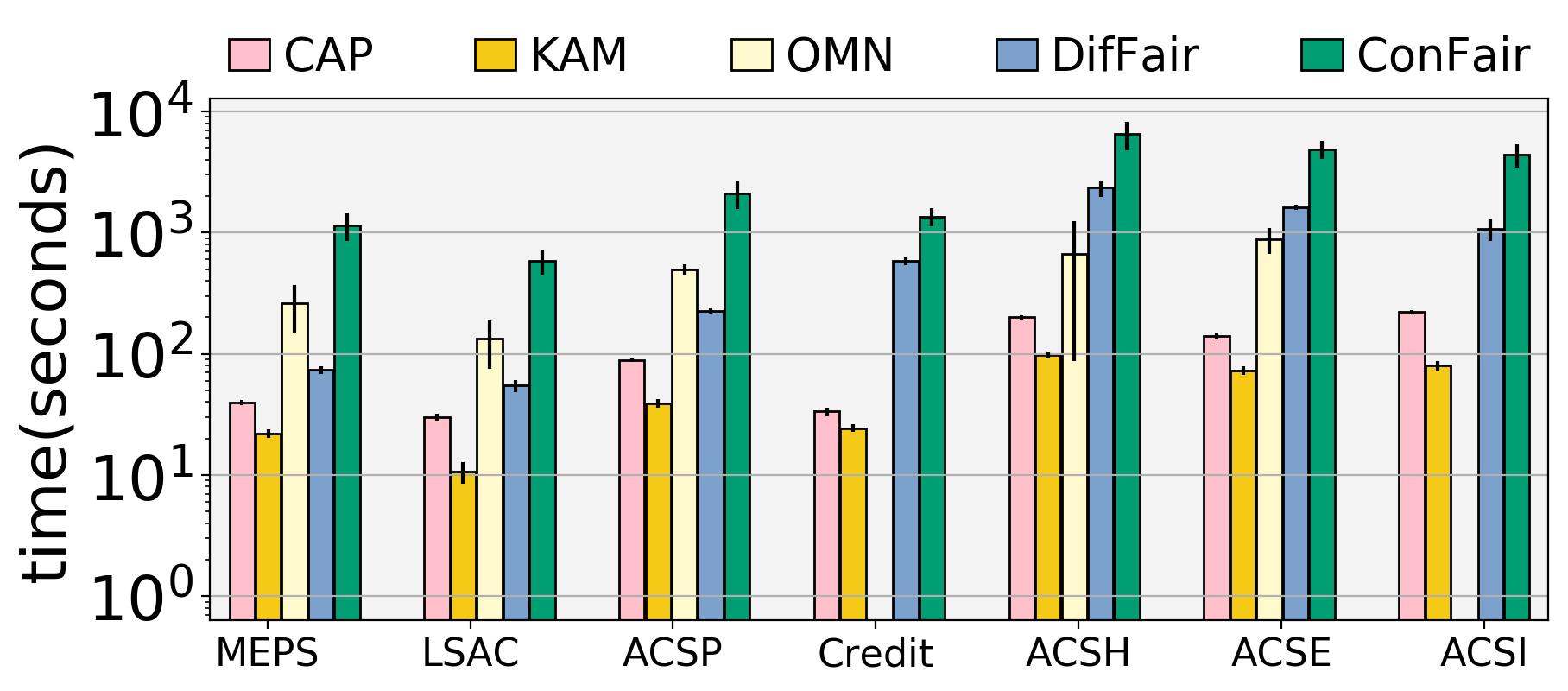}
 	\label{fig:time_tr}}
      \vspace{-0.1em}
    \caption{\scc has higher runtime due to CC derivation and weight calibration.  Its execution time can improve significantly with a user-supplied intervention degree.} 
    \vspace{-1em}
 	\label{fig:time}
\end{figure}

\revs{Figure~\ref{fig:mcc_syn} presents the results over the synthetic data for LR models\footnote{XGB models are not a good fit for the synthetic data due to the low separability of the minority group.}.} 
While \scc is generally the better choice over the real-world datasets, \mcc results in stronger fairness outcomes over the synthetic data.  These improvements have an impact on accuracy, which can be unavoidable in some cases, but the models remain reasonable. For comparison, we also display the results of \mbase, which uses the group mapping function $g$ to select which model to deploy for each tuple.  We observe that the use of CCs in \mcc results in starkly different behavior among the split-model strategies.
\revs{Figure~\ref{fig:mcc_real} presents the results over the real-world data for LR and XGB models.} \mcc performs comparably to \scc in most cases, but the latter is a better choice for two out of the five datasets.  The results considering \ead as the fairness metric display similar trends.

\smallskip
\begin{mdframed}[everyline=true,innertopmargin = 1mm,innerleftmargin = 1mm]
\textit{\textbf{Key takeaways:}} \mcc can be a better approach to improving fairness in learning, in scenarios where there is significant drift across groups and it is difficult to derive a single conforming model. 
\end{mdframed}

\subsection{Evaluation of CC optimization}
\label{sec:exp:opt}
Next, we examine the impact of the optimization in Algorithm~\ref{alg:density} on the performance of \mcc and \scc. Figure~\ref{fig:cc_opt} compares \mcc and\scc with variants {\mcc}$_0$ and {\scc}$_0$, which do not incorporate the density-based optimization, over the real-world datasets.  In both cases of LR (Figures~\ref{fig:cc_opt_lr_di},~\ref{fig:cc_opt_lr_aod}, and~\ref{fig:cc_opt_lr_balacc}) and XGB models (Figures~\ref{fig:cc_opt_tr_di},~\ref{fig:cc_opt_tr_aod}, and~\ref{fig:cc_opt_tr_balacc}), the density-based optimization of CCs leads to significant gains in the \edi metric. {\mcc}$_0$ in particular fails in most datasets, so this optimization is critical for \mcc.  The reason this optimization is so effective is that it significantly increases the discriminative power of the derived conformance constraints, thus rendering \mcc and \scc more effective.  The results follow similar trends for the \ead metric, and the utility of the models is not affected by the optimization. 

\subsection{Run-time evaluation}
\label{sec:exp:time}
In this section, we discuss the runtime of our methods with respect to other state of the art.  Figures~\ref{fig:time_lr} and~\ref{fig:time_tr} report the runtime of the algorithms over LR and XGB models, respectively. We note that these runtimes are largely dominated by model training, and the differences across reweighing techniques in particular depend on the number of times models are retrained.  

\kam is the fastest method, as it does not fine-tune its weights using the models.  Most of its execution time is taken up by training the final model, which is shared across methods.  The reason that \scc and \omn have higher execution times is because they calibrate their weights using the models.  This process involves several rounds of retraining during weight adjustment. 
We remind the reader that, while \scc often has higher runtime than \omn, the latter has significantly worse fairness and accuracy outcomes compared to \scc (see Section~\ref{sec:exp:scc}).  Note that Figure~\ref{fig:time_tr} does not report the \omn runtime for the \dcr and \dai datasets, because the method fails to produce a model in those cases, as its weights cause the XGB learner to not converge.

Our methods also consume some time in deriving the conformance constraints over each dataset, which is linear in the size of the data and cubic in the number of numerical attributes (see Section~\ref{sec:method_faircc} for details). Specifically, this cost dominates the runtime of \mcc, but this method does not need to retrain models as \scc does for its weight calibration.
We note that the runtime of \scc can be reduced significantly if users supply the intervention degree, instead of having the system automatically derive it through the fine-tuning process. This is not possible in \omn as it relies on the model output to fine-tune the weights, leading to a non-linear search space.

\capu's processing of the input data does not involve retraining the models, thus it can be faster than \scc and \omn.  Its runtime is similar to \mcc for the small datasets (\dmp, \dlg, and \dap), which means that \capu's data processing is similar to the cost of CC derivation in these cases.  In the larger datasets (\dcr, \dah, \dae, and \dai) CC derivation is about 10 times higher than \capu's processing time.

%% file: related.tex

\section{Related Work}
\label{sec:related}

\noindent
\textbf{Fair ML.} 
Algorithmic fairness has been studied extensively by researchers in machine learning and data management communities, among many others. Several technical reviews survey the topic from different perspectives in recent years~\cite{barocas2017fairness,DBLP:conf/icse/VermaR18,DBLP:conf/fat/FriedlerSVCHR19,DBLP:journals/cacm/ChouldechovaR20,berk2021fairness, doi:10.1146/annurev-statistics-042720-125902,IslamFMS2022}. 
Our method is similar to many methods that target group fairness or statistical parity to reduce ML bias, which requires equal decision rates across groups~\cite{DBLP:journals/datamine/CaldersV10,DBLP:journals/kais/KamiranC11,DBLP:conf/icdm/KamiranKZ12,DBLP:conf/pkdd/KamishimaAAS12,feldman2015certifying,DBLP:conf/nips/HardtPNS16,DBLP:conf/nips/CalmonWVRV17,DBLP:conf/kdd/ZhangWW17,DBLP:conf/nips/KusnerLRS17,DBLP:conf/aies/ZhangLM18,DBLP:journals/corr/abs-1808-00023,DBLP:conf/nips/LiptonMC18,DBLP:conf/icml/KearnsNRW18,DBLP:conf/icml/AgarwalBD0W18,DBLP:conf/icml/AgarwalDW19,johndrow2019algorithm,DBLP:conf/fat/CelisHKV19,SalimiRHS2019}. Another line of work focuses on individual fairness, motivated by the work of Dwork~\etal~\cite{DBLP:conf/innovations/DworkHPRZ12}, which requires that ML models assign predictions consistently for similar individuals.
Based on the stages of fairness interventions, research work can be categorized into three types: pre-, in-, and post-processing methods, while the interventions are applied on training data~\cite{DBLP:journals/kais/KamiranC11,feldman2015certifying,DBLP:conf/nips/CalmonWVRV17,DBLP:conf/kdd/ZhangWW17,SalimiRHS2019}, models~\cite{DBLP:conf/pkdd/KamishimaAAS12,DBLP:conf/aistats/ZafarVGG17,DBLP:conf/aies/ZhangLM18,DBLP:conf/icml/KearnsNRW18,DBLP:conf/icml/AgarwalBD0W18,DBLP:conf/icml/AgarwalDW19,DBLP:conf/fat/CelisHKV19,thomas2019preventing}, and predictions~\cite{DBLP:conf/icdm/KamiranKZ12,DBLP:conf/nips/HardtPNS16,DBLP:conf/nips/PleissRWKW17}, respectively. \scc can be considered a pre-processing intervention, as it operates before deriving ML models. 
\mcc would also be classified as a pre-processing technique, as it is based on splitting the data before model training.  While its primary insight lies in choosing which model to deploy based on conformance, it does not alter model outputs and thus does not fit in the post-processing category.
\rev{Data acquisition~\cite{chen2018my,asudeh2019assessing,tae2021slice,nargesian2021tailoring,chai2022selective} focuses on estimating the cost, benefit, and optimal strategies for collecting additional data; this setting is orthogonal to our case, where we only focus on the data that is available.}

The fairness literature has explored reweighing strategies as a method to improve model fairness~\cite{DBLP:journals/kais/KamiranC11,hashimoto2018fairness,lahoti2020fairness,jiang2020identifying,zhang2021omnifair}.
\scc and others~\cite{DBLP:journals/kais/KamiranC11,zhang2021omnifair} adjust the weights of tuples without modifying a learner, whereas many other approaches estimate the weights during the training of a model~\cite{krasanakis2018adaptive,hashimoto2018fairness,jiang2020identifying,lahoti2020fairness}. 
Li and Liu~\cite{li2022achieving} estimate the weights for tuples considering how they contribute to multiple fairness metrics and model loss by solving linear programs. \scc adjust weights based on the conformance of tuples to groups' data rather than merely on groups' representation~\cite{DBLP:journals/kais/KamiranC11} and on model outputs~\cite{zhang2021omnifair}. This also allows \scc to work with user-specified weights for designing fairness interventions tailored to specific tasks.

Kamiran~\etal~\cite{DBLP:journals/kais/KamiranC11} assigns identical weights for all tuples (within a group) to achieve a balanced representation of groups. The weights are computed based on the group's size and their original representation in the target label. Zhang~\etal~\cite{zhang2021omnifair} also follows this idea and adjusts the weights of tuples to achieve multiple fairness as measured by different metrics. Instead of assigning identical intervention on the weights for a group as in the previous two methods, \scc only increases the weights of the tuples that conform to the densest part of the group's data. This avoids amplifying outliers and noise that could mislead the training and harm model accuracy.  Hashimoto~\etal~\cite{hashimoto2018fairness} and Lahoti~\etal~\cite{lahoti2020fairness} iteratively adjust the tuples' weights during the continuous training of a fair model. The weights are modified based on how tuples contribute to the overall loss at each iteration, with the goal of achieving balanced loss across groups. Both of the approaches focus on the scenario where group membership is not available.
Nachum~\cite{jiang2020identifying} adjust tuples' weights iteratively during training based on how each tuple contributes to unbiased labels. Such unobserved labels are produced by perturbing the target labels~\cite{krasanakis2018adaptive} or are estimated by a mathematical framework~\cite{jiang2020identifying}.  
In contrast to these approaches, \scc does not modify weights during the training process but rather adjusts them prior to training. This allows \scc to work with user-specified weights for designing fairness interventions tailored to specific tasks.

\smallskip
\noindent\textbf{Data Drift.} ML researchers study data drift~\cite{quinonero2008dataset} with the goal of identifying the drift between different datasets such as training and serving data~\cite{DBLP:conf/sigmod/SchelterRB20} or identifying drift tuples such as out-of-distribution or misclassified records inside the input data~\cite{DBLP:conf/iclr/HendrycksG17,DBLP:journals/corr/abs-1812-02765,DBLP:conf/nips/JiangKGG18}. Examples of data drift include label~\cite{DBLP:conf/nips/HuangSGBS06,DBLP:conf/icml/ZhangSMW13,DBLP:conf/icml/LiptonWS18} and covariate shift~\cite{DBLP:journals/jmlr/BickelBS09, von2009finding, sugiyama2012machine}. 
Label shift assumes that the marginal distribution $p(\mathbf{Y})$ of labels $\mathbf{Y}$ changes, while the conditional distribution $p(\mathbf{X}|\mathbf{Y})$ of features $\mathbf{X}$ remains constant~\cite{DBLP:conf/nips/HuangSGBS06,DBLP:conf/icml/ZhangSMW13,DBLP:conf/icml/LiptonWS18}. 
Covariate shift assumes the opposite, where $p(\mathbf{X})$ changes with $p(\mathbf{Y}|\mathbf{X})$ being constant~\cite{DBLP:journals/jmlr/BickelBS09, von2009finding, sugiyama2012machine}. 
Lahoti~\cite{lahoti2021detecting} \etal focus on data shift between development and deployment stages by identifying different erroneous cases of a produced model on deploying data and advising actions (\eg collecting more training data) to mitigate the corresponding erroneous cases.
Instead of drift between multiple datasets, we focus on the drift over groups within an input, assuming that both label and covariate shifts might appear between groups. 

\smallskip
\noindent\textbf{Data Profiling.} 
Many data profiling techniques have been developed in data management to formalize different constraints that characterize the input data~\cite{DBLP:journals/vldb/AbedjanGN15}, such as functional dependencies~\cite{DBLP:journals/pvldb/PapenbrockEMNRZ15}, their
variants~\cite{DBLP:conf/sigmod/IlyasMHBA04, DBLP:conf/icde/KoudasSSV09,
DBLP:conf/icde/FanGLX09, DBLP:conf/ideas/CaruccioDP16,
DBLP:journals/pvldb/0001N18, DBLP:journals/pvldb/QahtanTOCS20,
DBLP:conf/sigmod/ZhangGR20, DBLP:conf/edbt/KaregarGGKSS21,DBLP:conf/icde/FanGLX09}, and the more general denial constraints~\cite{DBLP:journals/pvldb/ChuIP13,
DBLP:journals/pvldb/BleifussKN17, DBLP:journals/pvldb/PenaAN19,
DBLP:journals/pvldb/LivshitsHIK20}.
Compared to these, conformance constraints~\cite{DBLP:conf/sigmod/Fariha0RGM21} describe a dataset with arithmetic expressions of the relations among numerical attributes. Our methods can support other profiling techniques that provide quantitative measures of violations for the profiling constraints.

%% file: conc.tex

\section{Summary and Future Work}
\label{sec:conc}

In this paper, we recast the problem of fairness in ML models as a problem of data drift, and, consecutively, as an issue of conformance between data and models. 
We proposed two intervention strategies that employ conformance constraints in novel ways to achieve these conformance goals. Our model-splitting strategy, \mcc, trains separate models for different groups and uses conformance constraints (CCs) to determine the proper model to derive predictions. 
Our reweighing strategy, \scc, introduces a novel use of CCs to adjust the weights of tuples in the training data before feeding into an ML model. 
Our evaluation over seven real-world datasets showed that \scc outperforms prior art and is effectively model-agnostic, and \mcc can be a better option in cases of significant drift, where a single conforming model is unlikely. 

Future work can explore the integration of other data profiling techniques, which could potentially leverage a combination of attribute types to derive fairness interventions. 
An overarching goal is to automate fairness repairs and contribute to an end-to-end drift-driven repair system using techniques that detect internal drift and identify the relevant impacted subpopulations.

\smallskip
\noindent
\textbf{Acknowledgements.}  This research has been supported by the National Science Foundation through grant SHF-1763423, and by Google, through a Data Analytics and Insights (DANI) Award.

%% file: main.bbl
\begin{thebibliography}{10}
\providecommand{\url}[1]{#1}
\csname url@samestyle\endcsname
\providecommand{\newblock}{\relax}
\providecommand{\bibinfo}[2]{#2}
\providecommand{\BIBentrySTDinterwordspacing}{\spaceskip=0pt\relax}
\providecommand{\BIBentryALTinterwordstretchfactor}{4}
\providecommand{\BIBentryALTinterwordspacing}{\spaceskip=\fontdimen2\font plus
\BIBentryALTinterwordstretchfactor\fontdimen3\font minus
  \fontdimen4\font\relax}
\providecommand{\BIBforeignlanguage}[2]{{%
\expandafter\ifx\csname l@#1\endcsname\relax
\typeout{** WARNING: IEEEtran.bst: No hyphenation pattern has been}%
\typeout{** loaded for the language `#1'. Using the pattern for}%
\typeout{** the default language instead.}%
\else
\language=\csname l@#1\endcsname
\fi
#2}}
\providecommand{\BIBdecl}{\relax}
\BIBdecl

\bibitem{DBLP:journals/datamine/CaldersV10}
\BIBentryALTinterwordspacing
T.~Calders and S.~Verwer, ``Three naive bayes approaches for
  discrimination-free classification,'' \emph{Data Min. Knowl. Discov.},
  vol.~21, no.~2, pp. 277--292, 2010. [Online]. Available:
  \url{https://doi.org/10.1007/s10618-010-0190-x}
\BIBentrySTDinterwordspacing

\bibitem{DBLP:journals/kais/KamiranC11}
\BIBentryALTinterwordspacing
F.~Kamiran and T.~Calders, ``Data preprocessing techniques for classification
  without discrimination,'' \emph{Knowl. Inf. Syst.}, vol.~33, no.~1, pp.
  1--33, 2011. [Online]. Available:
  \url{https://doi.org/10.1007/s10115-011-0463-8}
\BIBentrySTDinterwordspacing

\bibitem{DBLP:conf/icdm/KamiranKZ12}
\BIBentryALTinterwordspacing
F.~Kamiran, A.~Karim, and X.~Zhang, ``Decision theory for discrimination-aware
  classification,'' in \emph{12th {IEEE} International Conference on Data
  Mining, {ICDM} 2012, Brussels, Belgium, December 10-13, 2012}, M.~J. Zaki,
  A.~Siebes, J.~X. Yu, B.~Goethals, G.~I. Webb, and X.~Wu, Eds.\hskip 1em plus
  0.5em minus 0.4em\relax {IEEE} Computer Society, 2012, pp. 924--929.
  [Online]. Available: \url{https://doi.org/10.1109/ICDM.2012.45}
\BIBentrySTDinterwordspacing

\bibitem{DBLP:conf/pkdd/KamishimaAAS12}
\BIBentryALTinterwordspacing
T.~Kamishima, S.~Akaho, H.~Asoh, and J.~Sakuma, ``Fairness-aware classifier
  with prejudice remover regularizer,'' in \emph{Machine Learning and Knowledge
  Discovery in Databases - European Conference, {ECML} {PKDD} 2012, Bristol,
  UK, September 24-28, 2012. Proceedings, Part {II}}, ser. Lecture Notes in
  Computer Science, P.~A. Flach, T.~D. Bie, and N.~Cristianini, Eds., vol.
  7524.\hskip 1em plus 0.5em minus 0.4em\relax Springer, 2012, pp. 35--50.
  [Online]. Available: \url{https://doi.org/10.1007/978-3-642-33486-3\_3}
\BIBentrySTDinterwordspacing

\bibitem{feldman2015certifying}
M.~Feldman, S.~A. Friedler, J.~Moeller, C.~Scheidegger, and
  S.~Venkatasubramanian, ``Certifying and removing disparate impact,'' in
  \emph{proceedings of the 21th ACM SIGKDD international conference on
  knowledge discovery and data mining}, 2015, pp. 259--268.

\bibitem{DBLP:conf/nips/HardtPNS16}
M.~Hardt, E.~Price, and N.~Srebro, ``Equality of opportunity in supervised
  learning,'' in \emph{NIPS}, 2016, pp. 3315--3323.

\bibitem{DBLP:conf/nips/CalmonWVRV17}
\BIBentryALTinterwordspacing
F.~P. Calmon, D.~Wei, B.~Vinzamuri, K.~N. Ramamurthy, and K.~R. Varshney,
  ``Optimized pre-processing for discrimination prevention,'' in \emph{Advances
  in Neural Information Processing Systems 30: Annual Conference on Neural
  Information Processing Systems 2017, December 4-9, 2017, Long Beach, CA,
  {USA}}, I.~Guyon, U.~von Luxburg, S.~Bengio, H.~M. Wallach, R.~Fergus,
  S.~V.~N. Vishwanathan, and R.~Garnett, Eds., 2017, pp. 3992--4001. [Online].
  Available:
  \url{https://proceedings.neurips.cc/paper/2017/hash/9a49a25d845a483fae4be7e341368e36-Abstract.html}
\BIBentrySTDinterwordspacing

\bibitem{DBLP:conf/kdd/ZhangWW17}
\BIBentryALTinterwordspacing
L.~Zhang, Y.~Wu, and X.~Wu, ``Achieving non-discrimination in data release,''
  in \emph{Proceedings of the 23rd {ACM} {SIGKDD} International Conference on
  Knowledge Discovery and Data Mining, Halifax, NS, Canada, August 13 - 17,
  2017}.\hskip 1em plus 0.5em minus 0.4em\relax {ACM}, 2017, pp. 1335--1344.
  [Online]. Available: \url{https://doi.org/10.1145/3097983.3098167}
\BIBentrySTDinterwordspacing

\bibitem{DBLP:conf/nips/KusnerLRS17}
\BIBentryALTinterwordspacing
M.~J. Kusner, J.~R. Loftus, C.~Russell, and R.~Silva, ``Counterfactual
  fairness,'' in \emph{Advances in Neural Information Processing Systems 30:
  Annual Conference on Neural Information Processing Systems 2017, December
  4-9, 2017, Long Beach, CA, {USA}}, I.~Guyon, U.~von Luxburg, S.~Bengio, H.~M.
  Wallach, R.~Fergus, S.~V.~N. Vishwanathan, and R.~Garnett, Eds., 2017, pp.
  4066--4076. [Online]. Available:
  \url{https://proceedings.neurips.cc/paper/2017/hash/a486cd07e4ac3d270571622f4f316ec5-Abstract.html}
\BIBentrySTDinterwordspacing

\bibitem{DBLP:conf/aies/ZhangLM18}
\BIBentryALTinterwordspacing
B.~H. Zhang, B.~Lemoine, and M.~Mitchell, ``Mitigating unwanted biases with
  adversarial learning,'' in \emph{Proceedings of the 2018 {AAAI/ACM}
  Conference on AI, Ethics, and Society, {AIES} 2018, New Orleans, LA, USA,
  February 02-03, 2018}, J.~Furman, G.~E. Marchant, H.~Price, and F.~Rossi,
  Eds.\hskip 1em plus 0.5em minus 0.4em\relax {ACM}, 2018, pp. 335--340.
  [Online]. Available: \url{https://doi.org/10.1145/3278721.3278779}
\BIBentrySTDinterwordspacing

\bibitem{DBLP:journals/corr/abs-1808-00023}
\BIBentryALTinterwordspacing
S.~Corbett{-}Davies and S.~Goel, ``The measure and mismeasure of fairness: {A}
  critical review of fair machine learning,'' \emph{CoRR}, vol. abs/1808.00023,
  2018. [Online]. Available: \url{http://arxiv.org/abs/1808.00023}
\BIBentrySTDinterwordspacing

\bibitem{DBLP:conf/nips/LiptonMC18}
\BIBentryALTinterwordspacing
Z.~C. Lipton, J.~J. McAuley, and A.~Chouldechova, ``Does mitigating ml's impact
  disparity require treatment disparity?'' in \emph{Advances in Neural
  Information Processing Systems 31: Annual Conference on Neural Information
  Processing Systems 2018, NeurIPS 2018, December 3-8, 2018, Montr{\'{e}}al,
  Canada}, S.~Bengio, H.~M. Wallach, H.~Larochelle, K.~Grauman,
  N.~Cesa{-}Bianchi, and R.~Garnett, Eds., 2018, pp. 8136--8146. [Online].
  Available:
  \url{https://proceedings.neurips.cc/paper/2018/hash/8e0384779e58ce2af40eb365b318cc32-Abstract.html}
\BIBentrySTDinterwordspacing

\bibitem{DBLP:conf/icml/KearnsNRW18}
\BIBentryALTinterwordspacing
M.~J. Kearns, S.~Neel, A.~Roth, and Z.~S. Wu, ``Preventing fairness
  gerrymandering: Auditing and learning for subgroup fairness,'' in
  \emph{Proceedings of the 35th International Conference on Machine Learning,
  {ICML} 2018, Stockholmsm{\"{a}}ssan, Stockholm, Sweden, July 10-15, 2018},
  ser. Proceedings of Machine Learning Research, J.~G. Dy and A.~Krause, Eds.,
  vol.~80.\hskip 1em plus 0.5em minus 0.4em\relax {PMLR}, 2018, pp. 2569--2577.
  [Online]. Available: \url{http://proceedings.mlr.press/v80/kearns18a.html}
\BIBentrySTDinterwordspacing

\bibitem{DBLP:conf/icml/AgarwalBD0W18}
\BIBentryALTinterwordspacing
A.~Agarwal, A.~Beygelzimer, M.~Dud{\'{\i}}k, J.~Langford, and H.~M. Wallach,
  ``A reductions approach to fair classification,'' in \emph{Proceedings of the
  35th International Conference on Machine Learning, {ICML} 2018,
  Stockholmsm{\"{a}}ssan, Stockholm, Sweden, July 10-15, 2018}, ser.
  Proceedings of Machine Learning Research, J.~G. Dy and A.~Krause, Eds.,
  vol.~80.\hskip 1em plus 0.5em minus 0.4em\relax {PMLR}, 2018, pp. 60--69.
  [Online]. Available: \url{http://proceedings.mlr.press/v80/agarwal18a.html}
\BIBentrySTDinterwordspacing

\bibitem{DBLP:conf/icml/AgarwalDW19}
\BIBentryALTinterwordspacing
A.~Agarwal, M.~Dud{\'{\i}}k, and Z.~S. Wu, ``Fair regression: Quantitative
  definitions and reduction-based algorithms,'' in \emph{Proceedings of the
  36th International Conference on Machine Learning, {ICML} 2019, 9-15 June
  2019, Long Beach, California, {USA}}, ser. Proceedings of Machine Learning
  Research, K.~Chaudhuri and R.~Salakhutdinov, Eds., vol.~97.\hskip 1em plus
  0.5em minus 0.4em\relax {PMLR}, 2019, pp. 120--129. [Online]. Available:
  \url{http://proceedings.mlr.press/v97/agarwal19d.html}
\BIBentrySTDinterwordspacing

\bibitem{johndrow2019algorithm}
J.~E. Johndrow and K.~Lum, ``An algorithm for removing sensitive information:
  application to race-independent recidivism prediction,'' \emph{The Annals of
  Applied Statistics}, vol.~13, no.~1, pp. 189--220, 2019.

\bibitem{DBLP:conf/fat/CelisHKV19}
\BIBentryALTinterwordspacing
L.~E. Celis, L.~Huang, V.~Keswani, and N.~K. Vishnoi, ``Classification with
  fairness constraints: {A} meta-algorithm with provable guarantees,'' in
  \emph{Proceedings of the Conference on Fairness, Accountability, and
  Transparency, FAT* 2019, Atlanta, GA, USA, January 29-31, 2019}, danah boyd
  and J.~H. Morgenstern, Eds.\hskip 1em plus 0.5em minus 0.4em\relax {ACM},
  2019, pp. 319--328. [Online]. Available:
  \url{https://doi.org/10.1145/3287560.3287586}
\BIBentrySTDinterwordspacing

\bibitem{SalimiRHS2019}
\BIBentryALTinterwordspacing
B.~Salimi, L.~Rodriguez, B.~Howe, and D.~Suciu, ``Interventional fairness:
  Causal database repair for algorithmic fairness,'' in \emph{Proceedings of
  the 2019 International Conference on Management of Data}, ser. SIGMOD
  '19.\hskip 1em plus 0.5em minus 0.4em\relax New York, NY, USA: Association
  for Computing Machinery, 2019, pp. 793–--810. [Online]. Available:
  \url{https://doi.org/10.1145/3299869.3319901}
\BIBentrySTDinterwordspacing

\bibitem{doi:10.1146/annurev-statistics-042720-125902}
\BIBentryALTinterwordspacing
S.~Mitchell, E.~Potash, S.~Barocas, A.~D'Amour, and K.~Lum, ``Algorithmic
  fairness: Choices, assumptions, and definitions,'' \emph{Annual Review of
  Statistics and Its Application}, vol.~8, no.~1, pp. 141--163, 2021. [Online].
  Available: \url{https://doi.org/10.1146/annurev-statistics-042720-125902}
\BIBentrySTDinterwordspacing

\bibitem{DBLP:journals/jmlr/BickelBS09}
\BIBentryALTinterwordspacing
S.~Bickel, M.~Br{\"{u}}ckner, and T.~Scheffer, ``Discriminative learning under
  covariate shift,'' \emph{J. Mach. Learn. Res.}, vol.~10, pp. 2137--2155,
  2009. [Online]. Available:
  \url{https://dl.acm.org/doi/10.5555/1577069.1755858}
\BIBentrySTDinterwordspacing

\bibitem{DBLP:journals/sigmod/KumarMNP15}
\BIBentryALTinterwordspacing
A.~Kumar, R.~McCann, J.~F. Naughton, and J.~M. Patel, ``Model selection
  management systems: The next frontier of advanced analytics,'' \emph{{SIGMOD}
  Rec.}, vol.~44, no.~4, pp. 17--22, 2015. [Online]. Available:
  \url{https://doi.org/10.1145/2935694.2935698}
\BIBentrySTDinterwordspacing

\bibitem{DBLP:conf/icml/LiptonWS18}
\BIBentryALTinterwordspacing
Z.~C. Lipton, Y.~Wang, and A.~J. Smola, ``Detecting and correcting for label
  shift with black box predictors,'' in \emph{Proceedings of the 35th
  International Conference on Machine Learning, {ICML} 2018,
  Stockholmsm{\"{a}}ssan, Stockholm, Sweden, July 10-15, 2018}, ser.
  Proceedings of Machine Learning Research, J.~G. Dy and A.~Krause, Eds.,
  vol.~80.\hskip 1em plus 0.5em minus 0.4em\relax {PMLR}, 2018, pp. 3128--3136.
  [Online]. Available: \url{http://proceedings.mlr.press/v80/lipton18a.html}
\BIBentrySTDinterwordspacing

\bibitem{DBLP:journals/sigmod/PolyzotisRWZ18}
\BIBentryALTinterwordspacing
N.~Polyzotis, S.~Roy, S.~E. Whang, and M.~Zinkevich, ``Data lifecycle
  challenges in production machine learning: {A} survey,'' \emph{{SIGMOD}
  Rec.}, vol.~47, no.~2, pp. 17--28, 2018. [Online]. Available:
  \url{https://doi.org/10.1145/3299887.3299891}
\BIBentrySTDinterwordspacing

\bibitem{DBLP:conf/sigmod/SchelterRB20}
\BIBentryALTinterwordspacing
S.~Schelter, T.~Rukat, and F.~Bie{\ss}mann, ``Learning to validate the
  predictions of black box classifiers on unseen data,'' in \emph{Proceedings
  of the 2020 International Conference on Management of Data, {SIGMOD}
  Conference 2020, online conference [Portland, OR, USA], June 14-19, 2020},
  D.~Maier, R.~Pottinger, A.~Doan, W.~Tan, A.~Alawini, and H.~Q. Ngo,
  Eds.\hskip 1em plus 0.5em minus 0.4em\relax {ACM}, 2020, pp. 1289--1299.
  [Online]. Available: \url{https://doi.org/10.1145/3318464.3380604}
\BIBentrySTDinterwordspacing

\bibitem{DBLP:conf/sigmod/Fariha0RGM21}
\BIBentryALTinterwordspacing
A.~Fariha, A.~Tiwari, A.~Radhakrishna, S.~Gulwani, and A.~Meliou, ``Conformance
  constraint discovery: Measuring trust in data-driven systems,'' in
  \emph{{SIGMOD} '21: International Conference on Management of Data, Virtual
  Event, China, June 20-25, 2021}, G.~Li, Z.~Li, S.~Idreos, and D.~Srivastava,
  Eds.\hskip 1em plus 0.5em minus 0.4em\relax {ACM}, 2021, pp. 499--512.
  [Online]. Available: \url{https://doi.org/10.1145/3448016.3452795}
\BIBentrySTDinterwordspacing

\bibitem{kappelhof2017survey}
J.~Kappelhof, ``Survey research and the quality of survey data among ethnic
  minorities,'' \emph{Total survey error in practice}, pp. 235--252, 2017.

\bibitem{hashimoto2018fairness}
T.~Hashimoto, M.~Srivastava, H.~Namkoong, and P.~Liang, ``Fairness without
  demographics in repeated loss minimization,'' in \emph{International
  Conference on Machine Learning}.\hskip 1em plus 0.5em minus 0.4em\relax PMLR,
  2018, pp. 1929--1938.

\bibitem{lahoti2020fairness}
P.~Lahoti, A.~Beutel, J.~Chen, K.~Lee, F.~Prost, N.~Thain, X.~Wang, and E.~Chi,
  ``Fairness without demographics through adversarially reweighted learning,''
  \emph{Advances in neural information processing systems}, vol.~33, pp.
  728--740, 2020.

\bibitem{zhang2021omnifair}
H.~Zhang, X.~Chu, A.~Asudeh, and S.~B. Navathe, ``Omnifair: A declarative
  system for model-agnostic group fairness in machine learning,'' in
  \emph{Proceedings of the 2021 International Conference on Management of
  Data}, 2021, pp. 2076--2088.

\bibitem{jiang2020identifying}
H.~Jiang and O.~Nachum, ``Identifying and correcting label bias in machine
  learning,'' in \emph{International Conference on Artificial Intelligence and
  Statistics}.\hskip 1em plus 0.5em minus 0.4em\relax PMLR, 2020, pp. 702--712.

\bibitem{DBLP:conf/innovations/DworkHPRZ12}
\BIBentryALTinterwordspacing
C.~Dwork, M.~Hardt, T.~Pitassi, O.~Reingold, and R.~S. Zemel, ``Fairness
  through awareness,'' in \emph{Innovations in Theoretical Computer Science
  2012, Cambridge, MA, USA, January 8-10, 2012}, 2012, pp. 214--226. [Online].
  Available: \url{https://doi.org/10.1145/2090236.2090255}
\BIBentrySTDinterwordspacing

\bibitem{DBLP:conf/nips/PleissRWKW17}
\BIBentryALTinterwordspacing
G.~Pleiss, M.~Raghavan, F.~Wu, J.~M. Kleinberg, and K.~Q. Weinberger, ``On
  fairness and calibration,'' in \emph{Advances in Neural Information
  Processing Systems 30: Annual Conference on Neural Information Processing
  Systems 2017, December 4-9, 2017, Long Beach, CA, {USA}}, I.~Guyon, U.~von
  Luxburg, S.~Bengio, H.~M. Wallach, R.~Fergus, S.~V.~N. Vishwanathan, and
  R.~Garnett, Eds., 2017, pp. 5680--5689. [Online]. Available:
  \url{https://proceedings.neurips.cc/paper/2017/hash/b8b9c74ac526fffbeb2d39ab038d1cd7-Abstract.html}
\BIBentrySTDinterwordspacing

\bibitem{DBLP:conf/aistats/ZafarVGG17}
\BIBentryALTinterwordspacing
M.~B. Zafar, I.~Valera, M.~Gomez{-}Rodriguez, and K.~P. Gummadi, ``Fairness
  constraints: Mechanisms for fair classification,'' in \emph{Proceedings of
  the 20th International Conference on Artificial Intelligence and Statistics,
  {AISTATS} 2017, 20-22 April 2017, Fort Lauderdale, FL, {USA}}, ser.
  Proceedings of Machine Learning Research, A.~Singh and X.~J. Zhu, Eds.,
  vol.~54.\hskip 1em plus 0.5em minus 0.4em\relax {PMLR}, 2017, pp. 962--970.
  [Online]. Available: \url{http://proceedings.mlr.press/v54/zafar17a.html}
\BIBentrySTDinterwordspacing

\bibitem{thomas2019preventing}
P.~S. Thomas, B.~Castro~da Silva, A.~G. Barto, S.~Giguere, Y.~Brun, and
  E.~Brunskill, ``Preventing undesirable behavior of intelligent machines,''
  \emph{Science}, vol. 366, no. 6468, pp. 999--1004, 2019.

\bibitem{martinez2020minimax}
N.~Martinez, M.~Bertran, and G.~Sapiro, ``Minimax pareto fairness: A multi
  objective perspective,'' in \emph{International Conference on Machine
  Learning}.\hskip 1em plus 0.5em minus 0.4em\relax PMLR, 2020, pp. 6755--6764.

\bibitem{krishnaswamy2021fair}
A.~Krishnaswamy, Z.~Jiang, K.~Wang, Y.~Cheng, and K.~Munagala, ``Fair for all:
  Best-effort fairness guarantees for classification,'' in \emph{International
  Conference on Artificial Intelligence and Statistics}.\hskip 1em plus 0.5em
  minus 0.4em\relax PMLR, 2021, pp. 3259--3267.

\bibitem{chhabra2021overview}
A.~Chhabra, K.~Masalkovait{\.e}, and P.~Mohapatra, ``An overview of fairness in
  clustering,'' \emph{IEEE Access}, vol.~9, pp. 130\,698--130\,720, 2021.

\bibitem{DBLP:journals/cacm/Bentley75}
\BIBentryALTinterwordspacing
J.~L. Bentley, ``Multidimensional binary search trees used for associative
  searching,'' \emph{Commun. {ACM}}, vol.~18, no.~9, pp. 509--517, 1975.
  [Online]. Available: \url{http://doi.acm.org/10.1145/361002.361007}
\BIBentrySTDinterwordspacing

\bibitem{scikit-learn}
F.~Pedregosa, G.~Varoquaux, A.~Gramfort, V.~Michel, B.~Thirion, O.~Grisel,
  M.~Blondel, P.~Prettenhofer, R.~Weiss, V.~Dubourg, J.~Vanderplas, A.~Passos,
  D.~Cournapeau, M.~Brucher, M.~Perrot, and E.~Duchesnay, ``Scikit-learn:
  Machine learning in {P}ython,'' \emph{Journal of Machine Learning Research},
  vol.~12, pp. 2825--2830, 2011.

\bibitem{wasserman2006all}
L.~Wasserman, \emph{All of nonparametric statistics}.\hskip 1em plus 0.5em
  minus 0.4em\relax Springer Science \& Business Media, 2006.

\bibitem{silverman2018density}
B.~W. Silverman, \emph{Density estimation for statistics and data
  analysis}.\hskip 1em plus 0.5em minus 0.4em\relax Routledge, 2018.

\bibitem{muller2004nonparametric}
P.~M{\"u}ller and F.~A. Quintana, ``Nonparametric bayesian data analysis,''
  \emph{Statistical science}, vol.~19, no.~1, pp. 95--110, 2004.

\bibitem{omohundro1989five}
S.~M. Omohundro, \emph{Five balltree construction algorithms}.\hskip 1em plus
  0.5em minus 0.4em\relax International Computer Science Institute Berkeley,
  1989.

\bibitem{aif360-oct-2018}
R.~K.~E. Bellamy, K.~Dey, M.~Hind, S.~C. Hoffman, S.~Houde, K.~Kannan,
  P.~Lohia, J.~Martino, S.~Mehta, A.~Mojsilovic, S.~Nagar, K.~N. Ramamurthy,
  J.~Richards, D.~Saha, P.~Sattigeri, M.~Singh, K.~R. Varshney, and Y.~Zhang,
  ``{AI Fairness} 360: An extensible toolkit for detecting, understanding, and
  mitigating unwanted algorithmic bias,''
  \url{https://arxiv.org/abs/1810.01943}, Oct. 2018.

\bibitem{kaggle_give_me_some_credit}
Kaggle, ``{Kaggle competition}: Give me some credit,''
  \url{https://www.kaggle.com/competitions/GiveMeSomeCredit/overview}, 2012.

\bibitem{ding2021retiring}
F.~Ding, M.~Hardt, J.~Miller, and L.~Schmidt, ``Retiring adult: New datasets
  for fair machine learning,'' \emph{Advances in neural information processing
  systems}, vol.~34, pp. 6478--6490, 2021.

\bibitem{adult}
\BIBentryALTinterwordspacing
R.~Kohavi and B.~Becker, ``{UCI} machine learning repository,'' 1994. [Online].
  Available: \url{http://archive.ics.uci.edu/ml/datasets/Adult}
\BIBentrySTDinterwordspacing

\bibitem{barocas2016big}
S.~Barocas and A.~D. Selbst, ``Big data's disparate impact,'' \emph{Calif. L.
  Rev.}, vol. 104, p. 671, 2016.

\bibitem{supp}
\BIBentryALTinterwordspacing
K.~Yang and A.~Meliou, ``Non-invasive fairness in learning through the lens of
  data drift (supplementary material),'' 2023. [Online]. Available:
  \url{https://github.com/DataProfilor/ConFair}
\BIBentrySTDinterwordspacing

\bibitem{krasanakis2018adaptive}
E.~Krasanakis, E.~Spyromitros-Xioufis, S.~Papadopoulos, and Y.~Kompatsiaris,
  ``Adaptive sensitive reweighting to mitigate bias in fairness-aware
  classification,'' in \emph{Proceedings of the 2018 world wide web
  conference}, 2018, pp. 853--862.

\bibitem{IslamFMS2022}
\BIBentryALTinterwordspacing
M.~T. Islam, A.~Fariha, A.~Meliou, and B.~Salimi,
  ``\href{https://dl.acm.org/doi/pdf/10.1145/3514221.3517841}{Through the Data
  Management Lens: Experimental Analysis and Evaluation of Fair
  Classification},'' in \emph{{SIGMOD} '22: International Conference on
  Management of Data}.\hskip 1em plus 0.5em minus 0.4em\relax {ACM}, 2022, pp.
  232--246. [Online]. Available: \url{https://doi.org/10.1145/3514221.3517841}
\BIBentrySTDinterwordspacing

\bibitem{barocas2017fairness}
S.~Barocas, M.~Hardt, and A.~Narayanan, ``Fairness in machine learning,''
  \emph{Nips tutorial}, vol.~1, p.~2, 2017.

\bibitem{DBLP:conf/icse/VermaR18}
\BIBentryALTinterwordspacing
S.~Verma and J.~Rubin, ``Fairness definitions explained,'' in \emph{Proceedings
  of the International Workshop on Software Fairness, FairWare@ICSE 2018,
  Gothenburg, Sweden, May 29, 2018}, Y.~Brun, B.~Johnson, and A.~Meliou,
  Eds.\hskip 1em plus 0.5em minus 0.4em\relax {ACM}, 2018, pp. 1--7. [Online].
  Available: \url{https://doi.org/10.1145/3194770.3194776}
\BIBentrySTDinterwordspacing

\bibitem{DBLP:conf/fat/FriedlerSVCHR19}
\BIBentryALTinterwordspacing
S.~A. Friedler, C.~Scheidegger, S.~Venkatasubramanian, S.~Choudhary, E.~P.
  Hamilton, and D.~Roth, ``A comparative study of fairness-enhancing
  interventions in machine learning,'' in \emph{Proceedings of the Conference
  on Fairness, Accountability, and Transparency, FAT* 2019, Atlanta, GA, USA,
  January 29-31, 2019}, danah boyd and J.~H. Morgenstern, Eds.\hskip 1em plus
  0.5em minus 0.4em\relax {ACM}, 2019, pp. 329--338. [Online]. Available:
  \url{https://doi.org/10.1145/3287560.3287589}
\BIBentrySTDinterwordspacing

\bibitem{DBLP:journals/cacm/ChouldechovaR20}
\BIBentryALTinterwordspacing
A.~Chouldechova and A.~Roth, ``A snapshot of the frontiers of fairness in
  machine learning,'' \emph{Commun. {ACM}}, vol.~63, no.~5, pp. 82--89, 2020.
  [Online]. Available: \url{https://doi.org/10.1145/3376898}
\BIBentrySTDinterwordspacing

\bibitem{berk2021fairness}
R.~Berk, H.~Heidari, S.~Jabbari, M.~Kearns, and A.~Roth, ``Fairness in criminal
  justice risk assessments: The state of the art,'' \emph{Sociological Methods
  \& Research}, vol.~50, no.~1, pp. 3--44, 2021.

\bibitem{chen2018my}
I.~Chen, F.~D. Johansson, and D.~Sontag, ``Why is my classifier
  discriminatory?'' \emph{Advances in neural information processing systems},
  vol.~31, 2018.

\bibitem{asudeh2019assessing}
A.~Asudeh, Z.~Jin, and H.~Jagadish, ``Assessing and remedying coverage for a
  given dataset,'' in \emph{2019 IEEE 35th International Conference on Data
  Engineering (ICDE)}.\hskip 1em plus 0.5em minus 0.4em\relax IEEE, 2019, pp.
  554--565.

\bibitem{tae2021slice}
K.~H. Tae and S.~E. Whang, ``Slice tuner: A selective data acquisition
  framework for accurate and fair machine learning models,'' in
  \emph{Proceedings of the 2021 International Conference on Management of
  Data}, 2021, pp. 1771--1783.

\bibitem{nargesian2021tailoring}
F.~Nargesian, A.~Asudeh, and H.~Jagadish, ``Tailoring data source distributions
  for fairness-aware data integration,'' \emph{Proceedings of the VLDB
  Endowment}, vol.~14, no.~11, pp. 2519--2532, 2021.

\bibitem{chai2022selective}
C.~Chai, J.~Liu, N.~Tang, G.~Li, and Y.~Luo, ``Selective data acquisition in
  the wild for model charging,'' \emph{Proceedings of the VLDB Endowment},
  vol.~15, no.~7, pp. 1466--1478, 2022.

\bibitem{li2022achieving}
P.~Li and H.~Liu, ``Achieving fairness at no utility cost via data reweighing
  with influence,'' in \emph{International Conference on Machine
  Learning}.\hskip 1em plus 0.5em minus 0.4em\relax PMLR, 2022, pp.
  12\,917--12\,930.

\bibitem{quinonero2008dataset}
J.~Qui{\~n}onero-Candela, M.~Sugiyama, A.~Schwaighofer, and N.~D. Lawrence,
  \emph{Dataset shift in machine learning}.\hskip 1em plus 0.5em minus
  0.4em\relax Mit Press, 2008.

\bibitem{DBLP:conf/iclr/HendrycksG17}
\BIBentryALTinterwordspacing
D.~Hendrycks and K.~Gimpel, ``A baseline for detecting misclassified and
  out-of-distribution examples in neural networks,'' in \emph{5th International
  Conference on Learning Representations, {ICLR} 2017, Toulon, France, April
  24-26, 2017, Conference Track Proceedings}.\hskip 1em plus 0.5em minus
  0.4em\relax OpenReview.net, 2017. [Online]. Available:
  \url{https://openreview.net/forum?id=Hkg4TI9xl}
\BIBentrySTDinterwordspacing

\bibitem{DBLP:journals/corr/abs-1812-02765}
\BIBentryALTinterwordspacing
T.~Denouden, R.~Salay, K.~Czarnecki, V.~Abdelzad, B.~Phan, and S.~Vernekar,
  ``Improving reconstruction autoencoder out-of-distribution detection with
  mahalanobis distance,'' \emph{CoRR}, vol. abs/1812.02765, 2018. [Online].
  Available: \url{http://arxiv.org/abs/1812.02765}
\BIBentrySTDinterwordspacing

\bibitem{DBLP:conf/nips/JiangKGG18}
\BIBentryALTinterwordspacing
H.~Jiang, B.~Kim, M.~Y. Guan, and M.~R. Gupta, ``To trust or not to trust {A}
  classifier,'' in \emph{Advances in Neural Information Processing Systems 31:
  Annual Conference on Neural Information Processing Systems 2018, NeurIPS
  2018, December 3-8, 2018, Montr{\'{e}}al, Canada}, S.~Bengio, H.~M. Wallach,
  H.~Larochelle, K.~Grauman, N.~Cesa{-}Bianchi, and R.~Garnett, Eds., 2018, pp.
  5546--5557. [Online]. Available:
  \url{https://proceedings.neurips.cc/paper/2018/hash/7180cffd6a8e829dacfc2a31b3f72ece-Abstract.html}
\BIBentrySTDinterwordspacing

\bibitem{DBLP:conf/nips/HuangSGBS06}
\BIBentryALTinterwordspacing
J.~Huang, A.~J. Smola, A.~Gretton, K.~M. Borgwardt, and B.~Sch{\"{o}}lkopf,
  ``Correcting sample selection bias by unlabeled data,'' in \emph{Advances in
  Neural Information Processing Systems 19, Proceedings of the Twentieth Annual
  Conference on Neural Information Processing Systems, Vancouver, British
  Columbia, Canada, December 4-7, 2006}, B.~Sch{\"{o}}lkopf, J.~C. Platt, and
  T.~Hofmann, Eds.\hskip 1em plus 0.5em minus 0.4em\relax {MIT} Press, 2006,
  pp. 601--608. [Online]. Available:
  \url{https://proceedings.neurips.cc/paper/2006/hash/a2186aa7c086b46ad4e8bf81e2a3a19b-Abstract.html}
\BIBentrySTDinterwordspacing

\bibitem{DBLP:conf/icml/ZhangSMW13}
\BIBentryALTinterwordspacing
K.~Zhang, B.~Sch{\"{o}}lkopf, K.~Muandet, and Z.~Wang, ``Domain adaptation
  under target and conditional shift,'' in \emph{Proceedings of the 30th
  International Conference on Machine Learning, {ICML} 2013, Atlanta, GA, USA,
  16-21 June 2013}, ser. {JMLR} Workshop and Conference Proceedings,
  vol.~28.\hskip 1em plus 0.5em minus 0.4em\relax JMLR.org, 2013, pp. 819--827.
  [Online]. Available: \url{http://proceedings.mlr.press/v28/zhang13d.html}
\BIBentrySTDinterwordspacing

\bibitem{von2009finding}
P.~Von~B{\"u}nau, F.~C. Meinecke, F.~C. Kir{\'a}ly, and K.-R. M{\"u}ller,
  ``Finding stationary subspaces in multivariate time series,'' \emph{Physical
  review letters}, vol. 103, no.~21, p. 214101, 2009.

\bibitem{sugiyama2012machine}
M.~Sugiyama and M.~Kawanabe, \emph{Machine learning in non-stationary
  environments: Introduction to covariate shift adaptation}.\hskip 1em plus
  0.5em minus 0.4em\relax MIT press, 2012.

\bibitem{lahoti2021detecting}
P.~Lahoti, K.~P. Gummadi, and G.~Weikum, ``Detecting and mitigating test-time
  failure risks via model-agnostic uncertainty learning,'' in \emph{2021 IEEE
  International Conference on Data Mining (ICDM)}.\hskip 1em plus 0.5em minus
  0.4em\relax IEEE, 2021, pp. 1174--1179.

\bibitem{DBLP:journals/vldb/AbedjanGN15}
\BIBentryALTinterwordspacing
Z.~Abedjan, L.~Golab, and F.~Naumann, ``Profiling relational data: a survey,''
  \emph{{VLDB} J.}, vol.~24, no.~4, pp. 557--581, 2015. [Online]. Available:
  \url{https://doi.org/10.1007/s00778-015-0389-y}
\BIBentrySTDinterwordspacing

\bibitem{DBLP:journals/pvldb/PapenbrockEMNRZ15}
\BIBentryALTinterwordspacing
T.~Papenbrock, J.~Ehrlich, J.~Marten, T.~Neubert, J.~Rudolph,
  M.~Sch{\"{o}}nberg, J.~Zwiener, and F.~Naumann, ``Functional dependency
  discovery: An experimental evaluation of seven algorithms,'' \emph{Proc.
  {VLDB} Endow.}, vol.~8, no.~10, pp. 1082--1093, 2015. [Online]. Available:
  \url{http://www.vldb.org/pvldb/vol8/p1082-papenbrock.pdf}
\BIBentrySTDinterwordspacing

\bibitem{DBLP:conf/sigmod/IlyasMHBA04}
\BIBentryALTinterwordspacing
I.~F. Ilyas, V.~Markl, P.~J. Haas, P.~Brown, and A.~Aboulnaga, ``{CORDS:}
  automatic discovery of correlations and soft functional dependencies,'' in
  \emph{Proceedings of the {ACM} {SIGMOD} International Conference on
  Management of Data, Paris, France, June 13-18, 2004}, G.~Weikum, A.~C.
  K{\"{o}}nig, and S.~De{\ss}loch, Eds.\hskip 1em plus 0.5em minus 0.4em\relax
  {ACM}, 2004, pp. 647--658. [Online]. Available:
  \url{https://doi.org/10.1145/1007568.1007641}
\BIBentrySTDinterwordspacing

\bibitem{DBLP:conf/icde/KoudasSSV09}
\BIBentryALTinterwordspacing
N.~Koudas, A.~Saha, D.~Srivastava, and S.~Venkatasubramanian, ``Metric
  functional dependencies,'' in \emph{Proceedings of the 25th International
  Conference on Data Engineering, {ICDE} 2009, March 29 2009 - April 2 2009,
  Shanghai, China}, Y.~E. Ioannidis, D.~L. Lee, and R.~T. Ng, Eds.\hskip 1em
  plus 0.5em minus 0.4em\relax {IEEE} Computer Society, 2009, pp. 1275--1278.
  [Online]. Available: \url{https://doi.org/10.1109/ICDE.2009.219}
\BIBentrySTDinterwordspacing

\bibitem{DBLP:conf/icde/FanGLX09}
\BIBentryALTinterwordspacing
W.~Fan, F.~Geerts, L.~V.~S. Lakshmanan, and M.~Xiong, ``Discovering conditional
  functional dependencies,'' in \emph{Proceedings of the 25th International
  Conference on Data Engineering, {ICDE} 2009, March 29 2009 - April 2 2009,
  Shanghai, China}, Y.~E. Ioannidis, D.~L. Lee, and R.~T. Ng, Eds.\hskip 1em
  plus 0.5em minus 0.4em\relax {IEEE} Computer Society, 2009, pp. 1231--1234.
  [Online]. Available: \url{https://doi.org/10.1109/ICDE.2009.208}
\BIBentrySTDinterwordspacing

\bibitem{DBLP:conf/ideas/CaruccioDP16}
\BIBentryALTinterwordspacing
L.~Caruccio, V.~Deufemia, and G.~Polese, ``On the discovery of relaxed
  functional dependencies,'' in \emph{Proceedings of the 20th International
  Database Engineering {\&} Applications Symposium, {IDEAS} 2016, Montreal, QC,
  Canada, July 11-13, 2016}, E.~Desai, B.~C. Desai, M.~Toyama, and
  J.~Bernardino, Eds.\hskip 1em plus 0.5em minus 0.4em\relax {ACM}, 2016, pp.
  53--61. [Online]. Available: \url{https://doi.org/10.1145/2938503.2938519}
\BIBentrySTDinterwordspacing

\bibitem{DBLP:journals/pvldb/0001N18}
\BIBentryALTinterwordspacing
S.~Kruse and F.~Naumann, ``Efficient discovery of approximate dependencies,''
  \emph{Proc. {VLDB} Endow.}, vol.~11, no.~7, pp. 759--772, 2018. [Online].
  Available: \url{http://www.vldb.org/pvldb/vol11/p759-kruse.pdf}
\BIBentrySTDinterwordspacing

\bibitem{DBLP:journals/pvldb/QahtanTOCS20}
\BIBentryALTinterwordspacing
A.~A. Qahtan, N.~Tang, M.~Ouzzani, Y.~Cao, and M.~Stonebraker, ``Pattern
  functional dependencies for data cleaning,'' \emph{Proc. {VLDB} Endow.},
  vol.~13, no.~5, pp. 684--697, 2020. [Online]. Available:
  \url{http://www.vldb.org/pvldb/vol13/p684-qahtan.pdf}
\BIBentrySTDinterwordspacing

\bibitem{DBLP:conf/sigmod/ZhangGR20}
\BIBentryALTinterwordspacing
Y.~Zhang, Z.~Guo, and T.~Rekatsinas, ``A statistical perspective on discovering
  functional dependencies in noisy data,'' in \emph{Proceedings of the 2020
  International Conference on Management of Data, {SIGMOD} Conference 2020,
  online conference [Portland, OR, USA], June 14-19, 2020}, D.~Maier,
  R.~Pottinger, A.~Doan, W.~Tan, A.~Alawini, and H.~Q. Ngo, Eds.\hskip 1em plus
  0.5em minus 0.4em\relax {ACM}, 2020, pp. 861--876. [Online]. Available:
  \url{https://doi.org/10.1145/3318464.3389749}
\BIBentrySTDinterwordspacing

\bibitem{DBLP:conf/edbt/KaregarGGKSS21}
\BIBentryALTinterwordspacing
R.~Karegar, P.~Godfrey, L.~Golab, M.~Kargar, D.~Srivastava, and J.~Szlichta,
  ``Efficient discovery of approximate order dependencies,'' in
  \emph{Proceedings of the 24th International Conference on Extending Database
  Technology, {EDBT} 2021, Nicosia, Cyprus, March 23 - 26, 2021},
  Y.~Velegrakis, D.~Zeinalipour{-}Yazti, P.~K. Chrysanthis, and F.~Guerra,
  Eds.\hskip 1em plus 0.5em minus 0.4em\relax OpenProceedings.org, 2021, pp.
  427--432. [Online]. Available: \url{https://doi.org/10.5441/002/edbt.2021.46}
\BIBentrySTDinterwordspacing

\bibitem{DBLP:journals/pvldb/ChuIP13}
\BIBentryALTinterwordspacing
X.~Chu, I.~F. Ilyas, and P.~Papotti, ``Discovering denial constraints,''
  \emph{Proc. {VLDB} Endow.}, vol.~6, no.~13, pp. 1498--1509, 2013. [Online].
  Available: \url{http://www.vldb.org/pvldb/vol6/p1498-papotti.pdf}
\BIBentrySTDinterwordspacing

\bibitem{DBLP:journals/pvldb/BleifussKN17}
\BIBentryALTinterwordspacing
T.~Bleifu{\ss}, S.~Kruse, and F.~Naumann, ``Efficient denial constraint
  discovery with hydra,'' \emph{Proc. {VLDB} Endow.}, vol.~11, no.~3, pp.
  311--323, 2017. [Online]. Available:
  \url{http://www.vldb.org/pvldb/vol11/p311-bleifub.pdf}
\BIBentrySTDinterwordspacing

\bibitem{DBLP:journals/pvldb/PenaAN19}
\BIBentryALTinterwordspacing
E.~H.~M. Pena, E.~C. de~Almeida, and F.~Naumann, ``Discovery of approximate
  (and exact) denial constraints,'' \emph{Proc. {VLDB} Endow.}, vol.~13, no.~3,
  pp. 266--278, 2019. [Online]. Available:
  \url{http://www.vldb.org/pvldb/vol13/p266-pena.pdf}
\BIBentrySTDinterwordspacing

\bibitem{DBLP:journals/pvldb/LivshitsHIK20}
\BIBentryALTinterwordspacing
E.~Livshits, A.~Heidari, I.~F. Ilyas, and B.~Kimelfeld, ``Approximate denial
  constraints,'' \emph{Proc. {VLDB} Endow.}, vol.~13, no.~10, pp. 1682--1695,
  2020. [Online]. Available:
  \url{http://www.vldb.org/pvldb/vol13/p1682-livshits.pdf}
\BIBentrySTDinterwordspacing

\end{thebibliography}
